\documentclass[5p,times]{elsarticle}
\usepackage{mathalfa}
 \usepackage{soul,xcolor}
\sethlcolor{yellow}
\usepackage{amsmath}
\usepackage{academicons}
\definecolor{orcidlogocol}{HTML}{A6CE39}
\usepackage{cite}
\usepackage{amsmath,amssymb,amsfonts}
 \usepackage{enumitem}          
\usepackage{adjustbox}
\usepackage{graphicx}
\usepackage{textcomp}
\usepackage{algpseudocode}
\usepackage{algorithm}
\usepackage{soul}
 \usepackage{multirow}
\usepackage{hyperref}
\usepackage[table]{xcolor}
\usepackage[graphicx]{realboxes}
 \usepackage{bm}
 \usepackage{booktabs}
\newcommand{\mat}[1]{\mathbf{#1}}   

\newcommand{\mX}{\mat{X}}   
   
\newcommand{\mW}{\mat{W}}
      
   \newcommand{\vtheta}{\bm{\theta}}

\newcommand{\mA}{\mat{A}}    
\newcommand{\mB}{\mat{B}} 
\makeatletter
\AtBeginDocument{\DeclareMathVersion{bold}
\DeclareSymbolFont{NewLetters}{T1}{times}{m}{it}
\SetSymbolFont{operators}{bold}{T1}{times}{b}{n}
\SetSymbolFont{NewLetters}{bold}{T1}{times}{b}{it}
\SetMathAlphabet{\mathrm}{bold}{T1}{times}{b}{n}
\SetMathAlphabet{\mathit}{bold}{T1}{times}{b}{it}
\SetMathAlphabet{\mathbf}{bold}{T1}{times}{b}{n}
\SetMathAlphabet{\mathtt}{bold}{OT1}{pcr}{b}{n}
\SetSymbolFont{symbols}{bold}{OMS}{cmsy}{b}{n}
\renewcommand\boldmath{\@nomath\boldmath\mathversion{bold}}}
\makeatother

\def\BibTeX{{\rm B\kern-.05em{\sc i\kern-.025em b}\kern-.08em
    T\kern-.1667em\lower.7ex\hbox{E}\kern-.125emX}}

\journal{Journal}

\begin{document}
\begin{frontmatter}
\title{
Refining Decision Boundaries In Anomaly Detection Using Similarity Search Within the Feature Space.}

\author[nyu]{Sidahmed Benabderrahmane\corref{cor1}}
\ead{sidahmed.benabderrahmane@nyu.edu}

\author[uqam]{Petko Valtchev}
\ead{valtchev.petko@uqam.ca}

\author[edin]{James Cheney}
\ead{jcheney@inf.ed.ac.uk}

\author[nyu]{Talal Rahwan}
\ead{talal.rahwan@nyu.edu}

\cortext[cor1]{Corresponding author}

\address[nyu]{New York University, NYUAD, Division of Science, Computer Science Department.}
\address[uqam]{University of Quebec in Montreal, Computer Science Department, Montreal.}
\address[edin]{University of Edinburgh, School of Informatics, Edinburgh.}



%

\begin{abstract}   
Detecting rare and diverse anomalies in highly imbalanced datasets—such as Advanced Persistent Threats (APTs) in cybersecurity—remains a fundamental challenge for machine learning systems. Active learning offers a promising direction by strategically querying an oracle to minimize labeling effort, yet conventional approaches often fail to exploit the intrinsic geometric structure of the feature space for model refinement. In this paper, we introduce SDA²E, a Sparse Dual Adversarial Attention-based AutoEncoder designed to learn compact and discriminative latent representations from imbalanced, high-dimensional data. We further propose a similarity-guided active learning framework that integrates three novel strategies to refine decision boundaries efficiently: mormal-like expansion, which enriches the training set with points similar to labeled normals to improve reconstruction fidelity; anomaly-like prioritization, which boosts ranking accuracy by focusing on points resembling known anomalies; and a hybrid strategy that combines both for balanced model refinement and ranking. A key component of our framework is a new similarity measure, Normalized Matching 1s ($SIM_{NM1}$), tailored for sparse binary embeddings. We evaluate SDA²E extensively across 52 imbalanced datasets, including multiple DARPA Transparent Computing scenarios, and benchmark it against 15 state-of-the-art anomaly detection methods. Results demonstrate that SDA²E consistently achieves superior ranking performance (nDCG up to 1.0 in several cases) while reducing the required labeled data by up to 80\% compared to passive training. Statistical tests confirm the significance of these improvements. Our work establishes a robust, efficient, and statistically validated framework for anomaly detection that is particularly suited to cybersecurity applications such as APT detection.\\
\end{abstract}
 \end{frontmatter}

\section{Introduction}
\textit{``It has long been an axiom of mine that the little things are infinitely the most important.''} --- Arthur Conan Doyle, The Memoirs of Sherlock Holmes. Nowhere is this insight more relevant than in anomaly detection, an essential task in domains such as fraud detection, cybersecurity, and industrial monitoring \citep{agrawal2015survey,taha2019anomaly,samariya2023comprehensive,inbookFontes22}. Unlike traditional classification tasks, anomaly detection deals with the identification of rare and unusual data patterns that significantly differ from the majority of normal samples. The rarity and diversity of anomalies pose challenges for most machine learning models, particularly in settings with high-dimensional or imbalanced datasets \citep{goldstein2016comparative,Kong2020}. A critical bottleneck is the severe imbalance between normal and anomalous samples, which can bias machine learning models toward overfitting to the dominant normal class, leading to reduced sensitivity to anomalies \citep{inbookKong20,Das10100135}. Furthermore, anomalies are often heterogeneous, with diverse characteristics that make it difficult for a single model to capture their patterns comprehensively. High-dimensional data exacerbate the issue by introducing noise and irrelevant features that obscure meaningful relationships \citep{piciarelli2019}. Additionally, labeled data for anomalies are typically scarce, as obtaining such labels is expensive, time-consuming, and sometimes infeasible \citep{Yanli22}. These challenges collectively hinder the ability of standard machine learning methods to learn precise decision boundaries, motivating the need for specialized approaches such as deep learning and active learning frameworks that are capable of exploiting structure within the feature space and incrementally improving with limited labeled data.

Active learning provides a promising solution for these challenges by allowing the model to iteratively query an oracle (e.g., a human expert) for labels on the most uncertain samples \citep{brame2016active}. This process would minimize the labeling effort while improving model performance. However, most existing active learning methods for anomaly detection rely on uncertainty sampling or query-by-committee strategies, treating each query in isolation without leveraging the intrinsic geometric relationships among data points in the feature space. This limitation is especially pronounced in highly imbalanced settings where anomalies are sparse and diverse; such approaches often fail to systematically refine decision boundaries because they do not exploit the latent structure that governs normal and anomalous behavior. To overcome this gap, we propose a novel integration of similarity search into the active learning loop, enabling the model to strategically exploit proximity and relational patterns within the feature space. Unlike conventional active learning, which focuses solely on selecting uncertain points, our approach uses similarity to:
\begin{itemize}
    \item     Expand and solidify the representation of normal behavior by retrieving unlabeled points that resemble labeled normal examples, thereby improving the model’s reconstruction fidelity;

    \item Prioritize potential anomalies by searching for points similar to known anomalies, boosting the ranking of true positives; and

    \item Balance exploration and refinement through a hybrid strategy that concurrently strengthens the model and sharpens the anomaly ranking.
\end{itemize}

This similarity-guided active learning framework is built on a newly designed Sparse Dual Adversarial Attention-based AutoEncoder (SDA²E), which combines sparsity constraints, adversarial training, and attention mechanisms to learn compact, discriminative latent representations from high-dimensional imbalanced data. The sparsity regularization forces the model to focus on salient features, adversarial learning enhances the realism and separability of latent codes, and attention dynamically weights important dimensions. When combined with similarity-driven sample selection, SDA²E iteratively refines decision boundaries with minimal oracle interaction, offering a principled way to jointly optimize model training and anomaly ranking.

Existing anomaly detection methods—whether based on statistical models, classical machine learning, or modern deep learning—often treat feature-space geometry as static or ignore it entirely during active learning. In contrast, our work explicitly grounds the active learning process in the geometric structure of the feature space, transforming similarity from a passive distance measure into an active refinement mechanism. This integration represents a significant departure from prior art and enables more efficient, robust, and interpretable anomaly detection, particularly in security-critical applications such as Advanced Persistent Threat (APT) detection, where anomalies are stealthy, heterogeneous, and extremely rare.

The primary objective of this paper is to establish a unified framework that seamlessly couples architectural innovation (SDA²E) with similarity-aware active learning, demonstrating its effectiveness across a wide range of imbalanced datasets while maintaining rigorous statistical validation. We focus particularly on APT detection as a demanding real-world application, but the framework is general and applicable to other domains where anomalies are rare and costly to label.
\noindent
The main contributions of this work can be summarized as follows:

\begin{itemize}
\item We propose a novel deep architecture, SDA²E, that combines sparsity regularization, attention mechanisms, and adversarial learning to produce compact and discriminative latent representations for anomaly detection in high-dimensional tabular data.
\item Similarity-Guided Active Learning:
We introduce a feedback-driven active learning loop that refines decision boundaries in the latent feature space through similarity search. Three complementary strategies (Normal-like, Anomalous-like, and Hybrid) guide sample selection for efficient label acquisition, moving beyond conventional uncertainty-based query strategies.

\item A New Similarity Metric:
We define a new binary-vector similarity measure, Normalized Matching 1s ($SIM_{NM1}$), that enhances discrimination among sparse high-dimensional embeddings and is computationally efficient for large-scale active learning.

\item Comprehensive and Statistically Validated Evaluation:
We conduct extensive experiments across 52 imbalanced datasets, including multiple DARPA Transparent Computing scenarios, and benchmark against fifteen state-of-the-art methods. Results are analyzed using ranking-based metrics and validated through statistical tests, demonstrating consistent and significant improvements in detection performance while reducing labeling effort by up to 80\% compared to passive training.
\end{itemize}

\noindent Together, these contributions establish a unified framework that integrates architectural innovation, active-learning efficiency, and statistical rigor for robust anomaly detection in cybersecurity and other domains.

The remainder of the paper is organized as follows. Section 2 reviews existing methods for anomaly detection across several domains. Section 3 describes the proposed framework, detailing the SDA²E architecture and the three similarity search strategies in active learning. Section 4 presents the experimental settings, results, and discussions. Finally, Section 5 summarizes the main results of this work and outlines future research directions.

\section{Related work}
\subsection{Overview}
Anomaly detection is a critical task in data analysis, aimed at identifying patterns or behaviors that deviate significantly from the norm \citep{chalapathy2019deep,qiao2025deep, zamanzadeh2024deep,liu2024time}. In most real-world datasets, the majority of observations follow a common distribution, representing typical system behavior. However, anomalies—also known as outliers—are data points that deviate from this expected pattern, often indicating irregular activities, faults, or malicious events \citep{lim2024future}. Anomalies can occur in various domains, often indicating fraud, system failures, or rare but critical events \citep{chalapathy2020robust}. For instance, in cybersecurity, anomalies may correspond to Advanced Persistent Threats (APTs), unauthorized access, or system intrusions \citep{rani2025comprehensive,ness2025anomaly}. In finance, anomalies may be observed during multiple rapid transactions of a debit card within a short time \citep{vilella2025weirdnodes}. In epidemiology, anomalies may represent the case where unexpected disease clusters (e.g. cancer) appear in a small geographic area \citep{de2025explainable}. In industrial and IoT systems, anomalies may appear when a factory suddenly consumes many times the usual power at night \citep{chen2025frequency,kundacina2025conformal,liu2025unsupervised,mancy2025swiniot}.

Detecting such anomalies is challenging due to their rarity and the presence of noise in high-dimensional data. Machine learning techniques, including AutoEncoders and clustering methods, help uncover these deviations by learning normal system behavior and flagging instances that exhibit significant divergence from learned patterns. Anomaly detection has been extensively studied for decades, and several surveys chronicle the evolution of its techniques, from early statistical models to modern machine learning approaches \citep{wang2025fusionformer,liu2025research}. These methods can broadly be categorized into several approaches, depending on their underlying assumptions, data requirements, and computational frameworks \citep{zhang2025frect,wang2025pre}. These include:

\subsection{Statistical Methods}
Statistical approaches to anomaly detection rely on the assumption that normal data can be modeled using a probability distribution, and anomalies are observations that deviate significantly from this distribution. These methods are often favored for their interpretability and mathematical rigor, particularly in low-dimensional or well-behaved data settings. Common statistical techniques include:

\begin{itemize}
\item Parametric models, such as Gaussian Mixture Models (GMMs), which approximate the data distribution using a weighted combination of Gaussian components. Anomalies are identified as points with low likelihood under the fitted model \citep{a16040195}.
\item Hypothesis testing methods, including the $t$-test, $z$-test, and Grubbs’ test, which evaluate whether a given observation likely originates from the same distribution as the majority of data \citep{KAMENIK2023137836}. These are effective for univariate or low-dimensional data but scale poorly to high-dimensional settings.
\item Density-based approaches, such as Kernel Density Estimation (KDE), which estimate the probability density function of the normal data non-parametrically. Observations lying in low-density regions are flagged as anomalies \citep{Frehner10453576}.
\end{itemize}

While statistical methods provide a principled foundation and are computationally efficient for small to moderate datasets, they face notable limitations in modern anomaly detection contexts. They often struggle with \textit{high-dimensional data} due to the curse of dimensionality, which makes density estimation unreliable. Moreover, they generally assume that the normal data follow a stationary distribution and are sensitive to \textit{nonlinear relationships} and complex dependencies among features. These constraints limit their applicability to real-world problems such as cybersecurity logs or system-level provenance data, where anomalies are sparse, high-dimensional, and exhibit non-Gaussian structure. Consequently, more flexible machine learning and deep learning paradigms have emerged to address these challenges.
\subsection{Machine Learning Approaches}
Machine learning (ML) methods for anomaly detection relax many of the strong distributional assumptions required by classical statistical techniques, offering greater flexibility in modeling complex and high-dimensional data. These approaches learn patterns directly from data and can be categorized into several families:

\begin{itemize}
\item Clustering-based methods, such as \textit{k-means} and \textit{DBSCAN}, group data points based on similarity measures. Anomalies are identified as points that do not belong to any cluster (noise) or lie far from cluster centroids \citep{munz2007traffic}. While effective in identifying global outliers, these methods can struggle with local anomalies and depend heavily on distance metrics and parameter choices (e.g., number of clusters, neighborhood radius).
\item Dimensionality reduction techniques, including \textit{Principal Component Analysis (PCA)} and its variants, project data into a lower-dimensional subspace that captures most of the variance. Anomalies are then detected as points with large reconstruction errors or significant deviations from the principal subspace \citep{ahmed2016survey}. Such methods are useful for reducing noise and redundancy but may discard discriminative features relevant to subtle anomalies.

\item Isolation Forests (iForest) employ an ensemble of random trees that recursively partition the feature space. Anomalies are points that can be isolated with fewer splits, indicating they reside in sparse regions of the feature space \citep{liu2012isolation}. iForest is computationally efficient and scalable to large datasets but may underperform when anomalies form dense, localized groups.

\item One-Class Support Vector Machines (OC-SVM) learn a hypersphere or hyperplane that encloses the majority of normal data in a high-dimensional feature space, often using kernel functions to handle nonlinearity. Observations lying outside the learned boundary are flagged as anomalies \citep{li2003improving}. OC-SVM is powerful in low-to-moderate dimensions but becomes computationally expensive and sensitive to kernel choice in very high-dimensional settings.
\end{itemize}

Compared to purely statistical approaches, ML-based methods impose fewer distributional assumptions and can handle higher-dimensional data more effectively. However, they still face important limitations: many rely on large volumes of training data to generalize well, assume that normal instances dominate the training set, and often treat anomaly detection as a static task without leveraging iterative feedback or active label acquisition. Moreover, they typically do not exploit the \textit{geometric structure} of the feature space to refine decision boundaries dynamically—a gap that modern deep learning and active learning frameworks aim to address.
\subsection{Neural Evolution Methods}
Neural evolution methods represent an alternative, optimization-driven paradigm for anomaly detection that combines evolutionary algorithms with deep learning. Rather than relying solely on gradient-based learning, these approaches treat the design of anomaly detection models—including architecture, hyperparameters, and training policies—as an evolutionary search problem \citep{pietron2024ad}.

A typical neural evolution workflow begins with a population of candidate detectors (e.g., autoencoders, one-class classifiers, or generative models). Each candidate is evaluated using a fitness function tailored to anomaly detection, such as reconstruction error, area under the ROC curve, or F1-score on a validation set. Through iterative cycles of selection, crossover, and mutation, the population evolves toward configurations that maximize detection performance on the target dataset \citep{zeng2025evolutionary,haq2025transnas}.

The primary appeal of neural evolution lies in its ability to automate model design and reduce reliance on manual tuning, which is especially valuable in domains where anomalies are rare and data characteristics are poorly understood. Moreover, evolutionary search can explore diverse model families and architectures that might be overlooked by conventional training pipelines. However, these methods are computationally intensive, often requiring thousands of model evaluations, and may not scale efficiently to very high-dimensional or streaming data. They treat model configuration as a static optimization problem, separate from the sample-selection and boundary-refinement processes that define adaptive anomaly detection systems.
\subsection{Deep Learning Techniques}
Deep learning has revolutionized anomaly detection by enabling the modeling of complex, non-linear relationships in high-dimensional data. Unlike traditional machine learning methods, deep models can learn hierarchical feature representations directly from raw inputs, making them particularly effective for large-scale and heterogeneous datasets. Prominent deep learning approaches for anomaly detection include:

\begin{itemize}
\item Generative Adversarial Networks (GANs): GAN-based anomaly detectors train a generator–discriminator pair to learn the distribution of normal data. Anomalies are identified via high reconstruction error or low discriminator confidence when reconstructing or generating abnormal samples \citep{sabuhi2021applications}. While powerful, GANs are prone to training instability and mode collapse, and they often require careful architectural design and hyperparameter tuning.
\item Recurrent Neural Networks (RNNs): RNNs and their variants (e.g., LSTMs, GRUs) are designed to capture temporal dependencies in sequential data, making them well-suited for time-series anomaly detection \citep{radford2018network}. However, they can struggle with very long sequences and are less effective in non-sequential or tabular settings.

\item Graph Neural Networks (GNNs): GNN-based methods have emerged as a powerful paradigm for anomaly detection in relational or graph-structured data \citep{zhang2025graph, qiao2025deep, shao2025gnn}. By modeling entities as nodes and interactions as edges, GNNs aggregate neighborhood information to detect nodes or edges that deviate from learned structural patterns. These methods excel in social network analysis, fraud detection, and system provenance analysis but are inherently limited to graph-representable data.

\item Transformers: Transformer-based models leverage self-attention mechanisms to capture long-range dependencies and contextual relationships in sequential or multivariate data \citep{wang2025fusionformer, marino2025self,dilek2025overview}. They have shown strong performance in domains such as network log analysis and sensor monitoring, though their quadratic complexity and large parameter counts can be prohibitive for resource-constrained deployments.
\end{itemize}

While deep learning methods have demonstrated state-of-the-art performance on many large-scale anomaly detection benchmarks, they are not without limitations. They are often computationally expensive, require extensive hyperparameter tuning, and depend on large volumes of labeled or well-curated normal data for training. More critically, most deep anomaly detectors treat the learning process as static—once trained, they do not actively refine their decision boundaries based on newly acquired labels or feature-space geometry. This oversight is especially problematic in highly imbalanced, sparse, or evolving environments such as cybersecurity, where anomalies are rare and continuously adapting.

Our work directly addresses this gap by integrating similarity search into an active learning loop built upon a novel deep architecture. Unlike conventional deep anomaly detectors, our framework explicitly exploits the geometric structure of the feature space to guide sample selection, iteratively refine decision boundaries, and improve ranking accuracy—all while minimizing labeling effort. This synergy of deep representation learning, similarity-based reasoning, and active refinement represents a principled advance over existing deep learning approaches for anomaly detection.
\section{Materials and Methods}
\subsection{Motivations }
\label{sec:overview}
Detecting anomalies in real-world datasets is particularly challenging due to their rarity, diversity, and the imbalanced nature of the data. Traditional anomaly detection methods—whether statistical, machine learning, or deep learning—often fail to adequately capture the underlying structure of such datasets, especially in high-dimensional settings where noise and redundant features obscure meaningful patterns. To address these limitations, we propose a unified framework that combines a novel Sparse Dual Adversarial Attention-based AutoEncoder (SDA²E) with similarity-guided active learning. This integration allows the model to focus labeling effort on the most informative regions of the feature space, iteratively refining both the representation and the decision boundary.

The SDA²E architecture incorporates three key innovations:
\begin{itemize}
    \item Sparse Latent Representations: By enforcing sparsity in the latent representation using a Kullback-Leibler (KL) divergence, the model is forced to focus on critical features that distinguish anomalies from normal data.
    \item Attention Mechanisms: Feature-wise attention weights dynamically highlight relevant dimensions, improving robustness against irrelevant or noisy attributes. 
    \item Adversarial Learning: By employing a discriminator and a generator AutoEncoders trained in an adversarial way, the model enhances the quality of its latent space representations, ensuring that reconstructions of normal data are highly accurate while anomalies produce distinct errors.
\end{itemize}
To tackle the label scarcity inherent in imbalanced anomaly detection, we augment SDA²E with an active learning loop driven by similarity search in the feature space. This enables two primary objectives:
\begin{itemize}

    \item Better Decision Boundaries: Oracle feedback and similarity-based sample selection progressively refine the separation between normal and anomalous regions.
    \item Efficient Use of Labels: By querying only the most informative points, active learning minimizes annotation cost while maximizing model improvement.
\end{itemize}
Three complementary similarity search strategies are introduced to guide the active learning process: 
\begin{itemize}
    \item Strategy 1 (Normal-like): expands the training dataset with normal-like points identified via similarity, for retraining and improving the robustness of the SDA²E.
    \item Strategy 2 (Anomaly-like): identifies anomalous-like points through similarity search, prioritizing them for enhancing the ranking process.
    \item Strategy 3 (Hybrid): combines both approaches, ensuring a balanced improvement of both ranking and model retraining.
\end{itemize}

The synergy between both components: SDA²E's representation learning capability and similarity‑guided active learning creates a cohesive framework that dynamically adapts to highly imbalanced and high‑dimensional data regimes, such as those encountered in cybersecurity log analysis or Advanced Persistent Threat (APT) detection, where anomalies are sparse, stealthy, and costly to label. The proposed framework iteratively refines the decision-making process by incorporating oracle feedback and leveraging similarity search to enhance the model's robustness and accuracy. 
\subsection{Algorithmic Overview}
Algorithm~\ref{alg:globaloverview} outlines the core procedure of the proposed SDA²E-based active anomaly detection framework. The process begins with a cold‑start phase, where the SDA²E is pre‑trained on a small initially labeled set $\mX_l$. In each active learning iteration, reconstruction errors defined as:
\[
\mathcal{E}_{SDA²E}(\mathbf{x}_i) = \|\mathbf{x}_i - \hat{\mathbf{x}}_i\|_2^2,
\]
with $\hat{\mathbf{x}}_i$ being the reconstructed output from the AutoEncoder, are computed for all data points $\mathbf{x}_i$ $\in$ $\mX$, and instances are ranked in descending order of these errors. Points whose reconstruction error exceeds a dynamic threshold $\tau$—set here as the 80th percentile of the error distribution—are presented to the oracle for labeling. This means only the top 20\% of the ranked list will be examined for potential anomalousness. This percentile-based query strategy maintains a constant query budget per iteration while adapting to the shifting error distribution as the model retrains. Newly acquired labels are used to update and to trigger one of the three similarity‑search strategies described above. The model is then retrained on the expanded labeled set, and the ranked anomaly list is updated. This iterative loop continues until a stopping criterion (e.g., a maximum number of iterations or stabilization of the ranking) is met, yielding a final ranked list of anomalies and a refined SDA²E model.

The threshold $\tau$ is chosen as the 80th percentile to balance sensitivity and specificity; a higher percentile reduces false positives at the cost of possibly missing subtle anomalies, whereas a lower percentile increases recall but may introduce more false alarms. This data‑driven threshold adapts to the error distribution of each dataset, providing a flexible and robust mechanism for candidate selection.
\begin{algorithm}[h]
\caption{SDA²E Anomaly Detection Method Enhanced With Similarity-Based Active Learning}
\begin{algorithmic}[1]
\Require Dataset $\mat{X}$, initial labeled subset $\mX_l$, anomaly threshold $\tau$, similarity metric, SDA²E AutoEncoder model
\Ensure Ranked list of anomalies, improved SDA²E
\State Initialize and train SDA²E AutoEncoder model with $\mX_l$
\For{each iteration $t$}
    \State Compute reconstruction errors $\mathcal{E}_{SDA^2E}(\mathbf{x}_i)$ for $\mathbf{x}_i \in \mX$
    \State Rank all points by $\mathcal{E}_{SDA^2E}(\mathbf{x}_i)$ (descending order)
    \State Select top points with $\mathcal{E}_{SDA^2E}(\mathbf{x}_i) > \tau$ and query oracle for labels
    \State Update labeled dataset $\mX_l$ with oracle feedback
        \If{Strategy 1 applied}
        \State Identify and integrate normal-like points into $\mX_l$.
        \State Retrain SDA²E with updated $\mX_l$
    \EndIf
    \If{Strategy 2 applied}
        \State Identify anomalous-like points and prioritize them in the ranked list of anomalies.
    \EndIf

    \If{Strategy hybrid applied}
        \State Perform Steps 8 to 9 and 12
    \EndIf
    \State Evaluate the model.
\EndFor
\State \textbf{return} Final ranked list of anomalies
\end{algorithmic}
\label{alg:globaloverview}
\end{algorithm}
\subsection{Illustrative Example}
Consider a cybersecurity setting where each data point corresponds to a system process described by binary features (e.g., file modifications, network connections, login attempts). When the oracle labels a process as anomalous—for instance, due to unusual login patterns combined with unauthorized file access—Strategy 2 (anomaly-like) will retrieve other processes with similar suspicious feature combinations, prioritizing them for investigation. Conversely, when the oracle confirms a process as normal, Strategy 1 (normal-like) expands the training set with behaviorally similar benign processes, strengthening the model's ability to reconstruct normal patterns and thereby sharpening the decision boundary. \\
By alternately querying anomaly-like and normal-like candidates, the active learning loop rapidly enriches both ends of the detection spectrum, enabling the model to focus on emerging threat clusters while solidifying its understanding of legitimate behavior—all with minimal labeling effort.

\begin{figure}[h]
    \centering
    \includegraphics[width=\linewidth]{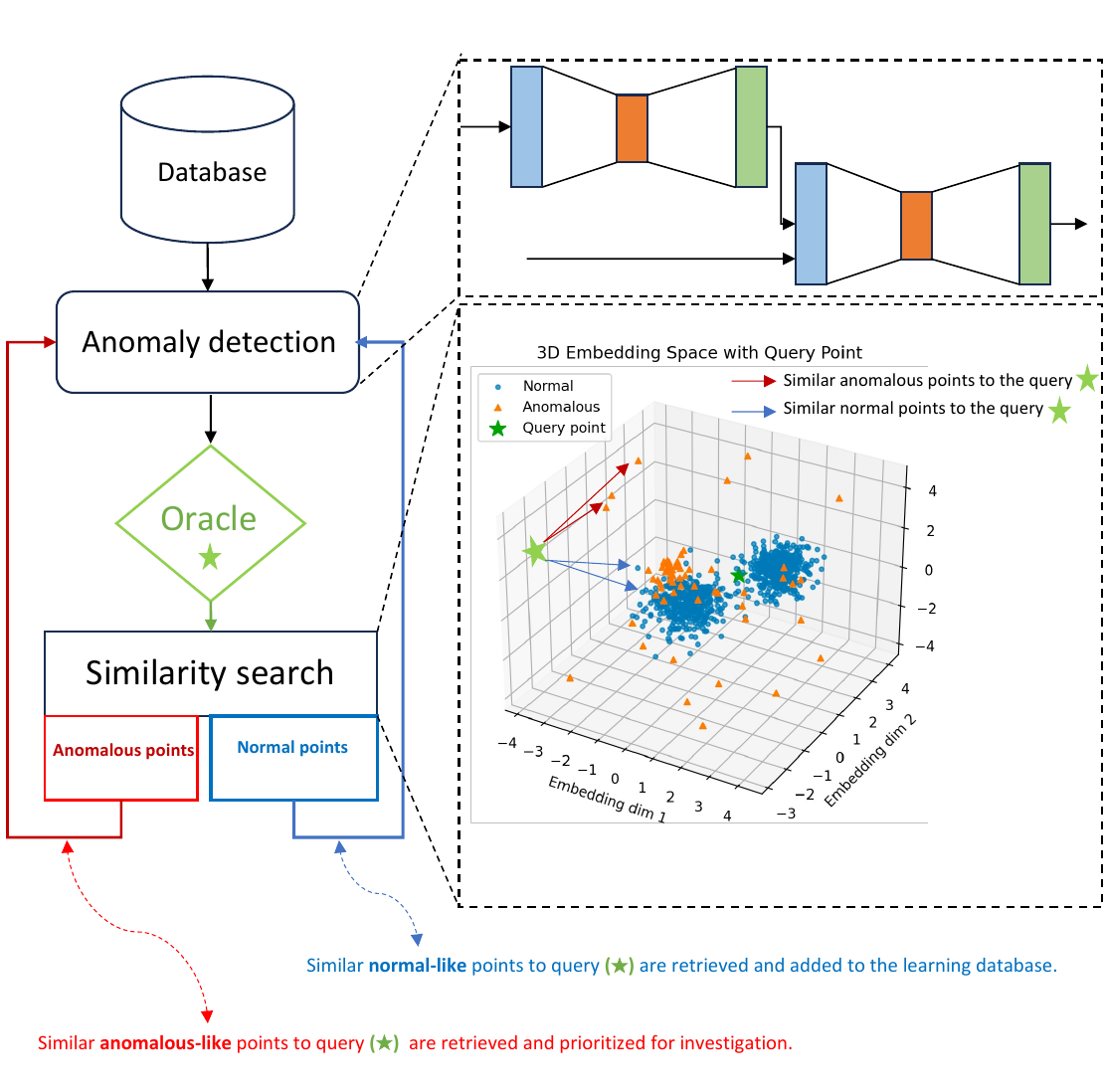}
    \caption{Overall architecture of the proposed SDA²E-based anomaly detection framework enhanced with similarity-guided active learning. The SDA²E encodes input data into a sparse latent representation, reconstructs it, and computes reconstruction errors for ranking. Points flagged as potentially anomalous (high reconstruction error) are sent to the oracle for labeling. Newly labeled anomalies are used to retrieve similar anomalous-like points via similarity search (Strategy 2) for priority analysis, while labeled normals are used to retrieve similar normal-like points (Strategy 1) which are incorporated into the training set, and the model is retrained iteratively to refine decision boundaries.}
    \label{fig:global-arch}
\end{figure}

Figure~\ref{fig:global-arch} summarizes the overall architecture, illustrating the interaction between the SDA²E, the oracle, and the similarity based sample‑selection strategies. By systematically integrating these components, the framework would effectively address the challenges of anomaly detection in imbalanced datasets, achieving robust performance while minimizing the labeling burden. \\
The following subsections detail the three core components: the SDA²E AutoEncoder (Section~\ref{sec:sda2e}), the similarity based active learning module (Section~\ref{sec:active-learning}), and the used similarity metric (Section~\ref{sec:similarity}).
\subsection{Sparse Dual Adversarial Attention-Based AutoEncoder}
\label{sec:sda2e}
\subsubsection{Architecture:}
The Sparse Dual Adversarial Attention-Based AutoEncoder (SDA²E) is a novel deep architecture designed to learn robust and discriminative representations for anomaly detection in high-dimensional and imbalanced datasets.
It integrates four core building blocks:
\begin{itemize}
    \item \emph{Dual AutoEncoding:} Two separate AutoEncoders—a generator and a discriminator—each with encoder--decoder structure.
    \item \emph{Adversarial Learning:} A reconstruction‑energy game between the two AutoEncoders, enforced by a margin.
    \item \emph{Attention Mechanism:} Feature‑wise gating that focuses on critical dimensions of the input.
    \item \emph{KL Divergence:} Sparsity constraints on the latent representations via Kullback--Leibler divergence.
\end{itemize}

SDA²E employs a generator AutoEncoder $G$ and a discriminator AutoEncoder $D$ trained adversarially. 
The generator $G$ learns to reconstruct normal data, capturing the dominant patterns in the feature space. 
The discriminator $D$ acts as an adversary by assigning \emph{low reconstruction energy} to real inputs and \emph{high reconstruction energy} to the generator's outputs, enforcing a margin $m$ between the two. 
This creates a min‑max game: $G$ tries to produce reconstructions that $D$ can also reconstruct well (low energy), while $D$ strives to separate real and generated samples by at least the margin $m$. 
The adversarial interplay sharpens the learned representations and improves anomaly‑detection robustness.

Sparsity constraints applied to the latent codes of both $G$ and $D$ force the networks to concentrate on salient features, discarding noise and irrelevant dimensions—a critical capability for high‑dimensional imbalanced data. 
An attention module further emphasizes informative features through a learnable gating mask. 
Together, the dual‑AE adversarial setup, sparsity regularization, and attention mechanism enable SDA²E to identify subtle anomalies while maintaining well‑separated decision boundaries, making it particularly suitable for cybersecurity, fraud detection, and industrial monitoring where anomalies are rare and diverse.
\subsubsection{Formal Definition:}
Given an input dataset $\mX = \{\mathbf{x}_1, \mathbf{x}_2, ..., \mathbf{x}_N\}$, where $\mathbf{x}_i \in \mathbb{R}^d$, an AutoEncoder is defined by an encoder $E$ and a decoder $\text{Dec}$, where the encoder $E$ maps $\mathbf{x}_i$ to a latent representation $\mathbf{z}_i$:
\[
\mathbf{z}_i = E(\mathbf{x}_i; \vtheta_E), \quad \mathbf{z}_i \in \mathbb{R}^k, \quad k < d
\]
The decoder $\text{Dec}$ reconstructs $\hat{\mathbf{x}}_i$ from $\mathbf{z}_i$:
\[
\hat{\mathbf{x}}_i = \text{Dec}(\mathbf{z}_i; \vtheta_{\text{Dec}}), \quad \hat{\mathbf{x}}_i \in \mathbb{R}^d
\]
The reconstruction loss for a mini-batch of size $B$ is the mean squared error:
\[
\mathcal{L}_{\text{recon}} = \frac{1}{B} \sum_{i=1}^B \| \mathbf{x}_i - \hat{\mathbf{x}}_i \|_2^2
\]

SDA²E employs a dual AutoEncoder configuration comprising a generator AutoEncoder $G$ and a discriminator AutoEncoder $D$ trained adversarially. The generator aims to reconstruct normal data faithfully, minimizing the reconstruction error for normal samples:
\[
\hat{\mathbf{x}} = G(\mathbf{x}; \vtheta_G), \quad 
\mathcal{L}_{\text{recon}}(G) = \frac{1}{B} \sum_{i=1}^B \|\mathbf{x}_i - \hat{\mathbf{x}}_i\|_2^2.
\]

The discriminator AutoEncoder $D$ assigns a reconstruction energy to any input, after attention modulation:
\[
E_D(\mathbf{x}) = \|\mathbf{x} - D(\mathbf{x}^*)\|_2^2,
\qquad 
E_D(\hat{\mathbf{x}}) = \|\hat{\mathbf{x}} - D(\hat{\mathbf{x}}^*)\|_2^2,
\]
where $\mathbf{x}^* = \mathbf{x} \odot \mA(\mathbf{x})$ and $\hat{\mathbf{x}}^* = \hat{\mathbf{x}} \odot \mA(\hat{\mathbf{x}})$ are the attention-modulated versions of the input and generator output, respectively. 
In practice we compute:
\[
E_{\text{real}} = \|\mathbf{x} - D(\mathbf{x}^*)\|_2^2, \qquad
E_{\text{fake}} = \|\hat{\mathbf{x}} - D(\hat{\mathbf{x}}^*)\|_2^2.
\]
It is trained to assign low reconstruction energy to real samples and higher reconstruction energy to generator reconstructions, enforced by a margin $m$. Its adversarial objective is defined as:
\[
\mathcal{L}_{\text{adv}}(D)
=
\mathbb{E}_{\mathbf{x}}\!\left[E_D(\mathbf{x})\right]
+
\mathbb{E}_{\mathbf{x}}\!\left[\max(0,\, m - E_D(\hat{\mathbf{x}}))\right].
\]

The margin parameter $m$ controls the minimum reconstruction-energy separation enforced by the discriminator between real and generated samples, and can be interpreted as an energy threshold below which an input is considered ``well reconstructed'' by $D$.\\
Conversely, the generator is encouraged to produce reconstructions that also receive low reconstruction energy under the discriminator:
\[
\mathcal{L}_{\text{adv}}(G)
=
\mathbb{E}_{\mathbf{x}}\!\left[E_D(\hat{\mathbf{x}})\right].
\]

This reconstruction-based adversarial interplay sharpens the latent representations and improves the model’s ability to separate normal from anomalous patterns. Simultaneously, sparsity constraints and attention gates force the network to concentrate on salient features, suppressing noise and irrelevant dimensions—a critical capability in high-dimensional, sparse domains such as cybersecurity logs.

%

\subsubsection{Sparsity in Latent Representations:}

To handle the challenges posed by high-dimensional data, sparsity constraints are imposed on the latent representations of both the generator and discriminator, denoted as $\mathbf{z}_G$ and $\mathbf{z}_D$, respectively. This ensures that the AutoEncoders focus on the most critical features, reducing the influence of noisy or irrelevant dimensions.\\
In the sparse-autoencoder formulation, each latent unit is treated as a Bernoulli random variable, where activation corresponds to ``success'' (1) and non-activation to ``failure'' (0). Sparsity is enforced by penalizing the mismatch between two key quantities:

\begin{itemize}
    \item $\rho$: The target activation probability (what we want to happen). For example, $\rho = 0.1$ means we aim for each neuron to be active approximately 10\% of the time.
    \item $\hat{\rho}_{G,j}$ and $\hat{\rho}_{D,j}$: The empirical activation frequencies (what actually happens) for the $j$-th latent units of the generator and discriminator, respectively.
\end{itemize}

The empirical activations are estimated over each mini-batch as:
\[
\hat{\rho}_{G,j} = \frac{1}{B} \sum_{i=1}^B \mathbb{I}[z_{G,j}^{(i)} > 0],
\qquad
\hat{\rho}_{D,j} = \frac{1}{B} \sum_{i=1}^B \mathbb{I}[z_{D,j}^{(i)} > 0],
\]
where $B$ is the batch size and $\mathbb{I}[\cdot]$ is the indicator function.

We then apply KL-based sparsity penalties that measure how ``surprised'' we would be if the true distribution had probability $\rho$ but we observed frequency $\hat{\rho}$:
\[
\mathcal{L}_{\text{sparse}}(G)
= \sum_{j=1}^k \text{KL}\big(\hat{\rho}_{G,j} \,\|\, \rho\big),
\quad
\mathcal{L}_{\text{sparse}}(D)
= \sum_{j=1}^k \text{KL}\big(\hat{\rho}_{D,j} \,\|\, \rho\big),
\]
where $k$ is the dimensionality of the latent representation (the number of latent units/neurons in the bottleneck layer). The KL divergence between two Bernoulli distributions is defined as:
\[
\text{KL}(\hat{\rho} \,\|\, \rho)
= \hat{\rho} \log \frac{\hat{\rho}}{\rho} 
+ (1-\hat{\rho})\log \frac{1-\hat{\rho}}{1-\rho}.
\]

During training, backpropagation adjusts the network weights to make all empirical frequencies $\hat{\rho}_{G,j}$ and $\hat{\rho}_{D,j}$ approach the target $\rho$. Initially, neurons activate randomly (some $\hat{\rho}$ too high, some too low), but the KL penalty gradually shapes the representation so that each neuron learns to activate only for specific, important patterns. We use a ReLU activation in the bottleneck layer, ensuring non-negative latent activations suitable for sparsity constraint.

This regularization encourages each latent unit to be active only a small fraction of the time (approximately $\rho$), promoting compact, selective, and interpretable latent representations. Sparsity provides several benefits:
\begin{itemize}

    \item it reduces the risk of overfitting by preventing both networks from relying on dense, entangled representations;
    \item it improves generalization by ensuring that only the most salient latent dimensions contribute significantly to reconstruction and discrimination; and
    \item it stabilizes adversarial training by discouraging overconfident discriminator behavior, a known factor in GAN training collapse.
\end{itemize}
The resulting sparse latent spaces make it easier to identify which features are most influential in distinguishing anomalous from normal behavior.

\subsubsection{Attention Mechanism:}
Attention mechanisms are integrated into both the generator and discriminator to dynamically focus on the most critical dimensions of the input data. For a given input $\mathbf{x}$, the attention mechanism computes a weight vector $\mA(\mathbf{x}) \in [0, 1]^d$ that highlights the importance of each feature. The modulated input is represented as:
\[
\mathbf{x}^* = \mathbf{x} \odot \mA(\mathbf{x}),
\]
where $\odot$ denotes element-wise multiplication. The computation of the attention weights $\mA(x)$ is performed using a fully connected neural network:
\[
\mA(\mathbf{x}) = \sigma(\mat{W}\mathbf{x} + \mathbf{b}),
\]
where $\mW \in \mathbb{R}^{d \times d}$ and $\mathbf{b} \in \mathbb{R}^d$ are the learnable parameters of the attention mechanism, and $\sigma$ is the sigmoid activation function that ensures weights lie in the range [0, 1].\\
The attention mechanism learns to prioritize features by assigning higher weights to the most relevant dimensions, effectively reducing the influence of noisy or irrelevant features. This adaptability is crucial in anomaly detection tasks, where only a subset of features might carry significant information about anomalies.\\
To enhance the learning of attention weights, an attention regularization term $\mathcal{L}_{\text{attn}}$ is introduced to encourage sparsity and interpretability of the attention weights:
\[
\mathcal{L}_{\text{attn}}
=
\lambda \lVert \mathbf{A}(\mathbf{x}) \rVert_1
=
\lambda \sum_{j=1}^{d} \left| A_j(\mathbf{x}) \right|,
\quad
A_j(\mathbf{x}) \in [0,1].
\]

where $\| \mA(\mathbf{x}) \|_1$ is the L1 norm of the attention weights, and $\lambda$ is a regularization coefficient.\\
The attention-modulated input $\mathbf{x}^*$ is fed into both the generator and discriminator to emphasize salient features during reconstruction and discrimination. By jointly learning sparse attention masks, SDA²E becomes more robust to high‑dimensional noise and better at isolating subtle anomalous patterns that would otherwise be obscured by dominant but irrelevant features.

\subsubsection{Total Loss with Dual Sparsity:}
The complete training objective for SDA²E is composed of a generator loss $\mathcal{L}_{\text{total}}(G)$ and a discriminator loss $\mathcal{L}_{\text{total}}(D)$.

The total loss function for the generator combines reconstruction, adversarial, sparsity, and attention regularization terms:
\[
\mathcal{L}_{\text{total}}(G)
=
\mathcal{L}_{\text{recon}}(G)
+
\alpha \mathcal{L}_{\text{adv}}(G)
+
\beta \mathcal{L}_{\text{sparse}}(G)
+
\gamma \mathcal{L}_{\text{attn}},
\]
where $\alpha,\beta,\gamma \geq 0$ control the contribution of each term.

The discriminator loss combines its reconstruction-energy adversarial objective with a sparsity penalty on its latent representation:
\[
\mathcal{L}_{\text{total}}(D)
=
\mathcal{L}_{\text{adv}}(D)
+
\delta \mathcal{L}_{\text{sparse}}(D),
\]
where $\delta \geq 0$ regulates the strength of the discriminator’s sparsity constraint. 

The discriminator $D$ distinguishes samples through their reconstruction energy; it learns to reconstruct real data with low energy while keeping the energy of generator outputs above the margin $m$. Hence, its loss contains the margin-based adversarial energy term $\mathcal{L}_{\text{adv}}(D)$ and a sparsity penalty $\mathcal{L}_{\text{sparse}}(D)$ to regularize its latent representation, without a separate classification head. 
\\    
The attention regularization $\mathcal{L}_{\text{attn}}$ is applied only to the generator because its primary role is to reconstruct normal data; focusing on salient features through attention directly improves reconstruction fidelity and anomaly separation. The discriminator, whose task is merely to distinguish real from reconstructed samples, does not require the same fine‑grained feature weighting, and omitting its attention loss simplifies training dynamics.\\
By enforcing sparsity in both the generator and discriminator, the model effectively learns robust and compact representations, focusing on critical features necessary for anomaly detection while mitigating the impact of noisy dimensions. During training, the two networks are optimized alternately: the generator minimizes $\mathcal{L}_{\text{total}}(G)$ to produce realistic reconstructions while maintaining sparse, interpretable representations; the discriminator minimizes $\mathcal{L}_{\text{total}}(D)$ to better distinguish real data from generator outputs. This dual‑sparsity design—enforced independently in both networks—ensures that the learned representations remain compact, focused on discriminative features, and robust to noise, which is critical for accurate anomaly detection in high‑dimensional, imbalanced datasets.

\subsubsection{Dual Adversarial Learning:}
SDA²E is trained using an alternating optimization scheme that separately updates the generator and discriminator.

\begin{itemize}
\item The generator minimizes $\mathcal{L}_{\text{total}}(G)$ to produce accurate reconstructions whose outputs also receive low reconstruction energy under the discriminator, while maintaining sparse latent codes and attention-guided feature selection.
\item The discriminator minimizes $\mathcal{L}_{\text{total}}(D)$ to assign low reconstruction energy to real samples and enforce a margin-based separation from generator reconstructions, with an auxiliary sparsity constraint on its latent representation.
\end{itemize}

This alternating min-max process continues until convergence, with the generator and discriminator progressively refining each other’s representations. The dual-sparsity regularization—applied independently to both networks—prevents overly dense or entangled latent spaces and stabilizes adversarial training. By integrating reconstruction-based adversarial learning with sparse, attention-guided feature extraction, SDA²E learns robust and interpretable representations that are highly effective for detecting rare and diverse anomalies in high-dimensional, imbalanced datasets.

%
\subsubsection{Training Sparse Dual Adversarial Attention-Based AutoEncoder}

Algorithm~\ref{alg:AE} presents the complete training procedure of SDA$^2$E.
The model is trained using mini-batch stochastic gradient descent with an alternating
optimization scheme for the generator, discriminator, and attention mechanism over
$M$ epochs.

At each iteration, attention is first applied to the input features, followed by
generator reconstruction, discriminator energy computation, and joint optimization
of all components.

\begin{enumerate}[label=\emph{Step (\arabic*)}, leftmargin=*, align=left]

\item \emph{Initialization:}
The trainable parameters of the generator, discriminator, and attention mechanism are
initialized as
$\boldsymbol{\theta}_G$, $\boldsymbol{\theta}_D$, and
$\boldsymbol{\theta}_A=\{\mathbf{W},\mathbf{b}\}$, respectively.
The attention parameters define a feature-wise gating function
$\mathbf{A}(\mathbf{x})=\sigma(\mathbf{W}\mathbf{x}+\mathbf{b})$.
All parameters are optimized jointly during training.

\item \emph{Attention computation:}
For each mini-batch $\{\mathbf{x}^{(i)}\}_{i=1}^{B}$, a feature-wise attention mask is
computed and applied:
\[
\mathbf{a}^{(i)}=\sigma(\mathbf{W}\mathbf{x}^{(i)}+\mathbf{b}), \qquad
\mathbf{x}^{*(i)}=\mathbf{x}^{(i)}\odot\mathbf{a}^{(i)}.
\]

\item \emph{Generator forward pass:}
The generator encodes the attention-modulated input and reconstructs it:
\[
\mathbf{z}_G^{(i)}=G_{\text{enc}}(\mathbf{x}^{*(i)}), \qquad
\hat{\mathbf{x}}^{(i)}=G_{\text{dec}}(\mathbf{z}_G^{(i)}).
\]

\item \emph{Attention on generated samples:}
The same attention mechanism is applied to the generator output:
\[
\hat{\mathbf{a}}^{(i)}=\sigma(\mathbf{W}\hat{\mathbf{x}}^{(i)}+\mathbf{b}), \qquad
\hat{\mathbf{x}}^{*(i)}=\hat{\mathbf{x}}^{(i)}\odot\hat{\mathbf{a}}^{(i)}.
\]

\item \emph{Discriminator forward pass and energy computation:}
The discriminator reconstructs both real and generated attention-modulated samples,
yielding reconstruction energies
\[
E_{\text{real}}^{(i)}=\|\mathbf{x}^{(i)}-D(\mathbf{x}^{*(i)})\|_2^2, \qquad
E_{\text{fake}}^{(i)}=\|\hat{\mathbf{x}}^{(i)}-D(\hat{\mathbf{x}}^{*(i)})\|_2^2.
\]

\item \emph{Loss evaluation:}
The following loss terms are computed:
\begin{align*}
\mathcal{L}_{\text{recon}}(G) &= \frac{1}{B}\sum_{i=1}^B
\|\mathbf{x}^{(i)}-\hat{\mathbf{x}}^{(i)}\|_2^2,\\
\mathcal{L}_{\text{adv}}(D) &= \frac{1}{B}\sum_{i=1}^B
\Big(E_{\text{real}}^{(i)}+\max(0,m-E_{\text{fake}}^{(i)})\Big),\\
\mathcal{L}_{\text{adv}}(G) &= \frac{1}{B}\sum_{i=1}^B E_{\text{fake}}^{(i)},\\
\mathcal{L}_{\text{sparse}}(G),\;\mathcal{L}_{\text{sparse}}(D) &\text{ via KL divergence},\\
\mathcal{L}_{\text{attn}} &= \frac{\lambda}{B}\sum_{i=1}^B\|\mathbf{a}^{(i)}\|_1.
\end{align*}

The total objectives are then formed as:
\begin{align*}
\mathcal{L}_{\text{total}}(G) &=
\mathcal{L}_{\text{recon}}(G)
+\alpha\mathcal{L}_{\text{adv}}(G)
+\beta\mathcal{L}_{\text{sparse}}(G)
+\gamma\mathcal{L}_{\text{attn}},\\
\mathcal{L}_{\text{total}}(D) &=
\mathcal{L}_{\text{adv}}(D)
+\delta\mathcal{L}_{\text{sparse}}(D).
\end{align*}

\item \emph{Parameter updates:}
The discriminator and generator are updated alternately, while the attention
parameters are updated jointly with the generator:
\begin{align*}
\boldsymbol{\theta}_D &\leftarrow
\boldsymbol{\theta}_D-\eta_D\nabla_{\boldsymbol{\theta}_D}\mathcal{L}_{\text{total}}(D),\\
\boldsymbol{\theta}_G &\leftarrow
\boldsymbol{\theta}_G-\eta_G\nabla_{\boldsymbol{\theta}_G}\mathcal{L}_{\text{total}}(G),\\
\boldsymbol{\theta}_A &\leftarrow
\boldsymbol{\theta}_A-\eta_A\nabla_{\boldsymbol{\theta}_A}\mathcal{L}_{\text{total}}(G).
\end{align*}

\end{enumerate}

After $M$ epochs, the trained parameters
$\{\boldsymbol{\theta}_G,\boldsymbol{\theta}_D,\boldsymbol{\theta}_A\}$ are obtained.
During inference, the anomaly score of a sample is computed as:
\[
\text{Ascore}(\mathbf{x})=\|\mathbf{x}-G(\mathbf{x}^*)\|_2^2,
\]
and samples whose scores exceed a threshold $\tau$ are flagged as anomalies (e.g., the 80th percentile of validation errors).\\
\begin{algorithm}[h!]
\caption{Training Sparse Dual Adversarial Attention-Based AutoEncoder (SDA$^2$E)}
\label{alg:AE}
\begin{algorithmic}[1]
\Require Dataset $\mX$, epochs $M$, batch size $B$, margin $m$, sparsity target $\rho$, 
attention weight $\lambda$, hyperparameters $\alpha,\beta,\gamma,\delta$, 
learning rates $\eta_G,\eta_D,\eta_A$
\Ensure Trained parameters $\vtheta_G, \vtheta_D, \vtheta_A$

\State Initialize generator parameters $\vtheta_G$
\State Initialize discriminator parameters $\vtheta_D$
\State Initialize attention parameters $\vtheta_A=\{\mW,\mathbf{b}\}$

\For{epoch $=1$ to $M$}
  \For{each mini-batch $\mB=\{\mathbf{x}^{(i)}\}_{i=1}^{B} \subset \mX$}

    \State \textbf{Attention on real inputs:}
    \For{$i=1$ to $B$}
      \State $\mathbf{a}^{(i)} = \sigma(\mW\mathbf{x}^{(i)} + \mathbf{b})$
      \State $\mathbf{x}^{\star(i)} = \mathbf{x}^{(i)} \odot \mathbf{a}^{(i)}$
    \EndFor

    \State \textbf{Generator forward pass:}
    \For{$i=1$ to $B$}
      \State $\mathbf{z}_G^{(i)} = G_{\text{enc}}(\mathbf{x}^{\star(i)};\vtheta_G)$
      \State $\hat{\mathbf{x}}^{(i)} = G_{\text{dec}}(\mathbf{z}_G^{(i)};\vtheta_G)$
    \EndFor

    \State \textbf{Attention on generated samples (for discriminator input):}
    \For{$i=1$ to $B$}
      \State $\hat{\mathbf{a}}^{(i)} = \sigma(\mW\hat{\mathbf{x}}^{(i)} + \mathbf{b})$
      \State $\hat{\mathbf{x}}^{\star(i)} = \hat{\mathbf{x}}^{(i)} \odot \hat{\mathbf{a}}^{(i)}$
    \EndFor

    \State \textbf{Discriminator reconstructions (energy model):}
    \For{$i=1$ to $B$}
      \State $\tilde{\mathbf{x}}^{(i)} = D(\mathbf{x}^{\star(i)};\vtheta_D)$
      \State $\tilde{\mathbf{x}}_{\text{fake}}^{(i)} = D(\hat{\mathbf{x}}^{\star(i)};\vtheta_D)$
      \State $E_{\text{real}}^{(i)} = \|\mathbf{x}^{(i)} - \tilde{\mathbf{x}}^{(i)}\|_2^2$
      \State $E_{\text{fake}}^{(i)} = \|\hat{\mathbf{x}}^{(i)} - \tilde{\mathbf{x}}_{\text{fake}}^{(i)}\|_2^2$
    \EndFor

    \State \textbf{Loss computation:}
    \State $\mathcal{L}_{\text{recon}}(G) = \frac{1}{B}\sum_{i=1}^{B}\|\mathbf{x}^{(i)}-\hat{\mathbf{x}}^{(i)}\|_2^2$
    \State $\mathcal{L}_{\text{adv}}(D) = \frac{1}{B}\sum_{i=1}^{B}\Big(E_{\text{real}}^{(i)} + \max(0,\, m-E_{\text{fake}}^{(i)})\Big)$
    \State $\mathcal{L}_{\text{adv}}(G) = \frac{1}{B}\sum_{i=1}^{B} E_{\text{fake}}^{(i)}$
    \State Compute $\mathcal{L}_{\text{sparse}}(G)$ and $\mathcal{L}_{\text{sparse}}(D)$ via KL divergence (target $\rho$)
    \State $\mathcal{L}_{\text{attn}} = \frac{\lambda}{B}\sum_{i=1}^{B}\|\mathbf{a}^{(i)}\|_1$

    \State \textbf{Total objectives:}
    \State $\mathcal{L}_{\text{total}}(G)=\mathcal{L}_{\text{recon}}(G)+\alpha\mathcal{L}_{\text{adv}}(G)+\beta\mathcal{L}_{\text{sparse}}(G)+\gamma\mathcal{L}_{\text{attn}}$
    \State $\mathcal{L}_{\text{total}}(D)=\mathcal{L}_{\text{adv}}(D)+\delta\mathcal{L}_{\text{sparse}}(D)$

    \State \textbf{Parameter updates (alternating optimization):}
    \State $\vtheta_D \leftarrow \vtheta_D - \eta_D \nabla_{\vtheta_D}\mathcal{L}_{\text{total}}(D)$
    \State $\vtheta_G \leftarrow \vtheta_G - \eta_G \nabla_{\vtheta_G}\mathcal{L}_{\text{total}}(G)$
    \State $\vtheta_A \leftarrow \vtheta_A - \eta_A \nabla_{\vtheta_A}\mathcal{L}_{\text{total}}(G)$ 
    \Comment{Attention is learned jointly via $\mathcal{L}_{\text{total}}(G)$}

  \EndFor
\EndFor

\State \Return $\vtheta_G, \vtheta_D, \vtheta_A$
\end{algorithmic}
\end{algorithm}

%
The anomaly detection decision rule with an AutoEncoder can be written as:
\begin{equation*}
\text{Anomaly}(\mathbf{x}) = 
\begin{cases} 
1 & \text{if } error(\mathbf{x}) > \tau \\
0 & \text{otherwise}
\end{cases}
\end{equation*}
\subsubsection{Complexity Remarks:}
Let $n = |\mathbf{X}|$, $d$ the input dimension, $k$ the latent dimension, and $B$ the batch size. One forward‑backward pass through SDA²E costs $O(B\cdot(dk + k^2))$ for the generator and $O(B\cdot dk)$ for the discriminator. The attention module adds $O(Bd^2)$. In practice, $k \ll d$ and $B$ is fixed, making the per‑epoch complexity linear in $n$, which is scalable to large high‑dimensional datasets.
\subsubsection{Analytical Perspective on Sparsity–Attention Interaction:}
The sparsity constraint enforced through the Kullback--Leibler (KL) divergence term encourages the latent distribution $q(\mathbf{z})$ to remain close to a low-entropy prior $p(\mathbf{z})$, penalizing redundant or weakly activated neurons. This regularization drives the encoder to allocate probability mass only to salient latent dimensions. When trained jointly with the attention module, this effect becomes complementary: the attention weights act as a differentiable gate that amplifies high-utility latent responses while further suppressing uninformative ones. Formally, the joint optimization can be interpreted as a constrained minimization problem in which the reconstruction loss is minimized subject to an information-sparsity penalty. Under adversarial learning, the discriminator amplifies this effect by forcing the generator to match the distribution of real embeddings, leading to an equilibrium in which compact yet expressive latent codes are preferred. Consequently, the KL term controls global sparsity, whereas attention provides localized feature selection, together yielding stable and interpretable representations that improve anomaly separability.
\subsection{Active Learning with Similarity Search}
\label{sec:active-learning}
\subsubsection{Oracle Feedback and Labeling:}

Active learning plays a crucial role in the proposed anomaly detection framework by iteratively identifying and selecting uncertain data points for labeling. This approach significantly reduces dependence on extensive human annotation efforts, which is especially important in resource‑limited scenarios. In anomaly detection, active learning is particularly effective for highly imbalanced datasets where anomalies constitute only a small fraction of the data and labeled examples are scarce. By focusing on the most informative points, active learning accelerates model convergence, sharpens decision boundaries, and lowers annotation costs. Commonly employed strategies include uncertainty sampling, which targets points where the model exhibits the highest uncertainty (often near decision boundaries); diversity sampling, which selects a varied subset of points to capture broader dataset patterns; and query‑by‑committee, which identifies points where multiple models disagree significantly, highlighting regions of the feature space that would benefit most from oracle feedback.

In our framework, uncertainty sampling is implemented via a ranked list of candidates based on reconstruction errors from SDA²E. Points with errors exceeding threshold $\tau$ are presented to the oracle for binary labeling (normal/anomalous). This addresses two key challenges: (i) optimal use of limited labeled data by prioritizing informative samples, and (ii) iterative expansion of the training set through oracle feedback.
\subsubsection{Similarity‑Guided Sample Selection}
To further enhance the active learning loop, we introduce three similarity‑search strategies that exploit geometric relationships in the feature space: Strategy 1 (Normal‑like) retrieves unlabeled points similar to oracle‑labeled normals and adds them to the training set to strengthen the AutoEncoder’s reconstruction of normal patterns; Strategy 2 (Anomaly‑like) retrieves points similar to known anomalies and prioritizes them in the ranked list to boost detection precision; and Strategy 3 (Hybrid) combines both approaches, simultaneously refining the model and improving ranking accuracy. These strategies transform passive uncertainty sampling into an active, geometry‑aware refinement process. The following subsections detail their algorithmic implementation (Algorithms 3‑5) and integration into the overall framework.
\subsubsection{Similarity Search for Normal-Like Points: 
Dataset Augmentation and Model Refining}
Strategy 1 enhances the model's representation of normal behavior by identifying unlabeled points that are similar to oracle confirmed normal examples. For each labeled normal point \( \mathbf{n} \in \mathcal{A}_{\text{normal}} \), we compute its similarity to all unlabeled points \( \mathbf{x} \in \mathcal{X}_u \) using a chosen similarity function $S(\cdot,\cdot)$. Points with similarity \( S(\mathbf{x}, \mathbf{n}) \geq \varrho \) (where $\varrho$ is a predefined threshold) are considered normal-like and are added to the training set. This iterative augmentation strengthens the AutoEncoder's reconstruction of normal patterns, thereby sharpening its sensitivity to anomalous deviations.
This strategy is summarized in Algorithm \ref{alg:strategy1}.\\
\begin{algorithm}[h!]
\caption{Strategy 1: Identify Normal‑Like Points for Model Refinement}
\begin{algorithmic}[1]
\Require Model, unlabeled data $\mathcal{X}u$, labeled normals $\mathcal{A}_{\text{normal}}$, threshold $\varrho$, similarity function $S$.
\Ensure Updated training set $\mathcal{X}_l'$
\State $\mathcal{X}_{\text{normal}} \gets \varnothing$
\For{each $\mathbf{n} \in \mathcal{A}_{\text{normal}}$}
\State $\mathcal{X}_n \gets [,]$
\For{each $\mathbf{x} \in \mathcal{X}u$}
\State $s \gets S(\mathbf{x}, \mathbf{n})$
\If{$s \geq \varrho$}
\State $\mathcal{X}_n.\text{append}(\mathbf{x})$
\EndIf
\EndFor
\State $\mathcal{X}_{\text{normal}} \gets \mathcal{X}_{\text{normal}} \cup \mathcal{X}_n$
\EndFor
\State $\mathcal{X}_l' \gets \mathcal{X}_l \cup \mathcal{X}_{\text{normal}}$
\State \textbf{return} $\mathcal{X}_l'$
\end{algorithmic}
\label{alg:strategy1}
\end{algorithm}

\subsubsection{Similarity Search for Anomalous-Like Points: Identification and Ranking Prioritization}
This strategy enhances the ranking performance of our anomaly detector by prioritizing unlabeled points that resemble known anomalies. For each oracle-confirmed anomaly \( \mathbf{a} \in A_{\text{anomaly}} \), we identify unlabeled points with high similarity (exceeding threshold \(\xi\)). These anomaly-like points, together with the oracle-labeled anomalies themselves, are placed at the top of the ranked list, ensuring they are investigated in subsequent active learning iterations.

Let \( \mathcal{X}_u \) be the set of unlabeled data points and \( \mathcal{A}_{\text{anomaly}} \) the set of anomalies confirmed by the oracle. For each labeled anomaly \( \mathbf{a} \in \mathcal{A}_{\text{anomaly}} \), we compute the similarity \( S(\mathbf{x}, \mathbf{a}) \) to every unlabeled point \( \mathbf{x} \in \mathcal{X}_u \). Points whose similarity exceeds a predefined threshold \(\xi\) are collected as anomaly-like candidates:

\[
\mathcal{X}_{\mathbf{a}} = \{ \mathbf{x} \in \mathcal{X}_u \mid S(\mathbf{x}, \mathbf{a}) \geq \xi \}.
\]

All such points across different \( \mathbf{a} \) are merged into a single set:

\[
\mathcal{X}_{\text{sim}} = \bigcup_{\mathbf{a} \in \mathcal{A}_{\text{anomaly}}} \mathcal{X}_{\mathbf{a}}.
\]

For each candidate \( \mathbf{x} \in \mathcal{X}_{\text{sim}} \), we retrieve its reconstruction error from the base anomaly detector. The final prioritized list is constructed by combining:
\begin{itemize}

    \item The oracle-labeled anomalies \( A_{\text{anomaly}} \)
    \item The anomaly-like candidates \( \mathcal{X}_{\text{sim}} \), sorted in descending order of reconstruction error
\end{itemize}

This combined set \( \mathcal{A}_{\text{anomaly}} \cup \mathcal{X}_{\text{sim}} \) is placed at the top of the main ranked anomaly list, ensuring that both confirmed anomalies and points similar to them are prioritized for investigation.

\begin{algorithm}[h!]
\caption{Strategy 2: Anomaly-Proximity Search and Ranking}
\begin{algorithmic}[1]
\Require Model, unlabeled data $\mathcal{X}_u $, labeled anomalies $\mathcal{A}_{\text{anomaly}} $, similarity function $S$, 
         threshold $\xi$, reconstruction error function $E$
\Ensure Prioritized list of points for investigation $ \mathcal{R}_{\text{priority}} $
\State $ \mathcal{X}_{\text{sim}} \gets \{\} $   
\For{each anomaly $ \mathbf{a} \in \mathcal{A}_{\text{anomaly}} $}
    \For{each  $ \mathbf{x} \in \mathcal{X}_u $}
        \If{$ S(\mathbf{x}, \mathbf{a}) \geq \xi $}
            \State $ \mathcal{X}_{\text{sim}} \gets \mathcal{X}_{\text{sim}} \cup \{\mathbf{x}\} $
        \EndIf
    \EndFor
\EndFor
\State For each $ \mathbf{x} \in \mathcal{X}_{\text{sim}} $, retrieve $ E(\mathbf{x}) $ from anomaly detector
\State Sort $ \mathcal{X}_{\text{sim}} $ in descending order of $ E(\mathbf{x}) $
\State $ \mathcal{R}_{\text{priority}} \gets \mathcal{A}_{\text{anomaly}} \cup \text{sorted } \mathcal{X}_{\text{sim}} $
\State \Return $ \mathcal{R}_{\text{priority}} $
\end{algorithmic}
\label{alg:strategy2}
\end{algorithm}

Algorithm~\ref{alg:strategy2} outlines the anomalous-like active learning strategy. The output \( \mathcal{R}_{\text{priority}} \) contains both the oracle-labeled anomalies and points similar to them, with the similar points ranked by their original anomaly scores. This list is placed at the beginning of the main anomaly ranking to ensure priority investigation of confirmed anomalies and points that closely resemble them.

\subsubsection{Hybrid Similarity Search: Combined Approach}

The hybrid similarity search approach integrates both normal-like and anomaly-like search strategies into a unified active learning framework. This combined methodology addresses the dual requirements of effective anomaly detection: robust modeling of normal patterns and targeted investigation of anomalous regions. By implementing Strategy 1 (normal augmentation) and Strategy 2 (anomaly prioritization) in parallel, the hybrid approach achieves synergistic benefits that enhance overall detection performance.

The implementation operates through two concurrent processes during each active learning iteration. The normal-like component identifies and incorporates points similar to confirmed normal examples into the training data, refining the model's baseline representation. Simultaneously, the anomaly-like component identifies points resembling confirmed anomalies and prioritizes them for investigation. These two processes create complementary reinforcement: improved normal modeling increases sensitivity to deviations, while targeted anomaly investigation efficiently discovers new anomalies.

This integrated approach dynamically adapts to the evolving understanding of both normal and anomalous patterns throughout the active learning cycle, optimizing both model refinement and anomaly discovery efficiency.

\subsubsection{Threshold Selection and Implementation Details}

The effectiveness of both strategies depends critically on appropriate threshold selection for similarity-based point identification. Rather than using fixed absolute thresholds, we employ a percentile-based approach that adapts to dataset-specific similarity distributions.

Both $\varrho$ (for normal augmentation) and $\xi$ (for anomaly prioritization) are set to the 80th percentile of their respective similarity score distributions (. This uniform threshold provides a consistent selection criterion across strategies while maintaining adaptability to different data characteristics. For each labeled normal point $\mathbf{n} \in A_{\text{normal}}$, we compute $\varrho$ as the 80th percentile of $\{S(\mathbf{x}, \mathbf{n}): \mathbf{x} \in \mathcal{X}_u\}$, selecting points with $S(\mathbf{x}, \mathbf{n}) \geq \varrho$ as normal-like candidates. Similarly, for each labeled anomaly $\mathbf{a} \in A_{\text{anomaly}}$, we compute $\xi$ as the 80th percentile of $\{S(\mathbf{x}, \mathbf{a}): \mathbf{x} \in \mathcal{X}_u\}$, identifying points with $S(\mathbf{x}, \mathbf{a}) \geq \xi$ as anomaly-like candidates.

This percentile-based methodology offers several advantages: adaptability to varying similarity distributions across datasets, consistency through unified thresholding logic, robustness to outliers, and simplified hyperparameter tuning with a single percentile parameter governing both strategies.

In practice, thresholds are computed dynamically during each active learning iteration. After receiving new oracle labels, we calculate pairwise similarities between labeled points and all unlabeled points. The 80th percentile of these similarity scores for each labeled point serves as its individual threshold for candidate selection in the subsequent iteration. This dynamic computation ensures that the selection process remains responsive to the evolving composition of labeled data throughout the active learning cycle.

\subsection{Similarity Measures and Distance Metrics}
\label{sec:similarity}
\subsubsection{Definition:}
Here, we describe the type of data used in our anomaly detection framework and the similarity metric applied to quantify relationships between data points. The metric forms the basis for effective anomaly ranking and prioritization.

We consider a binary dataset $\mX \in \{0,1\}^{n \times m}$, where:
\begin{itemize}
    \item $n$ is the number of data points (rows), and $m$ is the number of features (columns).
    \item Each entry $x_{i,j} \in \{0, 1\}$ indicates whether the $j$-th feature is active ($1$) or inactive ($0$) for the $i$-th data point. For instance, in a cybersecurity application, $i$ could represent a specific process recorded in the system logs, while $j$ could correspond to individual features such as access events, commands executed, or file modifications observed during that process.
\end{itemize}
For instance, consider the following dataset with $n=3$ data points and $m=4$ features:
\[
\mX = \begin{bmatrix}  
1 & 0 & 1 & 1&0 \\
0 & 1 & 1 & 1 \\  
1 & 1 & 1 & 0 & 0
\end{bmatrix}.
\]

The dataset represents three data points, where:
\begin{itemize}
    \item Data point $1$ has features $1$, $3$  and $4$ active.
    \item Data point $2$ has features $2$, $3$, and $4$ active.
    \item Data point $3$ has features $1$, $2$ and $3$ active.
\end{itemize}
Our goal is to compute similarity or distance metrics between rows (data points) to identify relationships and rank points for anomaly detection.
 \subsubsection{Normalized Matching 1s Measure:}
To compute the similarity between data points, we introduce here the Normalized Matching 1s  measure ($SIM_{NM1}$) to quantify the degree of similarity between two binary feature vectors by focusing exclusively on the positions where both vectors have exact active features (i.e., positions where both are `1`). Unlike more general similarity measures, Normalized Matching 1s excludes matches of `0` values and emphasizes the overlap of important active features.\\
Formally, for two binary vectors \(\mA\) and \(\mB\) of equal length \(n\):
\[
SIM_{NM1}(\mA,\mB) = \frac{|\mA \cap \mB|}{\max(|\mA|, |\mB|)}
\]
where:
\begin{itemize}
    \item \(|\mA \cap \mB|\): The number of positions where both \(\mA[i] = 1\) and \(\mB[i] = 1\).
    \item \(|\mA|\): The total number of `1`s in \(\mA\).
    \item \(|\mB|\): The total number of `1`s in \(\mB\).
\    \item \(\max(|\mA|, |\mB|)\): The maximum possible number of `1`s to ensure normalization.
\end{itemize}

This measure ensures the result lies in the range \([0, 1]\), with \(1\) representing perfect alignment of `1`s and \(0\) indicating no alignment. It is also particularly suited for applications involving binary feature vectors where active features (`1`s) represent critical attributes of a dataset. Examples include binary representations of categorical data, activation states, or binary presence-absence vectors. For instance, by considering the previous dataset \( \mX \), the $SIM_{NM1}$ similarity between Row1 and Row3 is computed as follows:
\[
SIM_{NM1}(Row1,Row3) = \frac{|Row1 \cap Row3|}{\max(|Row1|, |Row3|)} = \frac{2}{3} = 0.66
\] hence this similarity directly identifies the overlap of active features (positions with 1) and scales it by the maximum possible number of features.
\subsubsection{Computational Complexity:} In terms of computational complexity, $SIM_{NM1}$  is highly efficient, with a per-comparison cost linear in the number of features, $O(d)$. This is comparable to other metrics like Hamming and Jaccard, making it perfectly suitable for large-scale similarity search within the active learning loop, especially when the number of active features per vector ($|\mA|$, $|\mB|$) is precomputed.
\subsubsection{Discussion:}
In the context of anomaly detection, $SIM_{NM1}$ measure has several key advantages. Primarily, it focuses on active features; unlike measures such as Hamming or Cosine similarity that consider `0` values, this measure isolates the agreement between meaningful `1`s, which are often the primary interest in binary feature datasets. It also considers normalization by active features, by normalizing with \(\max(|\mA|, |\mB|)\), and adjusts for differences in sparsity between the compared vectors, making it robust to imbalanced densities of `1`s. It is also useful for the aim of interpretability, since the range \([0, 1]\) simplifies interpretation, where higher values indicate stronger overlap in active features.\\
$SIM_{NM1}$ similarity measure has also distinct properties compared to widely-used measures.
While Hamming similarity measures overall agreement (both `1`s and `0`s), it may overstate similarity by including non-active feature matches (`0`). $SIM_{NM1}$ focuses solely on shared `1`s, providing a more precise measure of overlap for active features. The Jaccard measure, defined as:
    $\frac{|\mA \cap \mB|}{|\mA \cup \mB|}$ penalizes both mismatched `1`s and additional `1`s in the union, whereas $SIM_{NM1}$ normalizes only by the maximum active features, making it more tolerant to feature sparsity. Dice Similarity, defined as: $\frac{2 |\mA \cap \mB|}{|\mA| + |\mB|}$ assigns double weight to intersections. While this may benefit highly sparse datasets, $SIM_{NM1}$ provides more balanced treatment of overlaps and non-overlapping regions. The Cosine similarity, defined as: $\frac{\mA \cdot \mB}{\|\mA\| \cdot \|\mB\|}$ focuses on vector direction rather than feature overlap. It is sensitive to vector magnitudes and may be less effective for purely binary data, unlike $SIM_{NM1}$, which prioritizes active feature overlap.\\%
The Normalized Matching 1s measure provides a focused, robust, and interpretable approach to evaluating similarity in binary feature vectors, particularly for datasets where active features (`1`s) are of primary importance. Its ability to highlight meaningful overlaps without being skewed by inactive features (`0`s) makes it an excellent choice for applications in anomaly detection, clustering, and feature analysis.

\section{Experimental settings and data}
\subsection{Evaluation Protocol:}
In this section, we present the datasets and benchmarking protocol used to evaluate the proposed framework, SDA²E, against various existing anomaly detection methods. Although the evaluation is performed across multiple datasets, our primary focus is the detection of Advanced Persistent Threats (APTs). APT cyber attacks are characterized by their prolonged, targeted nature, often aiming to steal sensitive data or disrupt critical systems. These attacks are notably stealthy, persistent, and adept at evading conventional security measures. Detecting APTs is a particularly challenging task due to the subtlety and rarity of the signals they generate, positioning them as archetypal, imbalanced, and rare outliers in cybersecurity contexts. Focusing on APTs highlights the relevance and robustness of our proposed framework in addressing real-world security threats.

Our method is evaluated against thirteen anomaly detection methods, each based on distinct underlying principles, ensuring a comprehensive comparative analysis across diverse detection approaches. This comparison is essential to assess the robustness and effectiveness of our framework across different anomaly detection paradigms, including statistical methods, rule-based approaches, density-based techniques, and machine learning-based models. By including methods with varied detection mechanisms, we ensure that our evaluation covers a broad spectrum of real-world anomaly detection scenarios, demonstrating the adaptability and generalization capability of our proposed approach.
The existing anomaly detection methods selected for comparison are as follows:
\begin{itemize}
\item ATDAD (Adversarial Tabular Deep Anomaly Detection): It is a recent one-class framework tailored for tabular data. It combines an autoencoder-style generator with an adversarial discriminator to learn a compact representation of normal behavior. The generator is trained to reconstruct normal samples, while the discriminator distinguishes real inputs from reconstructed ones, encouraging the model to tightly capture the normal data manifold. At test time, anomalies are identified using a score that integrates reconstruction error and discriminator confidence \citep{ATDADYANG2023103449}.
\item TTVAE (Transformer-based Tabular Variational AutoEncoder): It is a variational anomaly detection framework that uses two complementary teacher networks to guide a student Variational AutoEncoder (VAE) toward more robust latent representations. One teacher captures global normality, while the other enforces local structural consistency. Anomalies are detected via reconstruction and latent divergence scores, making TTVAD effective on tabular and structured data \citep{TTVAEWANG2025104292}.

\item AnoGAN (Anomaly Generative Adversarial Network):  In this framework, a tabular generator--discriminator pair is trained exclusively on normal processes to learn the manifold of benign system behavior. The generator maps latent variables to synthetic samples that mimic normal activity, while the discriminator distinguishes real normal samples from generated ones. Following the AnoGAN procedure, anomaly scoring is performed via per-sample latent optimization: for a given test process, a latent vector is iteratively optimized so that the generator output best reconstructs the input. The resulting reconstruction error, computed after latent refinement, serves as the anomaly score. Samples that cannot be well approximated by the generator---indicating deviation from the learned normal manifold---receive higher anomaly scores \citep{reddy2024anogan}.

    \item VF-ARM (Valid Frequent Itemset Anomaly Rule Mining): This approach focuses on mining frequent association rules in a dataset and considers deviations from these patterns as indicative of anomalies. By identifying frequent co-occurrences of itemsets in the data, this method assumes that anomalies often arise from rare combinations of attributes that violates the frequent rules extracted from the dataset. VF-ARM is particularly beneficial in scenarios where patterns in the data can be clearly defined (e.g., web traffic or transaction logs), and detecting deviations from these patterns could uncover anomalous events or behaviors \citep{Benabderrahmane21}.

    \item VR-ARM (Valid Rare Itemset Anomaly Rule Mining): It is an extension of VF-ARM, but with a focus on rare association rules in the data. It mines infrequent combinations of itemsets that might signal anomalous activities or attacks. While VF-ARM is geared towards deviations from frequent rules, VR-ARM is designed to highlight anomalies by identifying rules that occur infrequently but still have strong significance. This method is particularly suited to APT detection, as rare attack behaviors can be difficult to identify but can be equally indicative of underlying cyber threats \citep{Benabderrahmane21}.
    \item AVF (Attribute Value Frequency):  
    This method assigns anomaly scores to data instances based on the rarity of their attribute values. A higher score indicates that the data instance exhibits rare or unusual attribute values when compared to the rest of the dataset. AVF is effective in scenarios where certain attributes in the data (e.g., numerical or categorical values) are highly indicative of anomalies. This method is particularly suited for detecting outliers that deviate significantly from established patterns across multiple features \citep{koufakou_2007}.

    \item OC3 (One-Class Classification by Compression):
    It is based on one-class classification methods, which focus on defining a decision boundary around the normal class and detecting any data points that fall outside this boundary as anomalies. It employs compression to define normal behavior more rigorously and is particularly effective when anomalies are significantly different from normal data points, as is often the case with APTs. The major advantage of OC3 lies in its ability to operate in situations with an imbalanced dataset, where normal samples vastly outnumber the anomalous ones \citep{smets2011}.

    \item OD (Outlier Degree): This method calculates the distance between a data point and its neighbors in the feature space. The greater the distance, the higher the degree of outlierness of the data point. This approach is useful when anomalies are characterized by being significantly separated from the normal data points, which is often seen in the case of APTs that infiltrate a system in an unusual manner. OD captures the isolation aspect of outliers, making it effective in detecting rare events that stand apart from the majority of the dataset \citep{narita_2008}.

    \item IForest (Isolation Forest):  It is a decision-tree-based method specifically designed for anomaly detection. It isolates anomalies by creating random partitions in the dataset. Points that are isolated quickly by the tree structure are deemed anomalies. The strength of IForest lies in its efficiency and scalability, especially for large datasets. It is effective in detecting anomalies that are far from the core of the data, such as rare and novel attacks that differ from the bulk of the system's behavior \citep{xu2023deep}.

    \item OC-SVM (One-Class Support Vector Machine): It is a well-known method in the one-class classification category. It defines a decision function to separate normal data from the anomalous ones in the feature space using a hyperplane. This method aims to find a boundary that best contains most of the data, thereby identifying outliers that lie outside this region. OC-SVM is effective for detecting outliers, including APTs, by providing a clear separation between normal and anomalous patterns, especially when normal instances are well-defined  \citep{zhang2007one}.

    \item EE (Elliptic Envelope): It employs a statistical approach based on the assumption that the data follows a Gaussian distribution. By fitting an elliptical envelope around the data points, EE can identify anomalies as points lying outside this envelope. This method works well when the data is normally distributed, and anomalies deviate significantly from the expected covariance structure. While it is sensitive to the shape and scale of the data distribution, EE can effectively identify small deviations from the centroids of the data that could indicate underlying anomalies \citep{vishwakarma2023new}.

    \item LOF (Local Outlier Factor): It is a density-based anomaly detection algorithm that identifies anomalies by measuring the local deviation of a data point relative to its neighbors. It is often used for detecting outliers in datasets where anomalies are defined by their relative density compared to surrounding points \citep{zhang2007one}.

    \item The Frequent Pattern Outlier Factor (FPOF): It is a pattern-based anomaly detection method that identifies outliers by examining the frequency of patterns associated with each data point. Unlike distance- or density-based methods, FPOF analyzes the occurrence of frequent itemsets in the dataset. Each data point is represented as a transaction, and a frequent pattern mining algorithm identifies commonly occurring patterns above a specified support threshold. For a given data point, FPOF calculates an outlier score based on the number of frequent patterns the point supports, with fewer supported patterns indicating higher outlier potential \citep{he2005fp}.
\end{itemize}

In the first step of our evaluation, we will assess the impact of active learning on the anomaly detection performance of our framework by analyzing how iterative querying influences detection accuracy. Once the optimal configuration of our framework is determined, we will then compare its performance, using the best-selected parameters, against the aforementioned anomaly detection methods to provide a comprehensive benchmark analysis. While the compared methods employ different anomaly detection approaches, our primary focus is not merely to contrast SDA²E with these techniques. Instead, the key objective is to demonstrate how integrating active learning into SDA²E enhances anomaly detection performance, even when working with a limited amount of labeled data.\\
The key insight of our approach is that by continuously querying an oracle for the labels of the most uncertain samples in the dataset, we can refine the model’s understanding of both normal and anomalous data. This iterative process maximizes the utility of a small amount of labeled data, progressively enhancing the AutoEncoder model's ability to detect anomalies with each iteration. Thus, the comparison with other methods serves as a baseline, contextualizing the improvements achieved through active learning integration. Our experiments aim to demonstrate that even when starting with a limited labeled dataset, the framework can iteratively refine its performance and achieve results that are competitive, if not superior, to other state-of-the-art methods.

\subsection{Ranking Evaluation Method:}

To evaluate the effectiveness of the ranking produced by our anomaly detection framework, we use the Normalized Discounted Cumulative Gain (nDCG) scoring metric. nDCG is a widely used measure in ranking systems that quantifies how well the top positions in a ranked list are aligned with ground truth labels. This metric is particularly suitable for anomaly detection tasks, as it emphasizes the correct prioritization of anomalies at the top of the ranked list.\\
The Discounted Cumulative Gain (DCG) calculates the relevance of items in a ranked list, assigning higher importance to items at the top of the list. The DCG for a ranked list of $n$ items is defined as:
\[
\text{DCG}_n = \sum_{i=1}^n \frac{2^{r_i} - 1}{\log_2(i+1)},
\]
where:
\begin{itemize}
    \item $r_i$ is the relevance score of the item at position $i$, reflecting whether the item is anomalous or normal.
    \item $i$ is the rank position, with higher ranks (smaller values of $i$) receiving greater weight due to the logarithmic denominator.
\end{itemize}
The DCG score accumulates the relevance values, discounted logarithmically by the rank position, giving more importance to correctly identifying anomalies at the top of the list. 
To normalize the DCG score, the Ideal Discounted Cumulative Gain (IDCG) is computed for the same list sorted in perfect alignment with ground truth labels. The IDCG represents the maximum possible DCG achievable for a given dataset and is defined as:
\[
\text{IDCG}_n = \sum_{i=1}^n \frac{2^{r_i^*} - 1}{\log_2(i+1)},
\]
where $r_i^*$ represents the relevance scores in the ideal ranking order. The Normalized Discounted Cumulative Gain (nDCG) is then obtained by normalizing the DCG score by the IDCG score:
\[
\text{nDCG}_n = \frac{\text{DCG}_n}{\text{IDCG}_n}.
\]
nDCG values range from 0 to 1, where 1 indicates a perfectly ranked list that aligns with the ground truth. Higher nDCG scores indicate better ranking performance, as anomalies are prioritized closer to the top of the list.

In the context of anomaly detection, nDCG provides a meaningful evaluation metric, since it prioritizes anomalies in the ranked list, aligning with real-world requirements where critical anomalies must be identified promptly. By emphasizing the top-ranked items, nDCG captures the utility of the ranking system in reducing false positives and ensuring the effectiveness of anomaly detection models. This makes it a robust measure for comparing different ranking strategies and assessing model performance.

Although classical evaluation metrics such as the Receiver Operating Characteristic (ROC) curve or Area Under the Curve (AUC) are commonly used, they are not effective in this context due to the highly imbalanced nature of the datasets. Other alternative metrics such as Precision@K might be considered, however nDCG provides a more nuanced assessment in scenarios with highly imbalanced datasets. Indeed, unlike Precision@K, which evaluates the ratio of true anomalies in the top-K results, nDCG incorporates a graded relevance mechanism, accounting for the relative ranking positions of anomalies. These metrics tend to overestimate performance by ignoring the imbalance, failing to highlight the significance of properly ranking anomalies in such datasets.

\subsection{Datasets Description:}
The anomaly detection datasets used in this study come from two main sources: the first is the Defense Advanced Research Projects Agency (DARPA) \verb|Transparent| \verb|Computing TC| ~\citep{berrada_2019,BerradaCBMMTW20,Benabderrahmane21,darpa}, which contains records of user activities in various computer systems, including both normal activities and cyber-attacks carried out by malicious parties, and ingested in forty datasets of different attack scenarios (described in Table~\ref{datatable}). The second source of datasets are curated by \citep{pang2016unsupervised,pang2016outlier} (described in Table 2), and includes 12 different datasets spanning a wide variety of scenarios, including computer vision, advertising, public health, business,  etc \citep{pang2016unsupervised,pang2016outlier}. Since most datasets contained in this second source are not originally curated for anomaly detection tasks, we need to transform the datasets accordingly. Therefore, following \citep{pang2016unsupervised,pang2016outlier}, the smallest class(es) or a small random subset of the smallest class are selected as the anomaly class. \\
Taken together, both sources provide us with 52 datasets in total which can be categorized into 6 broad categories: cyber security, computer vision, business, health and demography, natural sciences, and miscellaneous.\\
The primary evaluation dataset for this study will be although the DARPA TC APT dataset, which contains multiple sub-datasets across various operating systems. This dataset is highly imbalanced and includes a small proportion of APT cyber attack (outliers), making it an ideal challenge for anomaly detection. The remaining datasets will be used primarily to validate the findings of the study and assess the robustness of the proposed anomaly detection framework.

\subsubsection{Cyber Security:}  
\subsubsection*{DARPA TC APT dataset:}

The first cyber security data source used in this paper comes from the Defense Advanced Research Projects Agency (DARPA)’s \verb|Transparent| \verb|Computing TC|\footnote{https://gitlab.com/adaptdata} program~\citep{darpa,berrada_2019,BerradaCBMMTW20,Benabderrahmane21,DBLP:journals/corr/abs-2006-07916,DBLP:journals/fgcs/BenabderrahmaneHVCR24}. The aim of this program is to provide transparent provenance data of system activities and component interactions across different operating systems (OS) and spanning all layers of software abstractions. Specifically, the datasets include system-level data, background activities, and system operations recorded while APT-style attacks are being carried out on the underlying systems. Preserving the provenance of all system elements allows for tracking the interactions and dependencies among components. Such an interdependent view of system operations is helpful for detecting activities that are individually legitimate or benign but collectively might indicate abnormal behavior or malicious intent.

Here we specifically employ DARPA’s data that has undergone processing conducted by the \verb|ADAPT| (Automatic Detection of Advanced Persistent Threats) project’s ingester~\citep{berrada_2019,BerradaCBMMTW20,Benabderrahmane21,DBLP:journals/corr/abs-2006-07916,DBLP:journals/fgcs/BenabderrahmaneHVCR24}. The records come from four different source OS, namely Android (called in the TC program Clearscope), Linux (called Trace), BSD (called Cadets), and Windows (called Fivedirections or 5dir). For each system, the data comes from two separate attack scenarios: scenario 1 and scenario 2, called Pandex (Engagement E1) and Bovia (Engagement E2), respectively. The processing includes ingesting provenance graph data into a graph database as well as additional data integration and de-duplication steps. The final data includes a number of Boolean-valued datasets (data aspects)
, with each representing an aspect of the behavior of system processes as illustrated in Table \ref{dataexample}. Each row in such a data aspect 
is a data point representing a single process run on the respective OS. It is expressed as a Boolean vector whereby a value of 1 in a vector cell indicates the corresponding attribute applies to that process. For instance, in Table \ref{dataexample}, the process with id\\ \verb|ee27fff2-a0fd-1f516db3d35f| has the following sequence of events: \verb|</usr/sbin/avahi-autoipd|, \verb|216.73.87.152|, \\\verb|EVENT_OPEN, EVENT_CONNECT, ...>|. Specifically, the relevant 
datasets are interpreted as follows:

\begin{itemize}
    \item \verb|ProcessEvent| (PE): Its attributes are event types performed by the processes. A value of 1 in \verb|process[i]| means the process has performed at least one event of type $i$.
    \item \verb|ProcessExec| (PX): The attributes are executable names that are used to start the processes.
    \item \verb|ProcessParent| (PP): Its attributes are executable names that are used to start the parents of the processes.
    \item \verb|ProcessNetflow| (PN): The attributes here represent IP addresses and port names that have been accessed by the processes.
    \item \verb|ProcessAll| (PA): This dataset is described by the disjoint union of all attribute sets from the previous datasets.
\end{itemize}
Overall, with two attack scenarios (Pandex, Bovia), four OS (BSD, Windows, Linux, Android) and five aspects (PE, PX, PP, PN, PA), a total of forty individual datasets are composed, as illustrated in Figure 2. 
\begin{figure}
    \centering
    \includegraphics[width=0.7\linewidth]{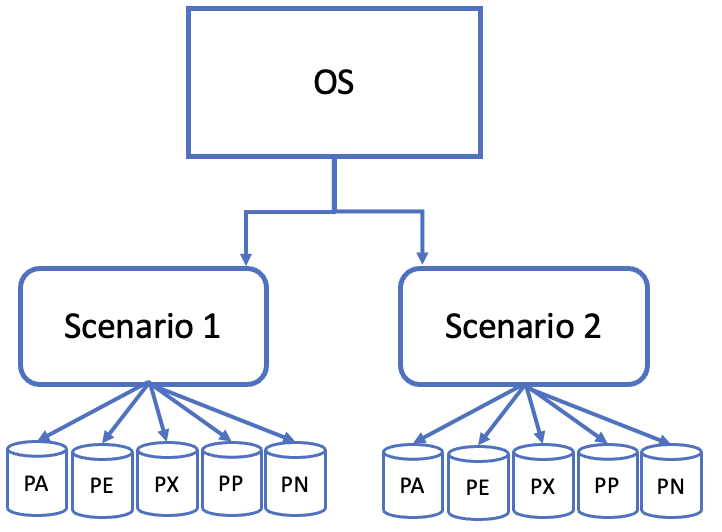}
    \caption{Organization of the DARPA's TC datasets. Each OS undergoes two attack scenarios, each of which contains five data aspects ets. With four OS (BSD, Windows, Linux, Android), two attack scenarios, and five aspects (PE, PX, PP, PN, PA), a total of forty individual datasets are composed. Each forensic configuration (OS$\times$attack scenario$\times$data aspect) represents a single dataset \citep{DBLP:journals/fgcs/BenabderrahmaneHVCR24}.}
    \label{fig:enter-label}
\end{figure}
They are described in Table~\ref{datatable} whereby the last column provides the number of attacks in each dataset. The substantially imbalanced nature of the datasets is clearly seen here. For each forensic configuration (OS$\times$attack scenario$\times$data aspect) we have the number of processes (instances) and the corresponding events (features). For instance, $Windows\_E1\_PE$ is the dataset represented by $PE$ aspect belonging to Windows OS, produced during the first attack scenario E1 (Pandex). It contains 17569 instances and 22 features with a total of 8 APTs anomalies (0.04\%). \\
Figure \ref{fig:densitywinner} illustrates some kernel density estimations derived from these datasets. This visual emphasizes the highly imbalanced nature of these datasets, particularly in their Advanced Persistent Threat (APT) annotations. The datasets exhibit vastly different distributions, with the majority of process attributes indicating routine operations while containing a negligible fraction of attack instances. This disparity underscores the challenge of detecting anomalies in a feature space dominated by normal operations. Such imbalance necessitates sophisticated mechanisms, including anomaly ranking and iterative refinement, to identify critical patterns that signify deviations or malicious activities.

\begin{table*}
\centering
\small
\resizebox{0.99\textwidth}{!}{
\begin{tabular}{|l|l||l|l|l|l|l|l|l|l|}
\hline & Scenario & Size& $PE$   & $PX$  & $PP$  & $PN$     & $PA$  & $nb\_attacks$    & $\%\frac{nb\_attacks}{nb\_processes}$     \\ \hline \hline
BSD    & 1 &288 MB &76903 / 29  & 76698 / 107  & 76455 / 24  & 31 / 136  & 76903 / 296 & 13&0.02\\  
    & 2 &1.27 GB &224624 / 31  &224246 / 135  & 223780 / 37  & 42888 / 62 &  224624 / 265      & 11&0.004\\ \hline
Windows & 1 &743 MB & 17569 / 22    &  17552 / 215  &   14007 / 77        &   92 / 13963      & 17569 / 14431& 8&0.04\\  
   & 2 &9.53 GB& 11151 / 30    &  11077 / 388  & 10922 / 84  & 329 / 125      &  11151 / 606    &8&0.07\\ \hline
Linux  & 1 &2858 MB &247160 / 24 & 186726 / 154 & 173211 / 40 & 3125 / 81 & 247160 / 299  &25&0.01\\
    & 2 &25.9 GB &282087 / 25 & 271088 / 140 & 263730 / 45 &6589 / 6225 &  282104 / 6435      &46&0.01\\ \hline
Android& 1 &2688 MB&102 / 21     &102 / 42&0 / 0&8 / 17& 102 / 80&9&8.8\\
&2 &10.9 GB&12106 / 27     &12106 / 44&0 / 0&4550 / 213&12106 / 295 &13&0.10\\ \hline
\end{tabular}
}
\caption{Summary of the first source of 40 benchmark datasets belonging to DARPA's TC program for APT detection. A dataset entry (columns 4 to 8) is described by a number of rows (processes) / number of columns (attributes). For instance, with ProcessAll (PA) obtained from the second scenario using Linux, the dataset has 282104 rows and 6435 attributes with 46 APT attacks (0.01\%) \citep{Benabderrahmane21}. }
 \label{datatable}
\end{table*}
\begin{table}[h!]
\centering
\scriptsize
\begin{tabular}{l|c|c|c}
\hline
{Dataset}                & \multicolumn{1}{c|}{{\# Instances}} & \multicolumn{1}{c|}{{\# Features}} & \multicolumn{1}{c|}{{\# Anomalies (\%)}} \\ \hline \hline
{KDD U2R}         &    60821                                        & 83                                          &                               228 (0.13\%)                 \\ \hline
{KDD Probe}       &  64760                                          &   83                                        &              4166    (6.43\%)                               \\ \hline
{CMC}             &     1474                                       &    23                                       &   29       (1.94\%)                                       \\ \hline
{AID362}          &  4280                                          &    118                                       &     60         (1.38\%)                                   \\ \hline
{aPascal (APas)}         & 12696                                           &  65                                         &     176   (1.38\%)                                           \\ \hline
{BM}              & 41189                                           &   54                                        &         4640      (11.26\%)                                    \\ \hline
{AD}              &  3280                                          &   1556                                        &    459      (14\%)                                        \\ \hline
{CoverType} (CT)       & 581013                                     & 45                                        &2747                (0.47\%)                                 \\ \hline
{CelebA}          &   202600                                         &       40                                  &               4547    (2.24\%)                              \\ \hline
{Reuters10} (R10)      &    12898                                        &  101                                         &        237          (1.83\%)                               \\ \hline
{Solar\_Flare} (SF)   &   1067                                         &   42                                        &        43            (4.02\%)                             \\ \hline
{w7a\_LibSVM} (W7A)   &   49750                                         &   301                                        &         1479       (2.97\%)                                 \\ \hline
\end{tabular}
\label{tabsecondsource}
\caption{Summary of the second source of benchmark datasets, including dataset names, number of instances, feature dimensions, and the proportion of anomalies present. The datasets span various domains such as cybersecurity, demographic studies, highlighting diverse challenges in handling imbalanced and high-dimensional data.}
\end{table}


\begin{table}
\scriptsize
\begin{tabular}{|l|c|c|c|c|c|c|c|}
\hline
                                           & {\rotatebox{90}{/usr/sbin/avahi-autoipd}} & {\rotatebox{90}{216.73.87.152}}   & {\rotatebox{90}{EVENT\_OPEN} }    & {\rotatebox{90}{EVENT\_EXECUTE}}  & {\rotatebox{90}{EVENT\_CONNECT} } & {\rotatebox{90}{EVENT\_SENDMSG} } & \multicolumn{1}{l|}{...} \\ \hline
{ee27fff2-a0fd-1f516db3d35f} & 1                                & 1                        & 1                        & 0                        & 1                        & 0                        & ...                      \\ \hline
{b2e7e930-8f25-4242a52c5d72} & 0                                & 1                        & 0                        & 1                        & 1                        & 1                        & ...                      \\ \hline
{07141a2a-832e-8a71ca767319} & 0                                & 0                        & 1                        & 1                        & 1                        & 1                        & ...                      \\ \hline
{b4be70a9-98ac-81b0042dbecb} & 1                                & 0                        & 1                        & 1                        & 0                        & 0                        & ...                      \\ \hline
{2bc3b5c6-9110-076710a13038} & 0                                & 0                        & 0                        & 0                        & 0                        & 1                        & ...                      \\ \hline
{ad7716e0-8d59-5d45d1742211} & 1                                & 1                        & 0                        & 1                        & 0                        & 1                        & ...                      \\ \hline
...                                           & \multicolumn{1}{l|}{...}         & \multicolumn{1}{l|}{...} & \multicolumn{1}{l|}{...} & \multicolumn{1}{l|}{...} & \multicolumn{1}{l|}{...} & \multicolumn{1}{l|}{...} & \multicolumn{1}{l|}{...} \\ \hline
\end{tabular}
\caption{Example of a boolean-valued dataset (data aspect) from the DARPA TC program. In each row, the boolean vector represents the list of features of the corresponding process. }
 \label{dataexample}
\end{table}
%
\begin{figure*}
\includegraphics[width=0.230\linewidth]{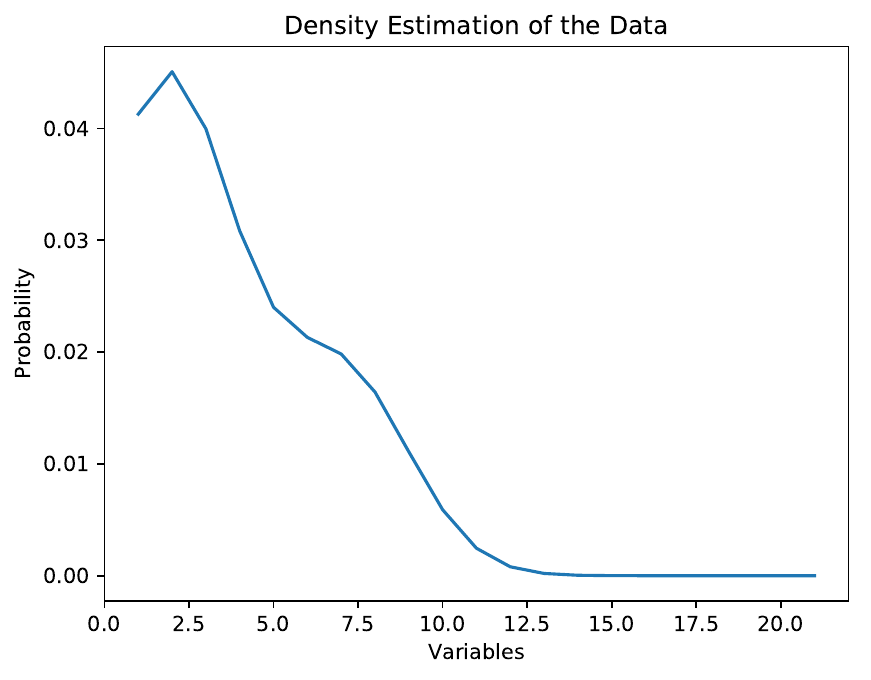} \quad
\includegraphics[width=0.230\linewidth]{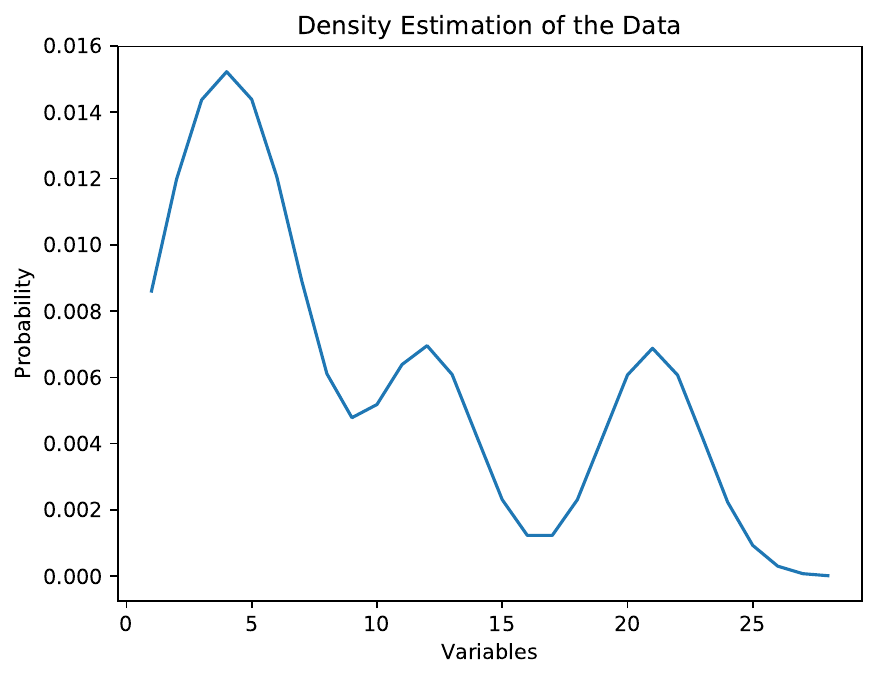}
\medskip
\includegraphics[width=0.230\linewidth]{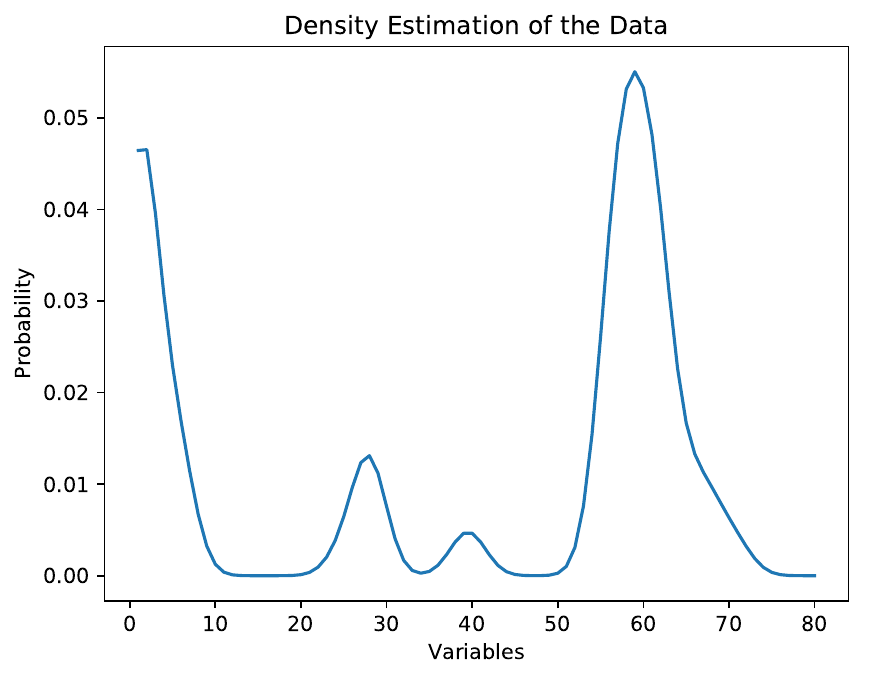}\quad
\includegraphics[width=0.23\linewidth]{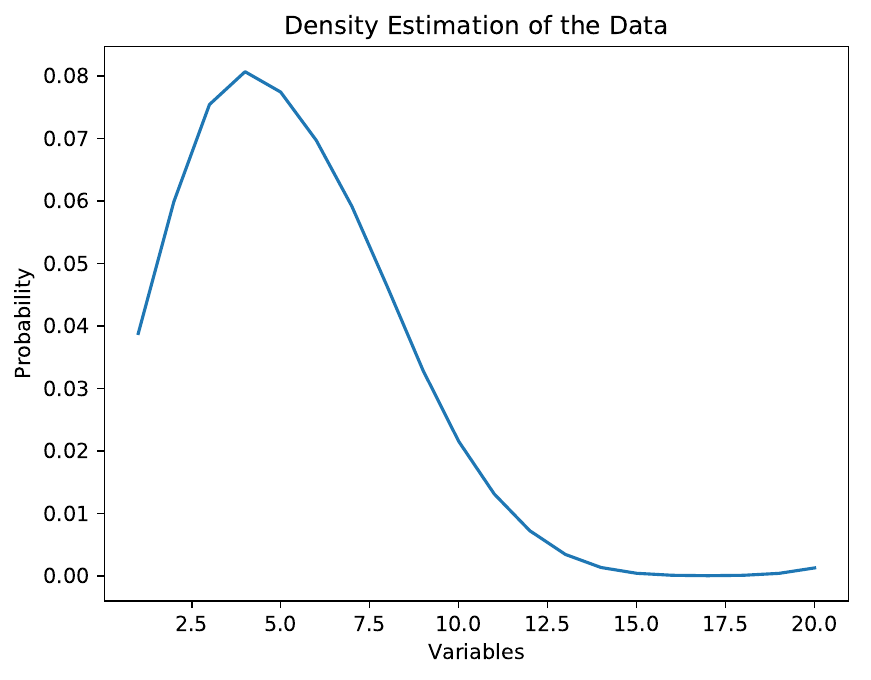}
\caption{Examples of some kernel density estimation of the DARPA's TC datasets showing the class imbalance (From left to right: Windows$\times$E1$\times$PE, BSD$\times$E1$\times$PE, Linux$\times$E1$\times$PN and Android$\times$E1$\times$PE. }
\label{fig:densitywinner}
\end{figure*}
\subsubsection*{NSL-KDD Probe:}  
The KDD Probe dataset\footnote{http://kdd.ics.uci.edu/databases/kddcup99/kddcup99.html} focuses on anomaly detection in computer network traffic and involves identifying suspicious or malicious activity based on the characteristics of network probes, such as connection attempts and server responses \citep{su2020bat,vibhute2024towards}. Probe or surveillance is an attack that tries to get information from a network. Key features in the dataset include network statistics, session data, and packet-level descriptors, which are particularly valuable for understanding the nature of computer network interactions and detecting potential threats. These features help distinguish between normal and abnormal behavior in a network environment. For anomaly detection, the Probe dataset can be used to identify suspicious network activities, such as abnormal connection attempts, unusual traffic patterns, or unsolicited requests that might indicate intrusion attempts or other cybersecurity threats.

\subsubsection*{NSL-KDD U2R:}  
The U2R (User to Root) dataset\footnote{https://github.com/thinline72/nsl-kdd} contains data on user-to-root attacks in cybersecurity \citep{thana2024machine,bala2019review}. It focuses on detecting anomalies where an attacker gains root access to a system, typically via various forms of exploitation and privilege escalation. U2R attacks start off with a normal user account and try to gain access to the system or network, as a super-user (root). The attacker attempts to exploit the vulnerabilities in a system to gain root privileges/access. Key features include system call data, user activity logs, and features based on the behavior of normal vs. anomalous system activities. 

\subsubsection{Computer Vision}  
\subsubsection*{aPascal (APAS):} This dataset\footnote{http://vision.cs.uiuc.edu/attributes/} is derived from the Pascal Visual Object Classes (VOC) benchmark and is used for object recognition and attribute prediction \citep{pascal2017pascal}. It labels objects in images with descriptive attributes (e.g., "furry," "metallic") and object categories. The key features of the dataset include attributes paired with object types, which enable attribute-based object analysis. It is well-annotated for tasks involving multi-class classification, clustering, and anomaly detection. The dataset provides a link between visual attributes and higher-level concepts in object recognition. In terms of anomaly detection, this dataset can be used to detect object attribute anomalies, such as objects with unusual attribute combinations (e.g., "furry laptop" or "metallic animal"). It can help identify unusual object occurrences by detecting objects that are visually inconsistent with the dataset’s main categories (e.g., detecting a "dog" object mislabeled as "cat"). It can also be used for attribute-based noise detection, identifying mislabeled attributes or data points that disrupt attribute-class relationships. 

\subsubsection*{CelebA (CelebFaces Attributes Dataset):}  
This dataset\footnote{http://mmlab.ie.cuhk.edu.hk/projects/CelebA.html} contains over 200,000 celebrity images, each annotated with 40 binary attribute labels, such as smiling, wearing glasses, and beard \citep{liu2015faceattributes}. Key features of the dataset include rich attribute annotations for each image, capturing detailed facial characteristics. The dataset includes a diverse range of face poses, expressions, and backgrounds, making it suitable for robust statistical analysis and deep learning tasks. In anomaly detection, CelebA can be used for identifying outlier attributes by detecting anomalous images with unusual attribute combinations, such as a "bearded" label in predominantly female faces or a "smiling" label paired with unexpected facial structures. It can help detect faces that deviate significantly in terms of attributes compared to the majority, such as rare attributes like unique eye-wear styles. Additionally, it can be used for outlier detection in data preprocessing by identifying mislabeled or corrupted data points that could impact model training.

\subsubsection*{AID362:}  
The AID362 dataset\footnote{https://sites.google.com/site/gspangsite/sourcecode/categoricaldata} is a bioassay database derived from the PubChem BioAssay repository, specifically designed for analyzing molecular activity against biological targets\citep{li2017adaptive,Sandeep24}. It consists of chemical compounds labeled as either active or inactive based on their experimental outcomes, making it suitable for binary classification tasks. The dataset is particularly challenging for machine learning models due to its significant class imbalance, as the majority of compounds are labeled as inactive. In the context of anomaly detection, AID362 can be utilized to identify rare active compounds, which are considered anomalies within the dataset. Detecting these rare compounds is critical in fields such as drug discovery, where uncovering potential therapeutic molecules among vast numbers of inactive candidates can save resources and accelerate research. By focusing on these anomalies, researchers can develop and evaluate models that not only classify compounds but also highlight outliers with potential biological significance.

\subsubsection*{LIBSVM W7A:}  
The w7a LIBSVM dataset, originating from the PASCAL Large Scale Learning Challenge, is a high-dimensional, sparse dataset often used to evaluate machine learning models for binary classification tasks \citep{jiang2020svm,platt1998fast}. Characterized by severe class imbalance, it is particularly suited for anomaly detection studies, where the minority class can be treated as anomalous or outliers. Its text-derived nature, encoding sparse feature representations typical of natural language processing tasks, makes it a challenging benchmark for models to detect rare events amidst dominating patterns. The dataset's structure and imbalance closely mimic real-world scenarios like fraud detection or rare class identification, enabling researchers to assess and refine the robustness, scalability, and accuracy of anomaly detection methods under difficult conditions.

\subsubsection{Business}  
\subsubsection*{Bank Marketing (BM):} The Bank Marketing dataset\footnote{https://archive.ics.uci.edu/ml/datasets/Bank+Marketing} originates from a direct marketing campaign of a Portuguese banking institution \citep{elsalamony2014bank}. The primary goal is to predict whether a client will subscribe to a term deposit. Key features of the dataset include client demographic information, bank account details, and marketing campaign responses. It provides rich categorical and numerical features for customer behavior analysis. The dataset includes over 45,000 data points with balanced labels for deposit subscription. In the context of anomaly detection, the dataset can be used to identify customer profiling anomalies, by detecting atypical clients who deviate significantly from typical customer profiles. It helps detect outlier marketing responses, such as unusual responses to marketing campaigns, repeated refusals, or unexpected acceptance. The dataset can also be used for data quality verification by ensuring consistency and detecting anomalies like unrealistic ages or missing account details.

\subsubsection*{Internet Advertisements (AD):}  
This dataset\footnote{https://archive.ics.uci.edu/ml/datasets/Internet+Advertisements} contains manually labeled web ad content aimed at distinguishing between advertisements and non-advertisement content. Key features include text and image-based attributes for web page content, but it is a challenging dataset with noisy and partially missing data. It is suitable for supervised classification and outlier detection tasks. In anomaly detection, this dataset can be used for ad detection anomalies by identifying mislabeled web content as ads or non-ads. It helps detect data noise, identifying noisy features and anomalous attribute values within web page data. It can also detect unusual content patterns by spotting rare and unconventional advertisements.

\subsubsection{Public Health and Demography}  
\subsubsection*{Contraceptive Method Choice (CMC):}   This dataset\footnote{https://archive.ics.uci.edu/ml/datasets/Contraceptive+Method+Choice} focuses on demographic and socioeconomic factors influencing contraceptive methods among women in Indonesia. Key features include a multivariate dataset with numeric and categorical features, capturing attributes such as age, education, and marital status. This dataset is related to a sensitive and socially significant domain. For anomaly detection, the dataset can be used to detect demographic outliers, identifying women with unusually rare combinations of attributes, such as education level and marital status. It helps spot behavioral anomalies by identifying atypical contraceptive choices compared to the dominant patterns. Additionally, it can detect data noise, identifying errors in demographic or response attributes.

 %
%
\subsubsection{Natural Sciences}  
\subsubsection*{Solar Flare (SF):}  
The Solar Flare dataset\footnote{https://archive.ics.uci.edu/ml/datasets/Solar+Flare} records historical solar activity and features attributes predicting the occurrence of solar flares based on geographical and magnetic characteristics. Key features include a predictive dataset for solar event classification, composed mostly of ordinal data derived from astrophysical observations. The dataset is used in reliability studies for space-weather forecasting. In anomaly detection, this dataset can be used to detect event prediction outliers by identifying regions or conditions with unexpectedly high or low solar activity. It helps analyze unusual patterns by detecting unusual correlations or configurations in magnetic parameters. The dataset is also useful for data consistency checking, highlighting anomalies due to misrecorded observational data.

\subsubsection*{CoverType (CT):}  
The CoverType dataset\footnote{https://archive.ics.uci.edu/ml/datasets/covertype} predicts forest cover type based on cartographic and ecological data from US forest regions. Key features include a large dataset with over 500,000 data points and 54 attributes, such as elevation, aspect, soil type, and wilderness areas. The dataset serves as a robust benchmark for classification and clustering tasks. In anomaly detection, CoverType can be used to detect ecological outliers by identifying forest patches with unexpected attributes, such as extreme elevation or soil characteristics. It can help predict cover anomalies by identifying instances where the cover type is inconsistent with ecological features. Additionally, it is useful for detecting geographic deviations, highlighting abnormalities that may indicate data recording errors or rare natural phenomena.

\subsubsection{Miscellaneous}  
\subsubsection*{Reuters10 (R10):}  
This dataset\footnote{https://sites.google.com/site/gspangsite/sourcecode/categoricaldata} is a subset of the Reuters Corpus Volume 1 (RCV1) focused on text mining and clustering. It is particularly tailored for high-dimensional datasets, often used in text categorization and anomaly detection research. Key features include labeled textual data structured into topics or categories. The dataset’s high dimensionality and sparsity make it challenging for text-based outlier detection. It is often vectorized into categorical or numerical features for machine learning studies. For anomaly detection, Reuters10 can be used to identify topic outliers by detecting documents that deviate significantly from typical topics, such as a finance article appearing in a dataset dominated by sports topics. It helps detect word distribution anomalies by identifying documents with unusual or sparse word distributions that don't align with the dominant document clusters. The dataset is also useful for noise removal, helping identify corrupted or mislabeled documents in preprocessing pipelines for text models.

\section{Results:}
\subsection{Computational environment}
All experiments were conducted on a machine running macOS 14.5 with an Apple M1 Max chip, 64 GB RAM.
\subsection{Code and data availability:}
The full implementation, preprocessing scripts, databases, and configuration files needed to reproduce all experiments are publicly available in the project's repository.
\subsection{Hyperparameter Selection Strategy}
The SDA$^2$E framework integrates multiple loss components— reconstruction, sparsity, adversarial, and attention—each controlled by a corresponding coefficient ($\alpha$, $\beta$, $\gamma$, $\delta$), in addition to structural
hyperparameters governing sparsity, attention, and adversarial separation. Because the datasets evaluated in this study exhibit substantial heterogeneity in scale, noise level, and feature redundancy, exhaustive per-dataset tuning would both risk overfitting and significantly expand the paper's scope. Instead, an exploratory grid search was conducted on a small subset of
representative datasets (one per domain) to identify a stable configuration that consistently balanced detection performance and training stability. The loss-weight coefficients were fixed for all experiments as
$\alpha = 1.0$, $\beta = 0.1$, $\gamma = 0.01$, and $\delta = 0.1$. This setting prioritizes accurate reconstruction while enforcing controlled
latent sparsity and stable adversarial learning, without allowing either the discriminator or attention mechanism to dominate optimization. The adversarial margin parameter was set to $m = 1.0$, which defines the minimum
reconstruction-energy separation enforced by the discriminator between real and generated samples. This value was found to provide sufficient separation
without destabilizing adversarial training. The attention regularization weight was fixed to $\lambda = 0.01$, encouraging sparse and interpretable feature-wise attention masks while preserving reconstruction fidelity. The sparsity target for latent activations was set to
$\rho = 0.1$, enforcing selective neuron activation consistent with standard sparse autoencoder practice.
All hyperparameters were held constant across datasets to ensure reproducibility and to demonstrate the robustness of SDA$^2$E under a unified configuration, rather than relying on dataset-specific optimization.

\subsection{Training loss of the AutoEncoder:}
\begin{figure}[h!]
    \centering
    \includegraphics[width=0.9\linewidth]{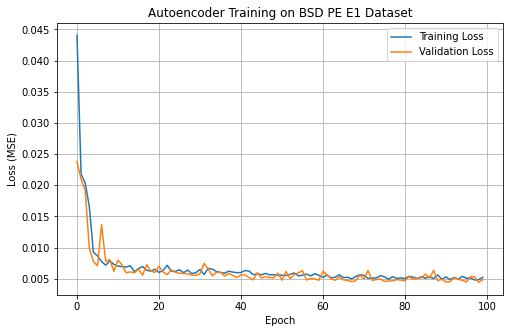}
    \caption{Learning curves illustrating the training and validation losses of the proposed Autoencoder trained on the BSD PE E1 Dataset, measured by Mean Squared Error (MSE) across 100 epochs.}
    \label{fig:AAELoss}
\end{figure}

The presented figure \ref{fig:AAELoss} shows the learning behavior of the Attention-based Autoencoder during training and validation phases on the BSD$\times$PE$\times$E1 dataset. Both curves exhibit a rapid initial decrease in loss, indicating efficient learning of the underlying data representation during early epochs. As training progresses, the training and validation losses stabilize and converge closely, signifying that the model effectively captures the key data patterns without significant overfitting. The close alignment between training and validation curves implies good generalization performance and supports the effectiveness of the AutoEncoder's attention mechanism in prioritizing informative features and generating robust embeddings suitable for subsequent anomaly detection tasks. Similar behavior has been observed for other datasets.

\subsection{Embeddings visualization:}
To ensure the practicality of our approach, we limited the active learning process to 20 iterations. This decision balances efficient forensic investigation with real-world constraints, such as the high cost of oracle interactions. Indeed, limiting queries reduces the overall cost while emphasizing rapid convergence of the detection rate, which is critical in anomaly detection. For instance, in cybersecurity, delaying responses with a large number of iterations can create during the oracle querying, windows of opportunity for attackers to deepen infiltration and potentially cause significant harm. Throughout the experiments that we performed, the model reliably detected true positives early, prioritizing anomalies within a restricted budget. This ability to converge on accurate results quickly demonstrates its suitability for scenarios where timely detection and resource efficiency are paramount.
Rapid convergence in anomaly detection is essential to maintain operational relevance in high-stakes scenarios like advanced threat detection. Extending the iterations to analyze the entire dataset would demand excessive time and resources, potentially overwhelming analysts with prolonged investigations. Instead, emphasizing swift detection minimizes the risk of delayed responses while alleviating the burden on analysts, aligning closely with the demands of live cybersecurity systems. By balancing thoroughness with timeliness, our model effectively supports operational environments requiring fast and reliable responses to mitigate harm.\\
\begin{figure*}[th!]
    \centering
   
     \includegraphics[width=0.35\linewidth]{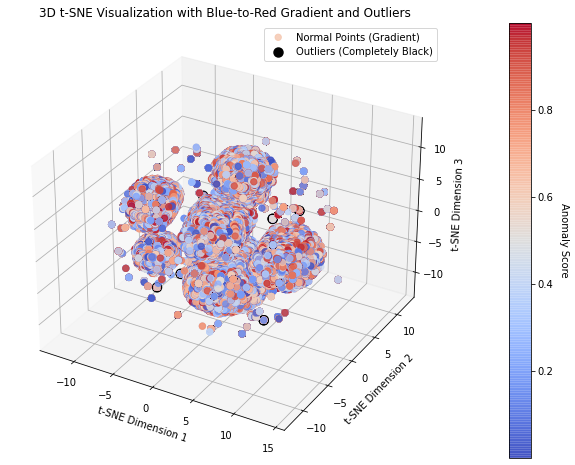} 
     \includegraphics[width=0.35\linewidth]{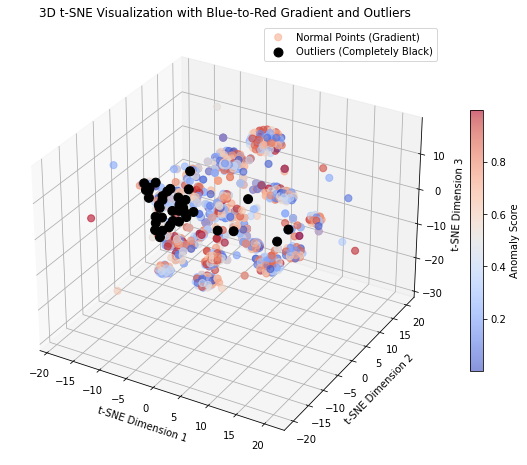}    
     \includegraphics[width=0.35\linewidth]{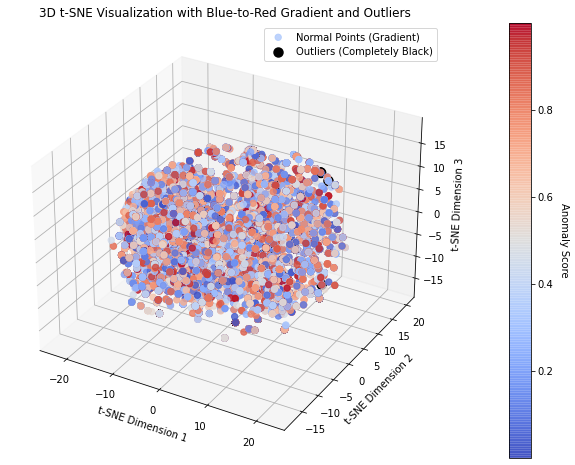}
     \includegraphics[width=0.35\linewidth]{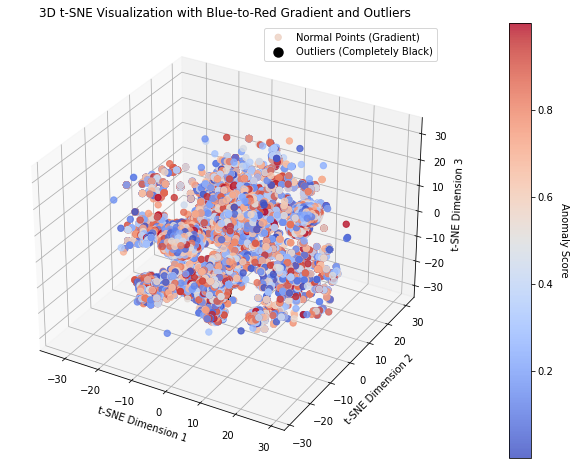}   
    \includegraphics[width=0.35\linewidth]{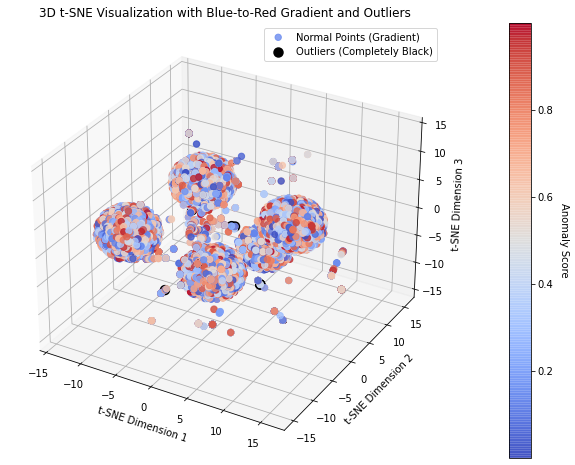} 
     \includegraphics[width=0.35\linewidth]{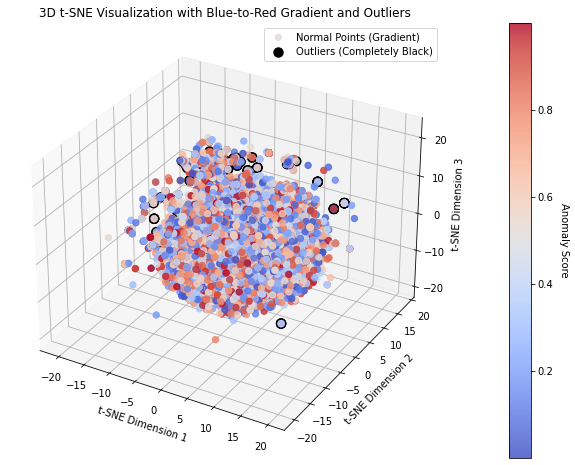}      
     \includegraphics[width=0.35\linewidth]{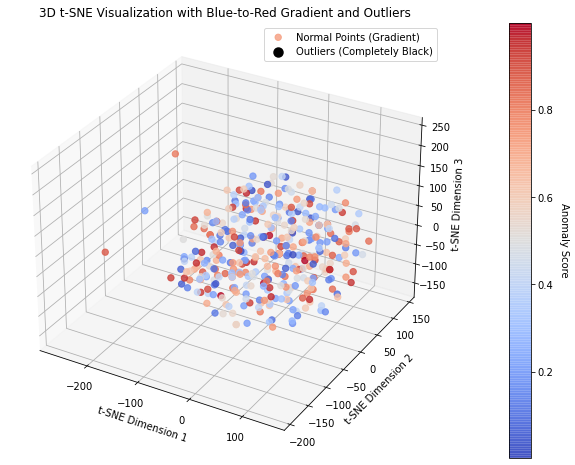} 
     \includegraphics[width=0.35\linewidth]{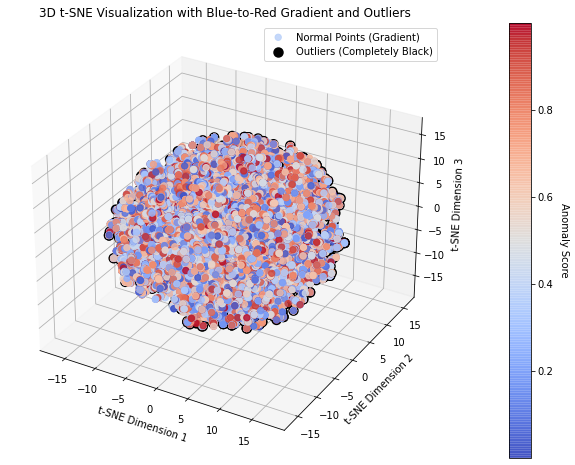} 
    \caption{3D t-SNE projection of the feature space derived from some datasets (Left from top to bottom: Linux$\times$E1$\times$PE, BSD$\times$E1$\times$PE, Windows$\times$E1$\times$PA, Android$\times$E1$\times$PA. Right from top to bottom: Solar Flare, Apascal, Reuters10, Bank). The embedding shows clusters of normal points (blue-to-red gradient based on anomaly scores: blue represents processes with low anomaly scores and red represents highly anomalous processes), with solid black points marking the outliers in the feature space. Around the black points, which represent detected outliers, lie regions of uncertainty where the model should focus its exploration to identify potential anomalies, leveraging active learning to refine its detection capabilities.} 
    \label{fig:t-SNE}
\end{figure*}
%
%

To have a global overview on the data, we initially generated the t-SNE projections as illustrated in Figure \ref{fig:t-SNE}, which provide the embedding vectors of the data based on their feature interactions, allowing us to visualize the structure of the feature space. The points in the plot represent individual items in the datasets, clustered based on their embedding similarity in the feature space. Normal points are colored in a gradient from blue to red, reflecting their anomaly scores during the training process. Outliers are distinctly highlighted as solid dark markers, representing individuals that are separated from the main clusters or located in sparse regions of the embedding. Their isolated positions suggest a significant deviation from the normal patterns, marking them as anomalous in the feature space. However, for some datasets, certain outlier points were found in close proximity to uncertain points, further complicating the anomaly detection task by making it challenging to distinguish true anomalies from samples in ambiguous regions. The distribution of points further indicates that the embedding captures variations in feature interactions, separating normal and anomalous behavior into meaningful regions.

\subsection{nDCG scoring during active learning:}
Next, we present the variation of the nDCG scores during active learning using the different datasets. We firstly illustrate results of the DARPA TC datasets, with which we show the results by operating system, aligning with the methodology adopted in previous studies \citep{BerradaCBMMTW20,berrada_2019,benabderrahmane_2019,Benabderrahmane21,DBLP:journals/fgcs/BenabderrahmaneHVCR24}. Within each operating system, the most effective active learning strategy is identified as the one yielding the highest nDCG scores, regardless of the specific sub-dataset considered. This ensures consistent evaluation of each method’s performance across sub-datasets while maintaining comparability with established benchmarks.\\
The following figures (\ref{fig:bsdPandexBoviaSAL} to \ref{fig:w7a}) illustrate the variation of nDCG scores over active learning iterations, comparing the performance of three similarity search strategies: S1, S2, and H (line plot on the left). Each figure also includes box plots on the right summarizing the overall distribution of nDCG scores for each strategy across iterations. This dual representation helps to analyze both the progression of decision boundary refinement and the consistency of the strategies. 
Following, is a summary of the different findings.
\subsection{BSD Operating System}

\begin{figure*}[th!]
    \centering
 \includegraphics[width=0.4\linewidth]{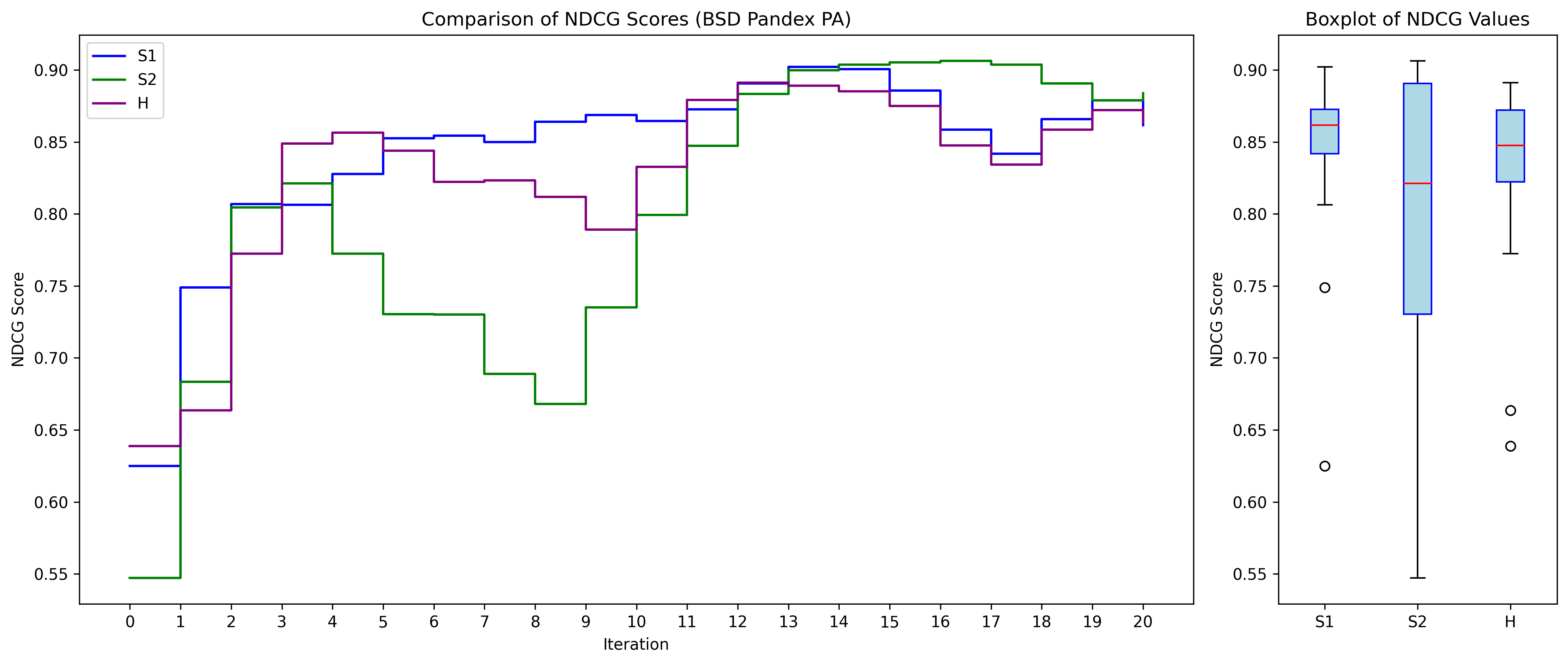}
         \includegraphics[width=0.4\linewidth]{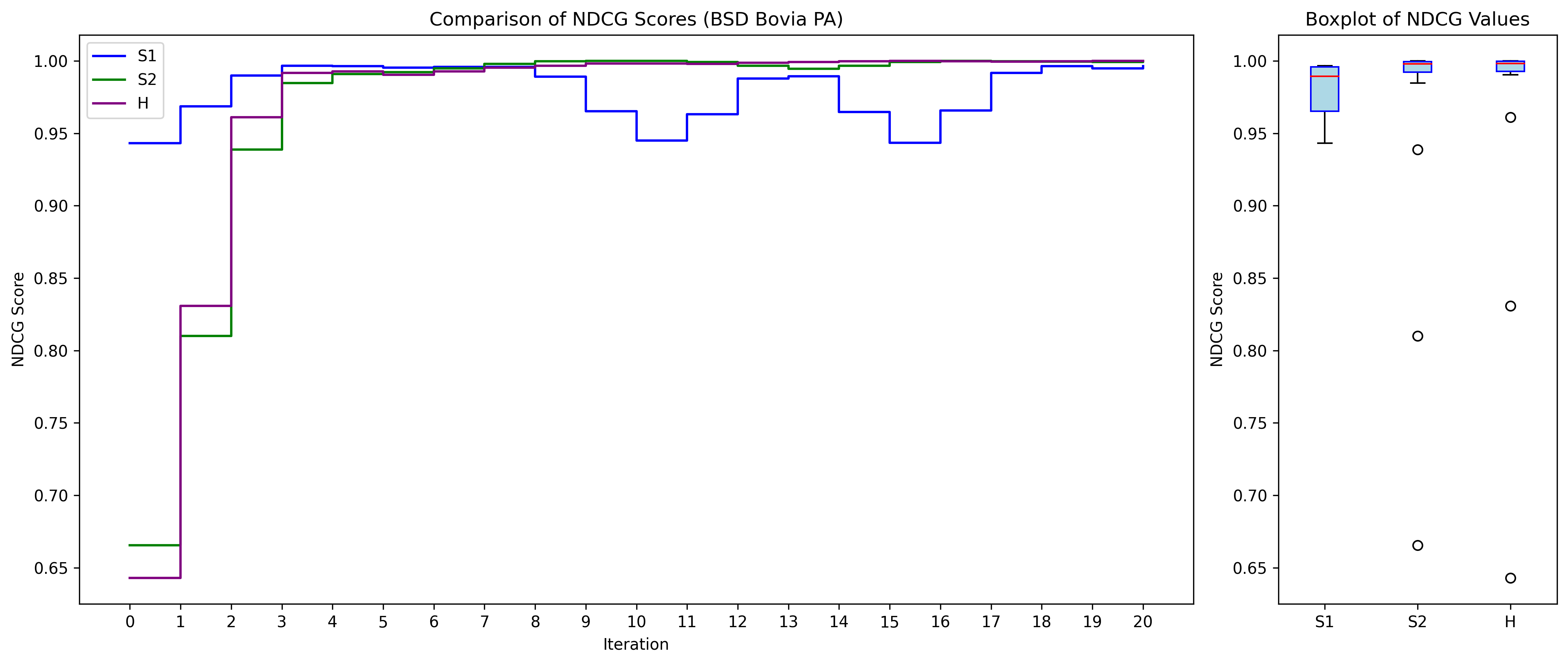} 
        
       \includegraphics[width=0.4\linewidth]{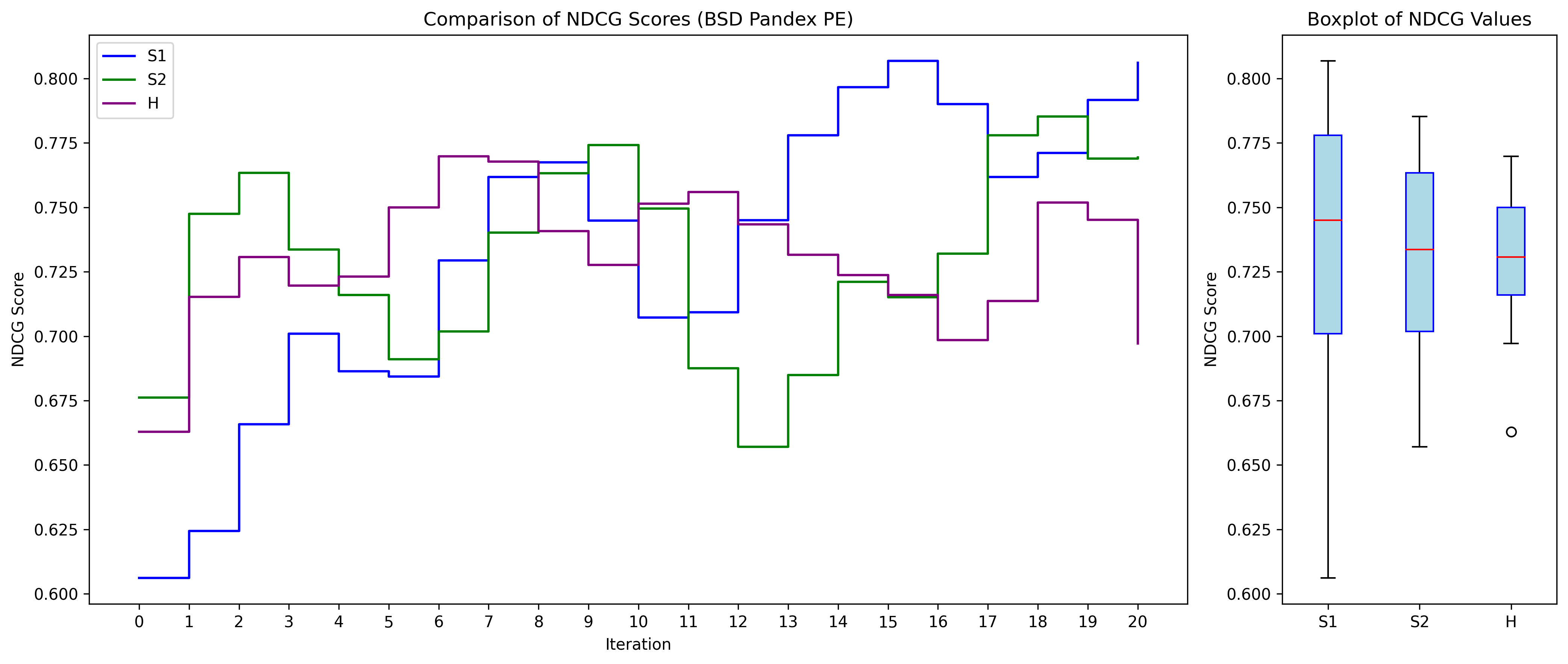} 
        \includegraphics[width=0.4\linewidth]{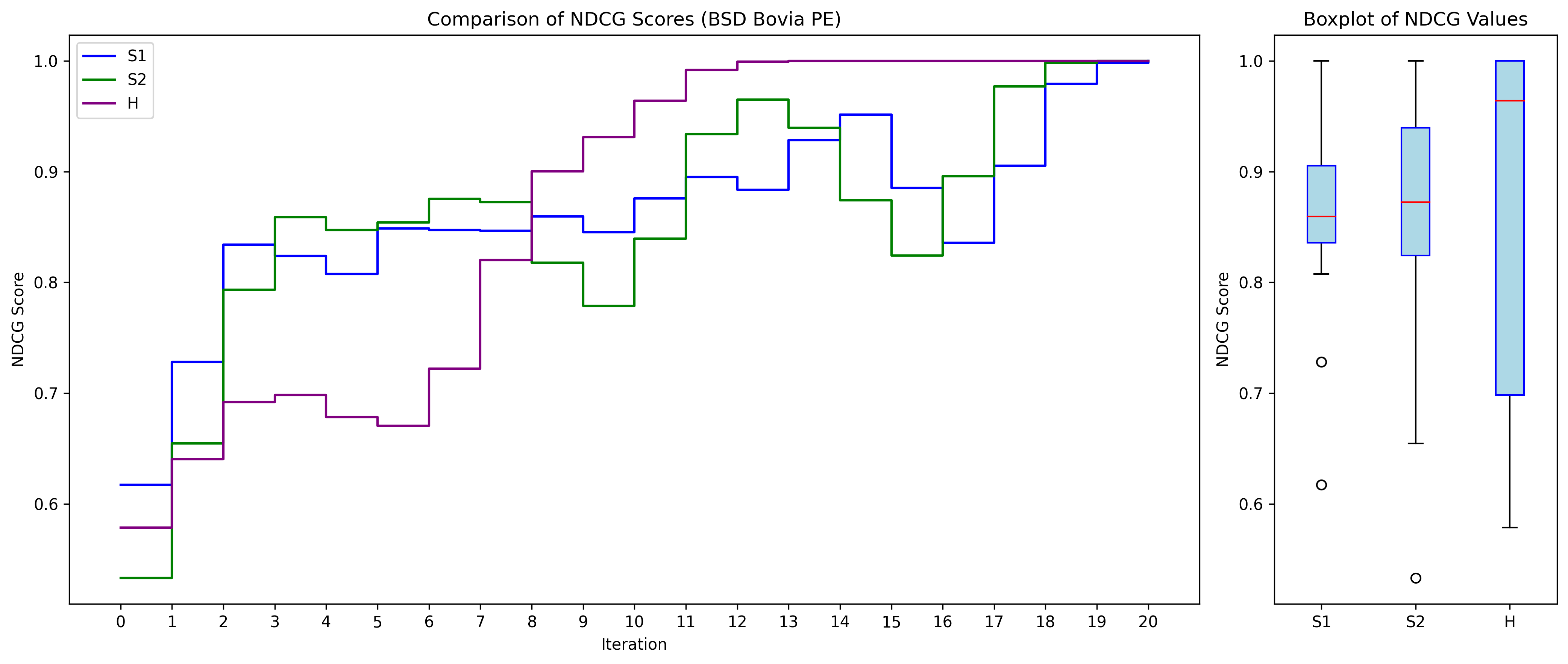} 
        
       \includegraphics[width=0.4\linewidth]{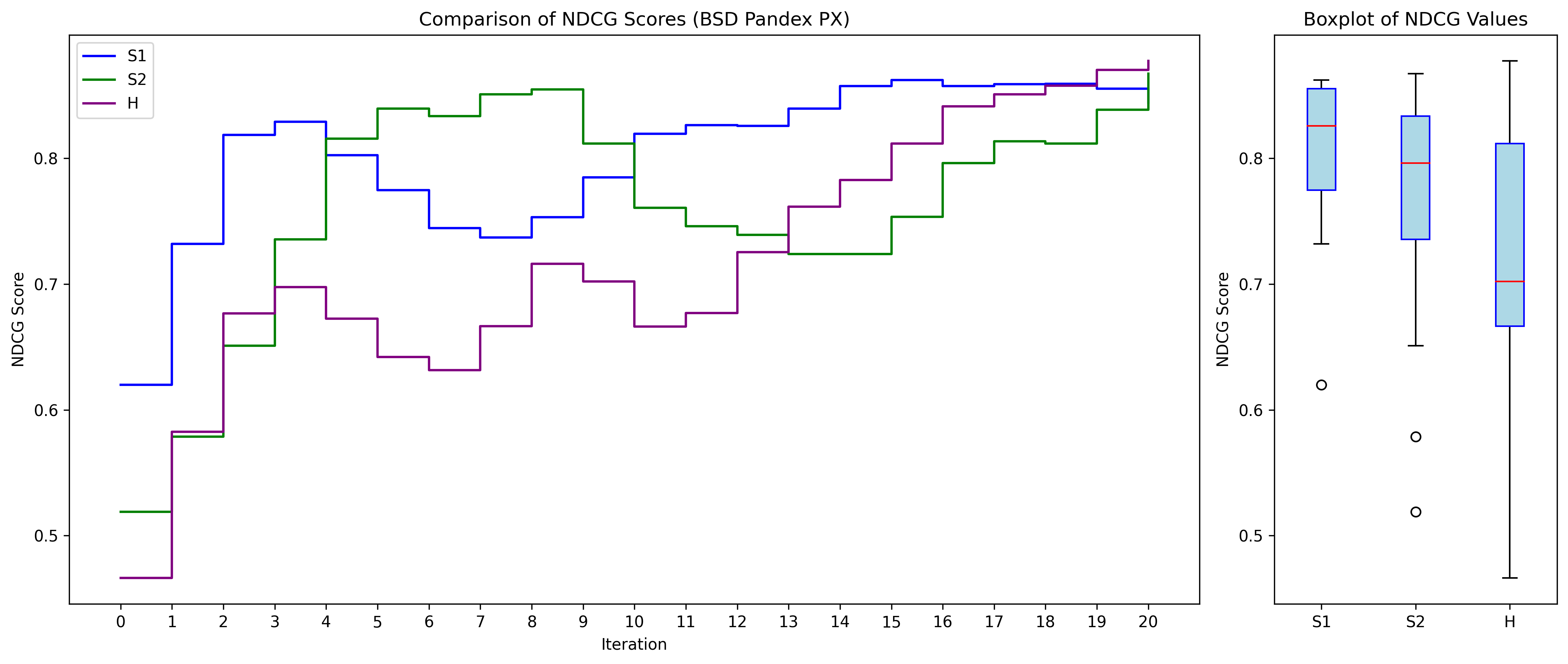} 
       \includegraphics[width=0.4\linewidth]{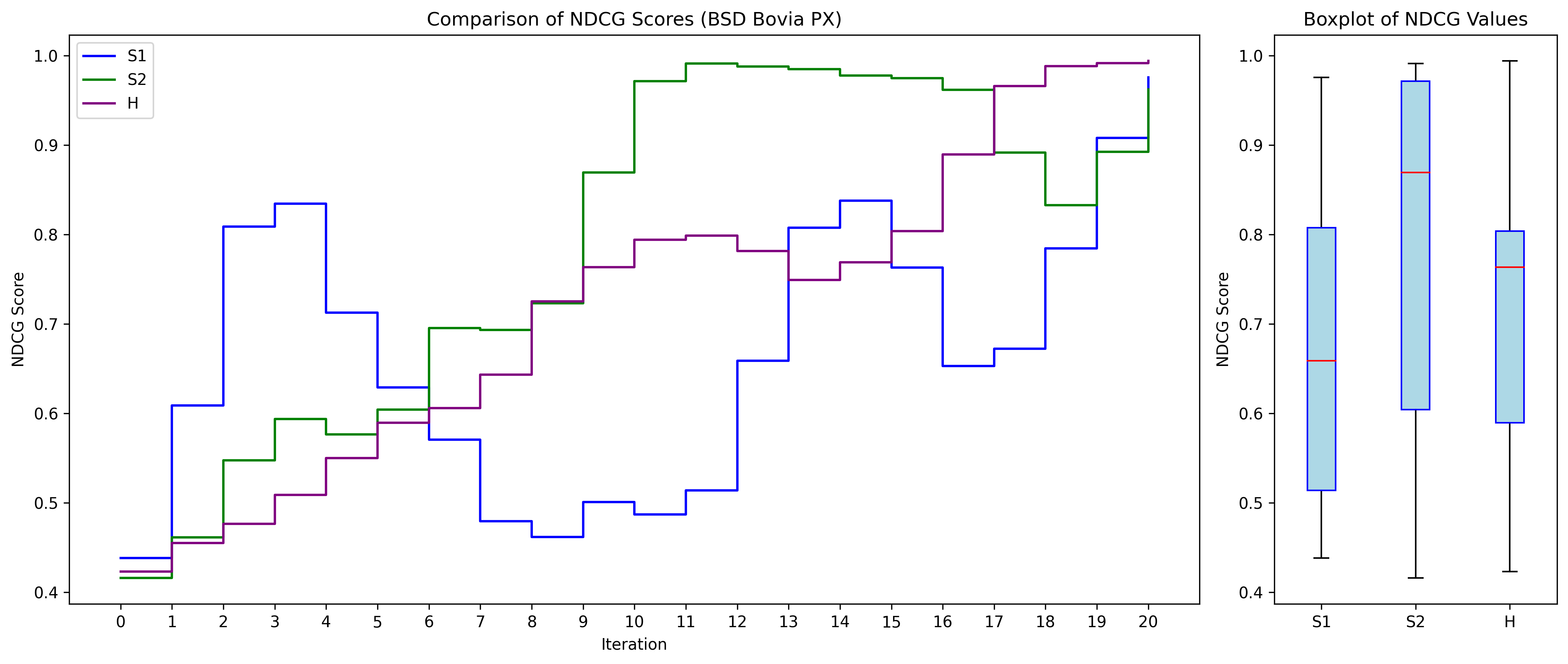} 
        
     \includegraphics[width=0.4\linewidth]{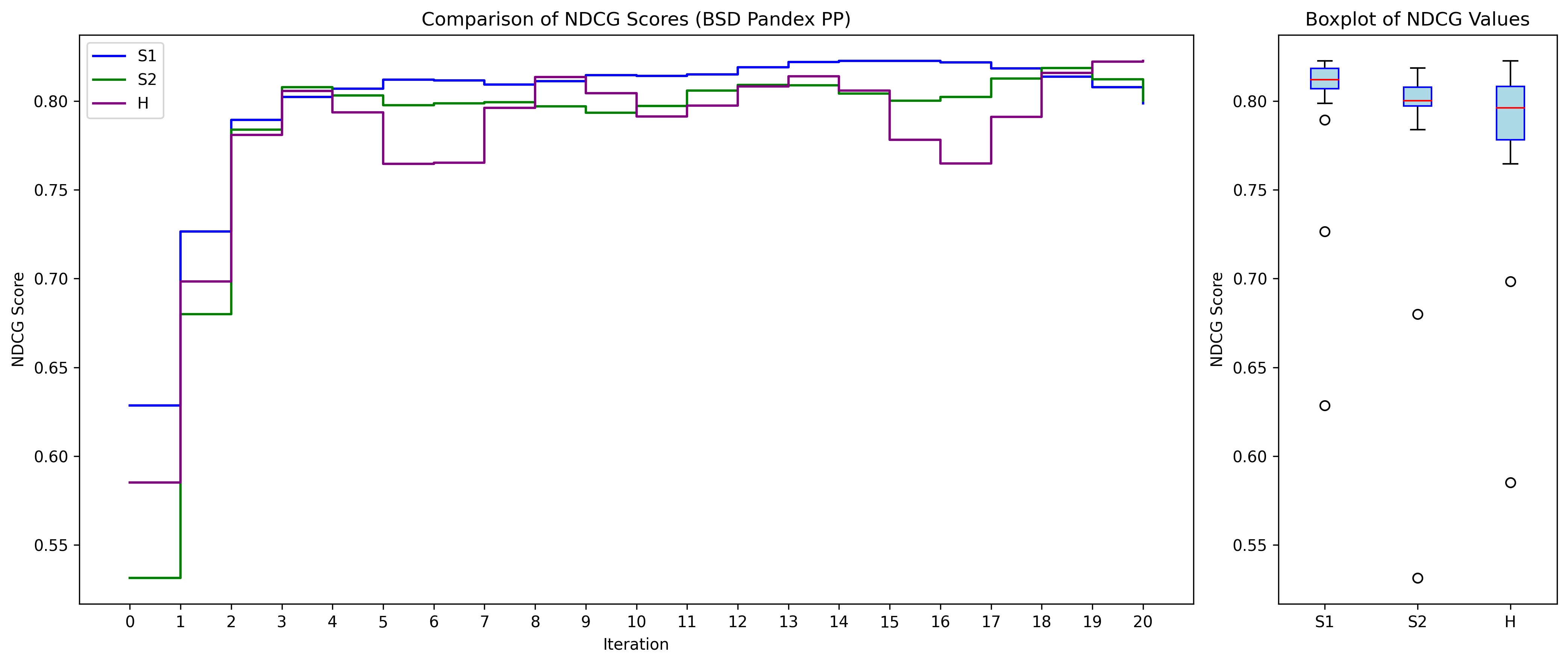} 
         \includegraphics[width=0.4\linewidth]{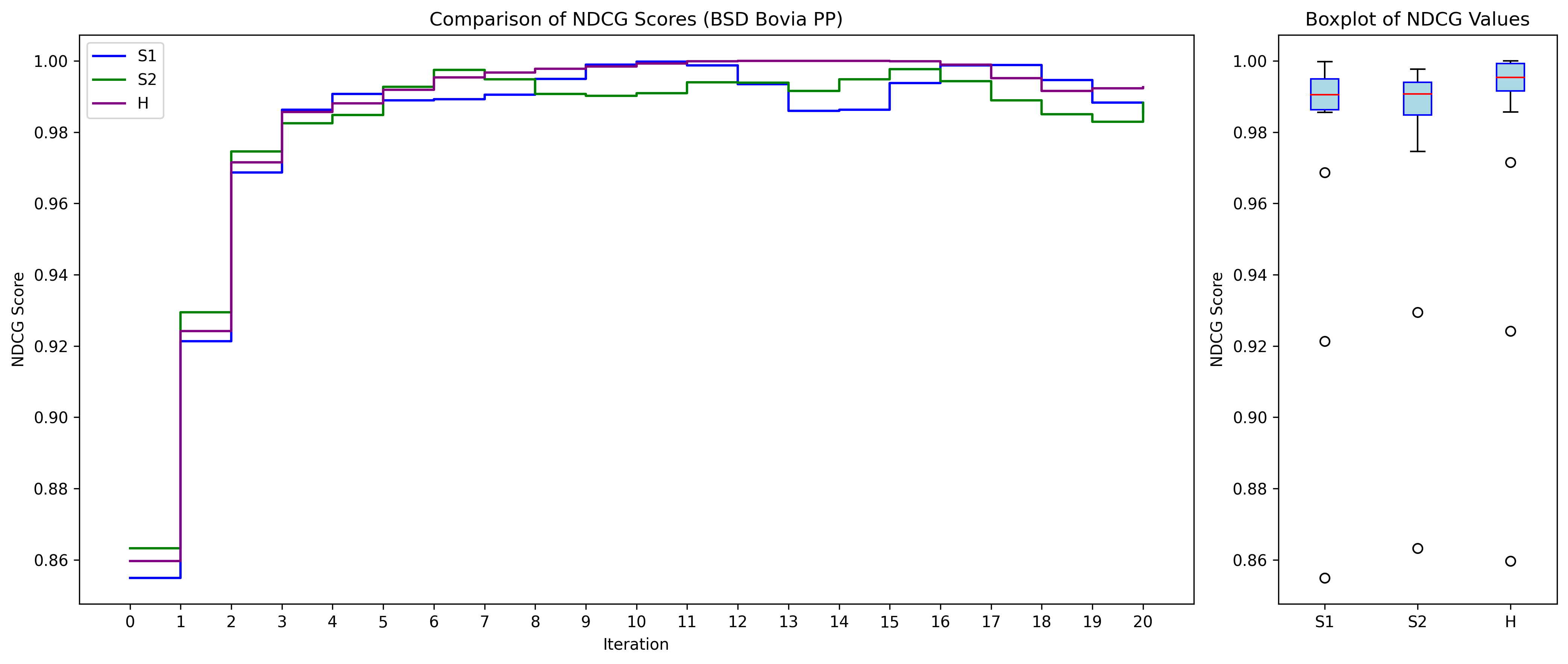} 
        
       \includegraphics[width=0.4\linewidth]{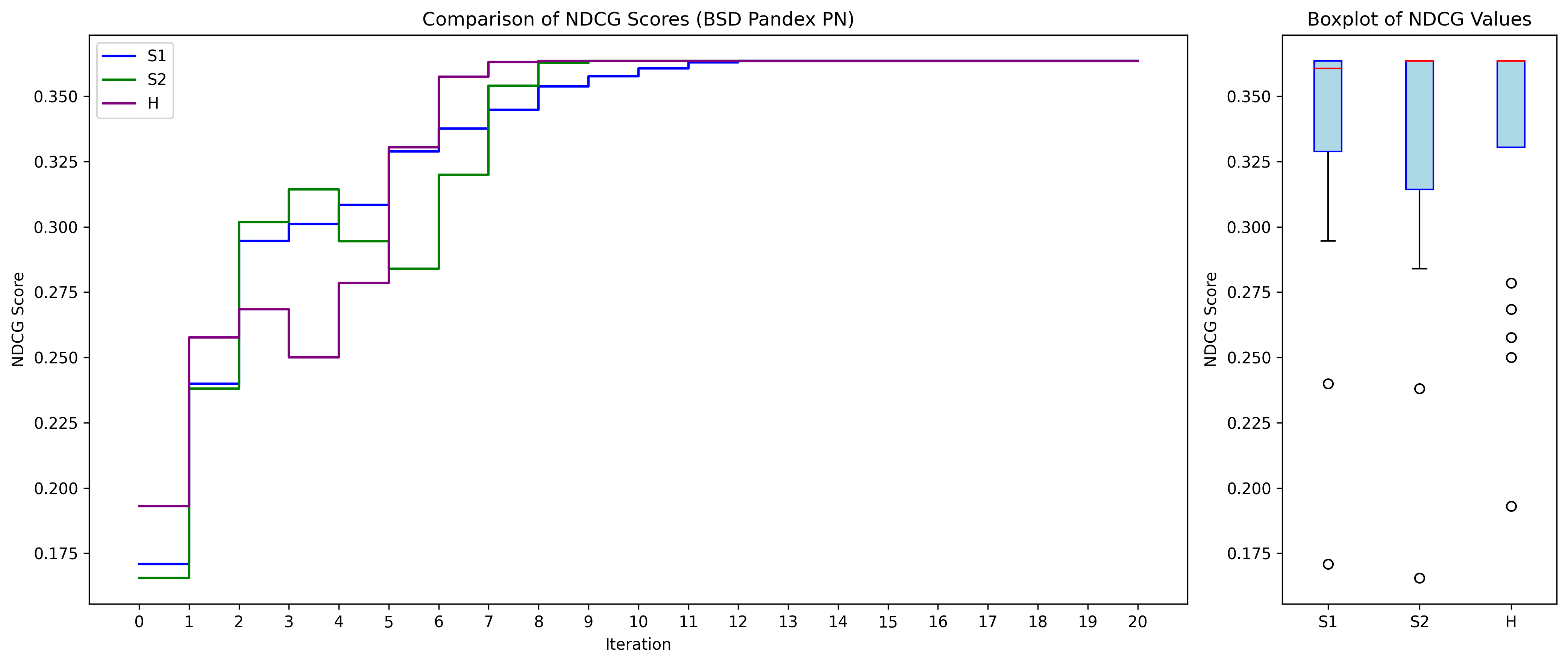} 
         \includegraphics[width=0.4\linewidth]{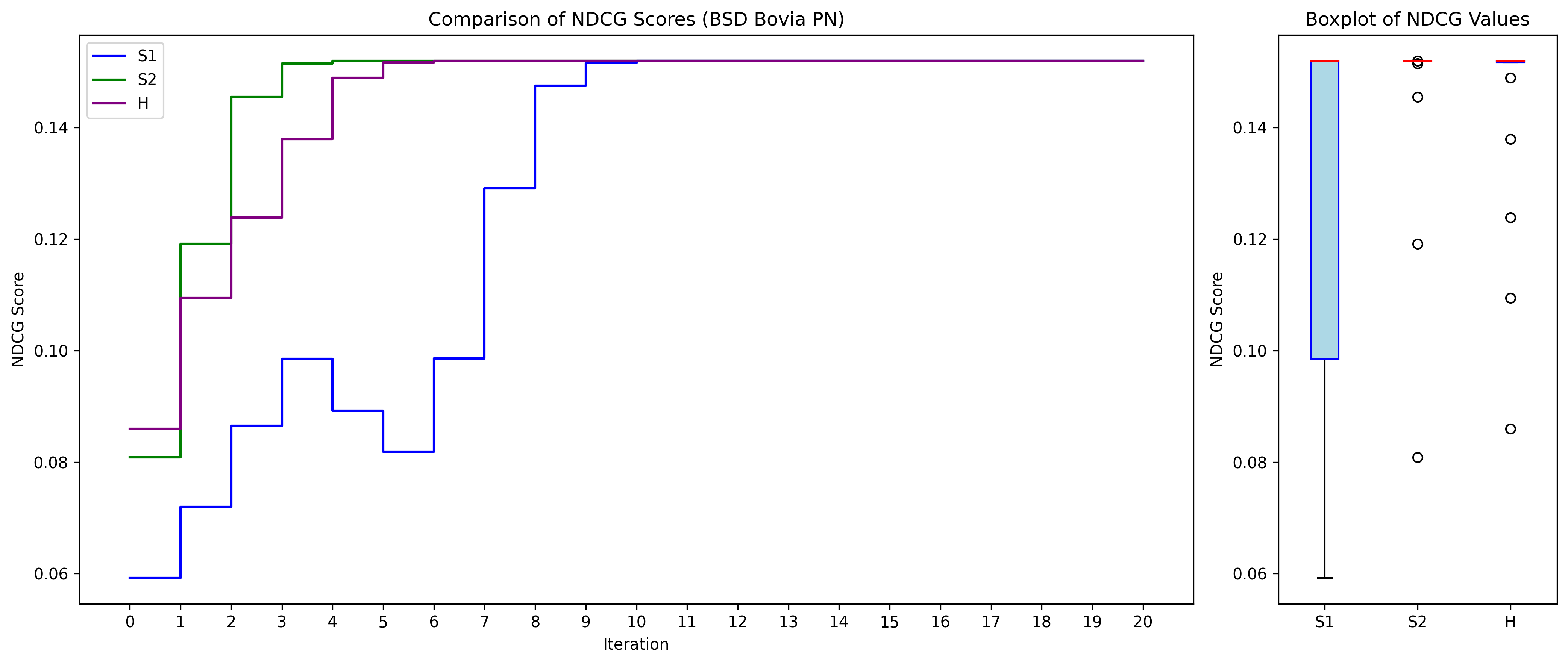}

    \caption{Comparison of nDCG scores across active learning iterations for three similarity search strategies (S1, S2, and H) on the BSD datasets. The left column represents the Pandex attack scenario, while the right column represents the Bovia attack scenario. Each row corresponds to a different subdataset: PA (ProcessALL), PE (ProcessEvent), PX (ProcessExec), PP (ProcessParent), and PN (ProcessNetflow). The line plots (left side of each subplot) illustrate the evolution of nDCG scores, while the boxplots (right side of each subplot) summarize the overall distribution of nDCG scores, highlighting variability and accuracy.
    }
    
    \label{fig:bsdPandexBoviaSAL}
\end{figure*}

Figure~\ref{fig:bsdPandexBoviaSAL} presents the variation of nDCG scores over active learning iterations for the BSD datasets, across two attack scenarios: Pandex (left) and Bovia (right). Each row corresponds to a different sub-dataset (PA, PE, PX, PP, PN), allowing for a comparative analysis of how well each strategy enhances anomaly detection through active learning.

Across both attack scenarios, the nDCG scores exhibit an overall increasing trend, indicating that active learning effectively improves anomaly ranking performance as more informative samples are selected. Some datasets, such as PX, PP, and PN, exhibit early-stage fluctuations due to the sparsity of labeled anomalies, causing temporary drops before stabilizing in later iterations.

For the Pandex attack scenario, strategy H achieves the highest nDCG scores in most sub-datasets, including PA, PE, and PX, demonstrating faster convergence and more stable improvements over iterations. In the PA dataset, strategy H reaches approximately 0.85 after just three iterations. More specifically, in PA (ProcessALL), strategies S1 and S2 reach nDCG values of approximately 0.90 after thirteen iterations. In PE (ProcessEvent), strategy H peaks at an nDCG of approximately 0.85, slightly ahead of S2 and S1. In PX (ProcessExec), H again leads in the final iterations with an nDCG of approximately 0.87. For PP (ProcessParent) and PN (ProcessNetflow), strategies S1, S2, and H perform competitively, reaching maximum nDCG values of approximately 0.80 and 0.36, respectively.

Under the Bovia attack scenario, strategy H consistently achieves the highest nDCG scores in PA, PE, and PX, confirming its effectiveness across different datasets. In PA, H reaches an nDCG of approximately 1.0, indicating perfect anomaly ranking. In PE, H achieves an nDCG of approximately 0.98, with S2 following closely. In PX, H attains an nDCG of approximately 0.97, significantly improving upon initial values. For PP and PN, strategies S1 and S2 are more competitive, peaking at nDCG values of approximately 0.99 and 0.93, respectively.

A comparison between the two attack scenarios shows that Bovia generally exhibits higher final nDCG scores across all sub-datasets, suggesting that the proposed framework is more effective in refining anomaly detection under this attack scenario. Pandex, while showing consistent improvement over iterations, achieves slightly lower maximum nDCG scores, indicating that anomalies in this scenario may be more complex or harder to rank correctly.

Overall, strategy H emerges as the dominant approach, particularly in PA, PE, and PX, achieving the highest nDCG scores in both Pandex (0.89, 0.85, 0.87) and Bovia (1.0, 0.98, 0.97). Strategy S2 remains competitive in PP and PN, achieving nDCG values of approximately 0.36 for Pandex and 0.99 for Bovia. Across all sub-datasets, the Bovia attack scenario leads to better nDCG scores, highlighting that similarity search strategies may be more effective in this setting.

\subsection{Windows Operating System:}

\begin{figure*}[th!]
    \centering
  \includegraphics[width=0.45\linewidth]{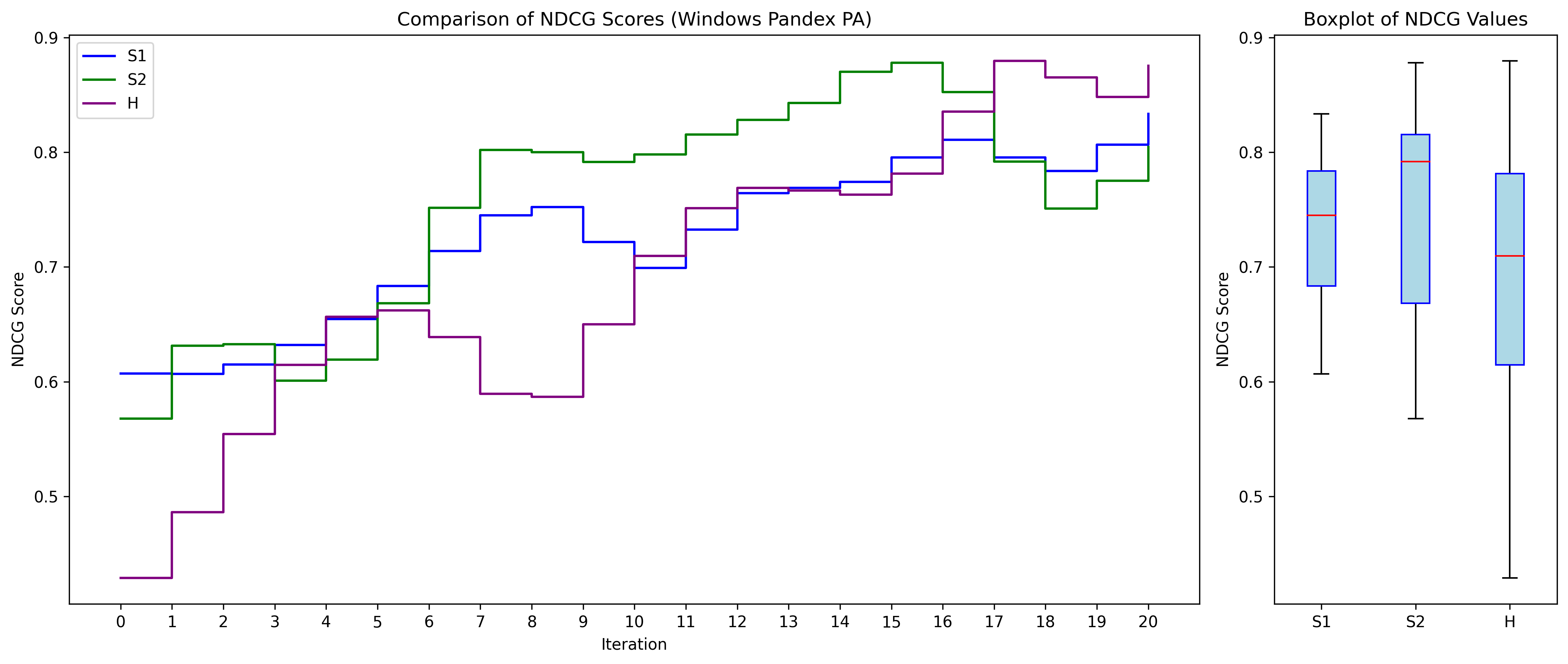} 
         \includegraphics[width=0.45\linewidth]{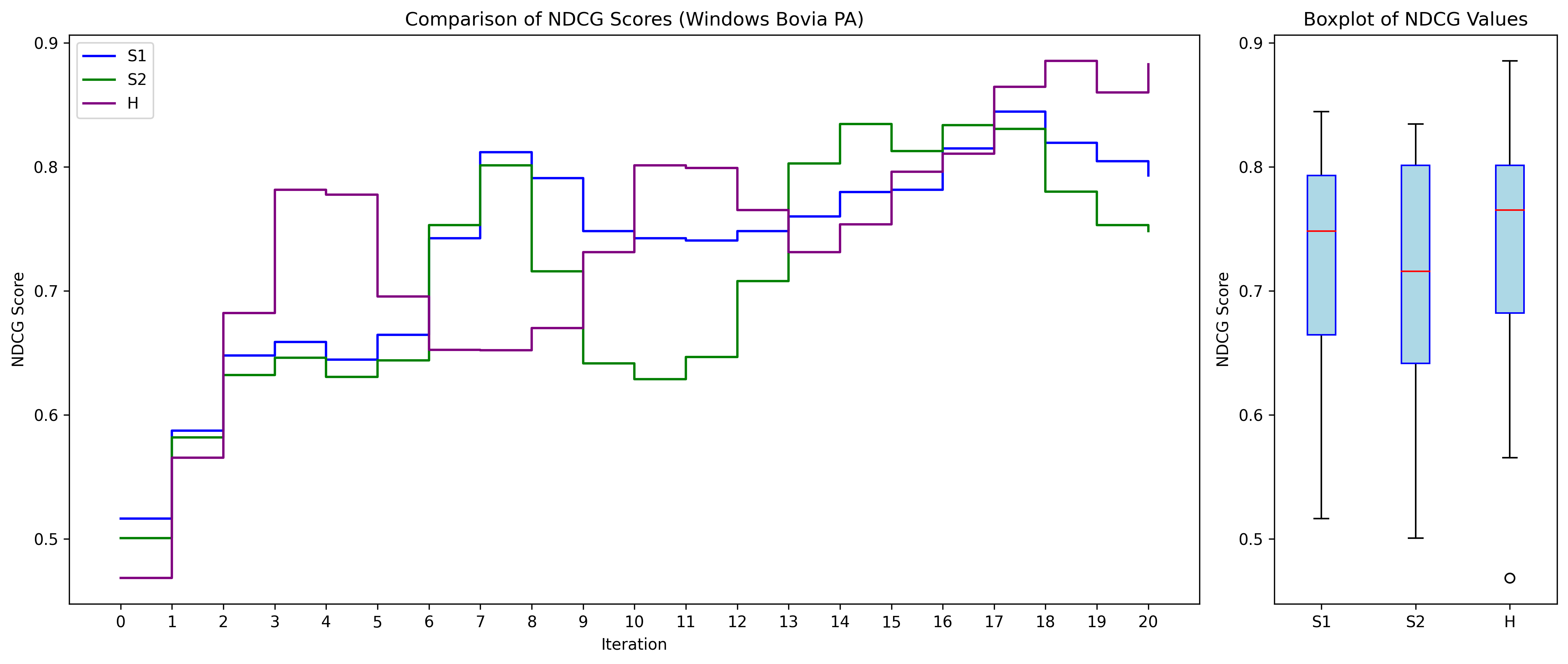} 
        
         \includegraphics[width=0.45\linewidth]{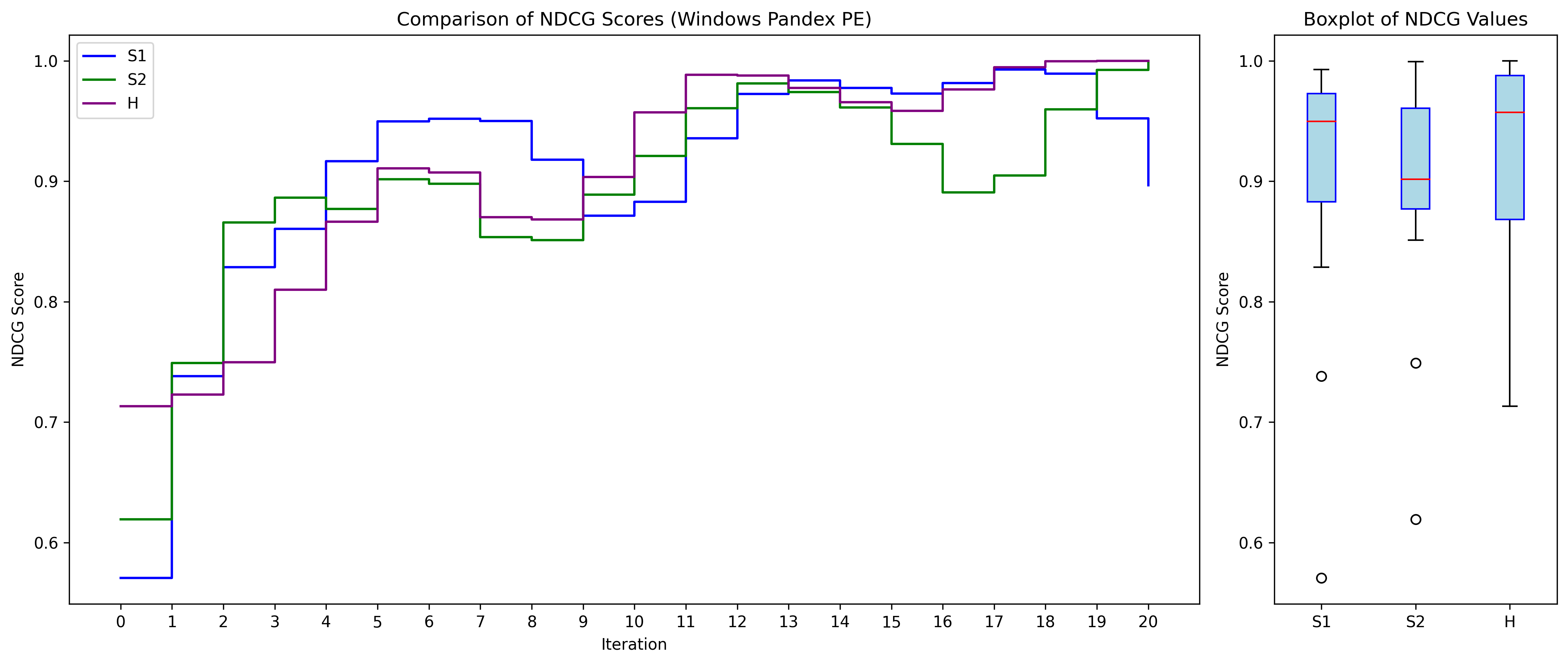} 
         \includegraphics[width=0.45\linewidth]{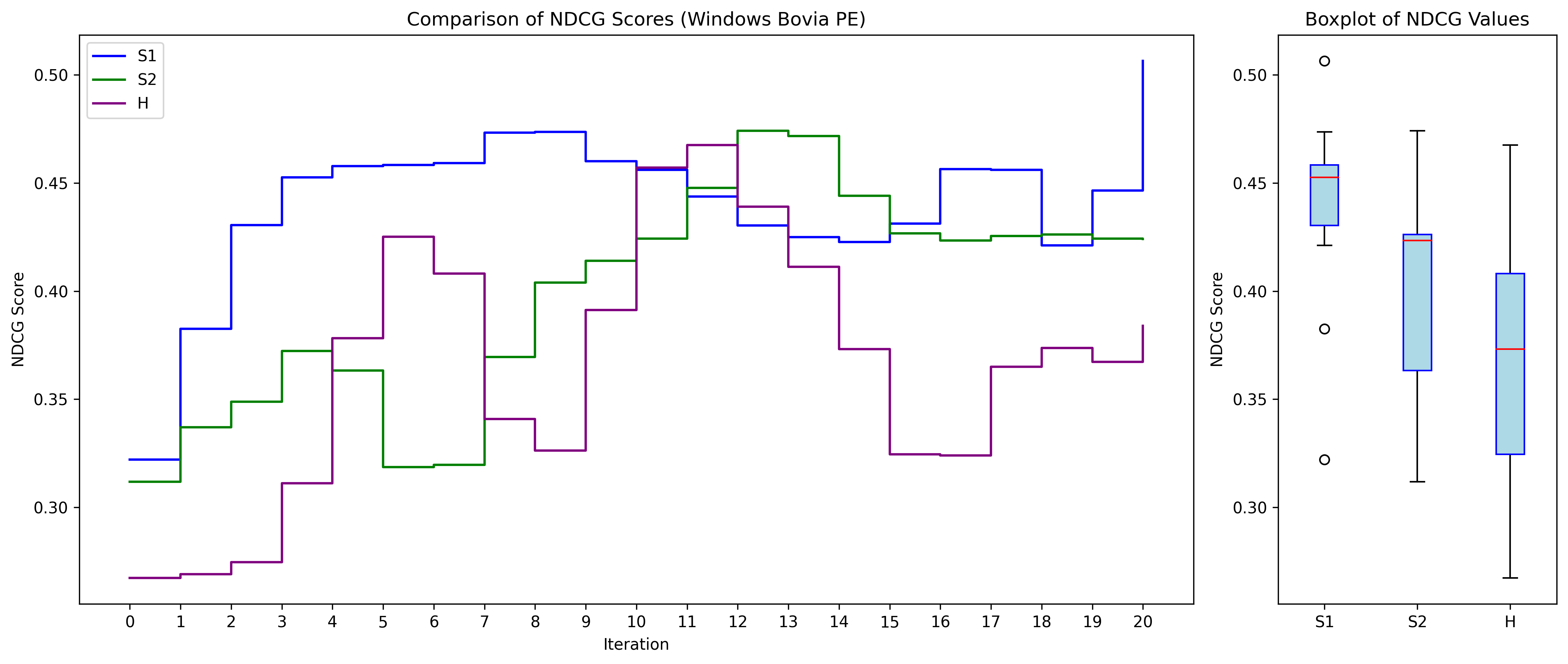} 
        
         \includegraphics[width=0.45\linewidth]{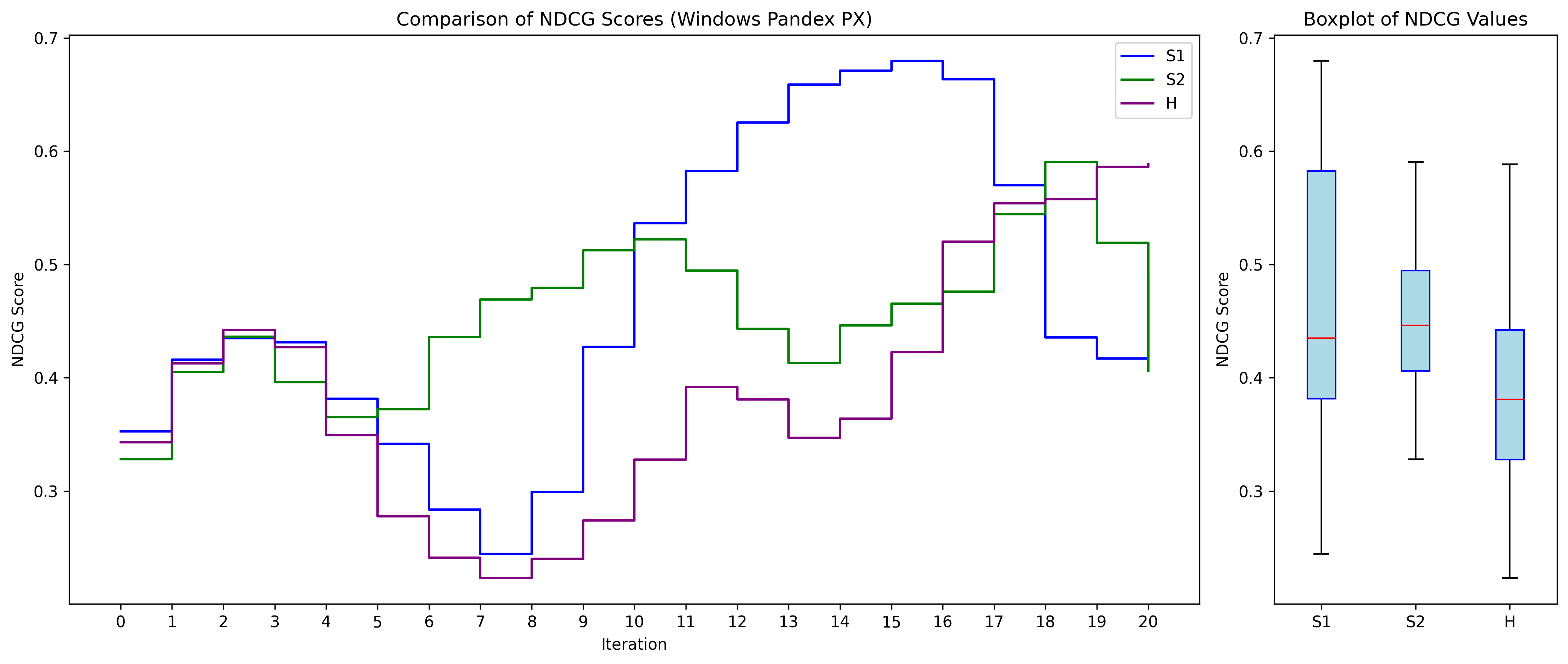} 
         \includegraphics[width=0.45\linewidth]{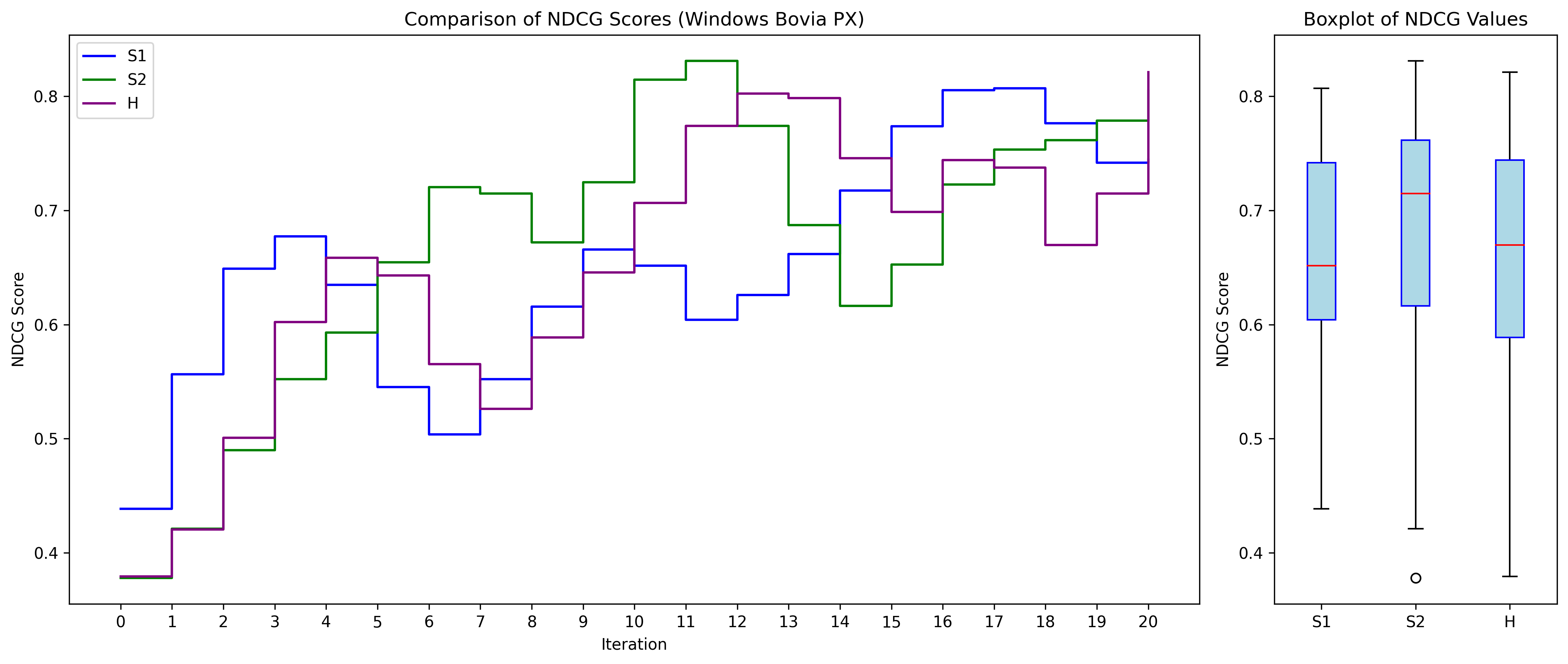} 
        
         \includegraphics[width=0.45\linewidth]{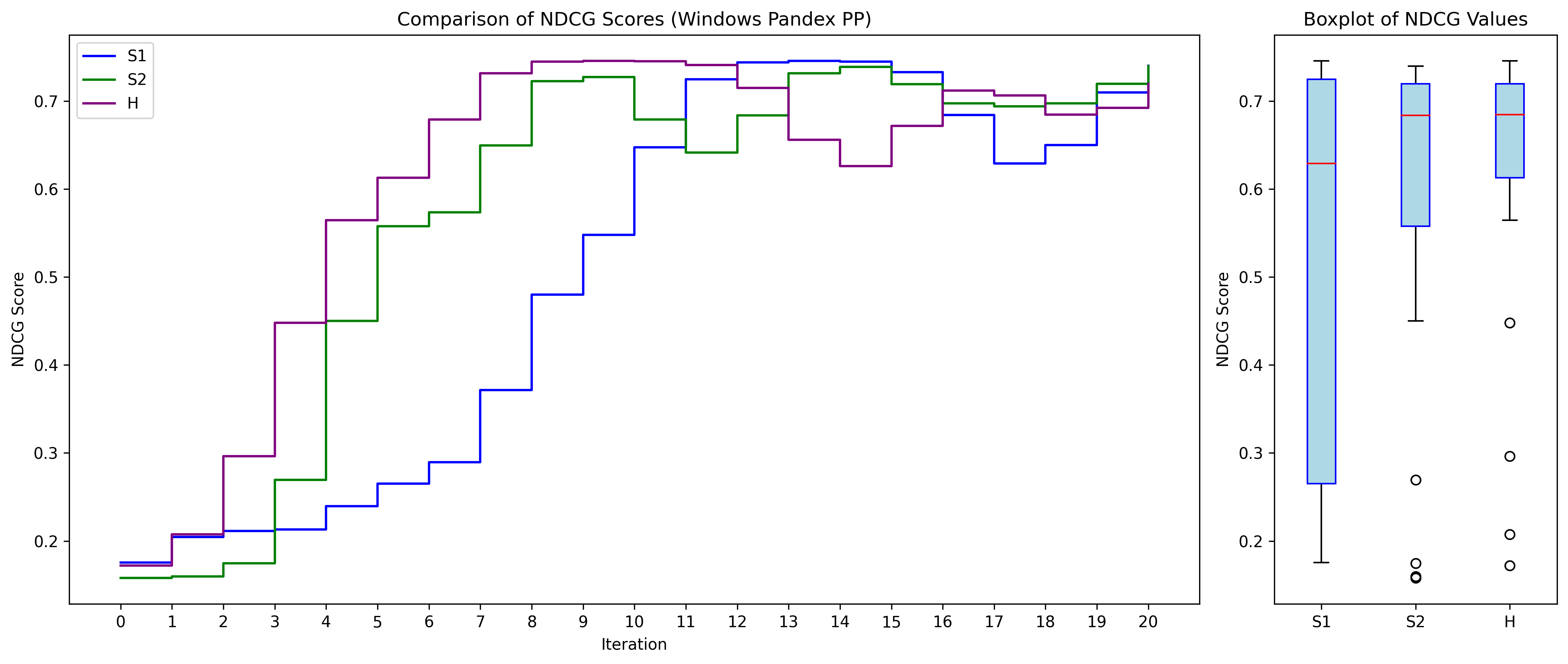} 
         \includegraphics[width=0.45\linewidth]{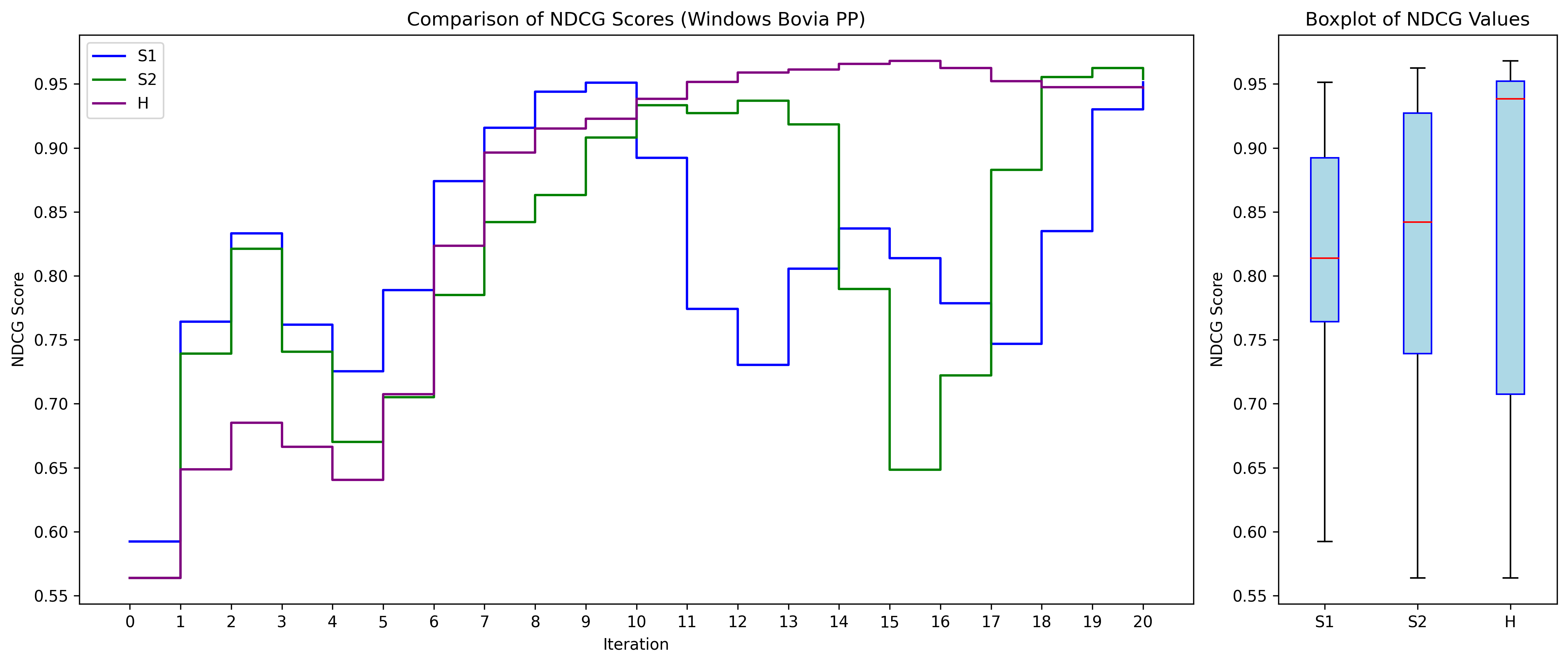} 
        
         \includegraphics[width=0.45\linewidth]{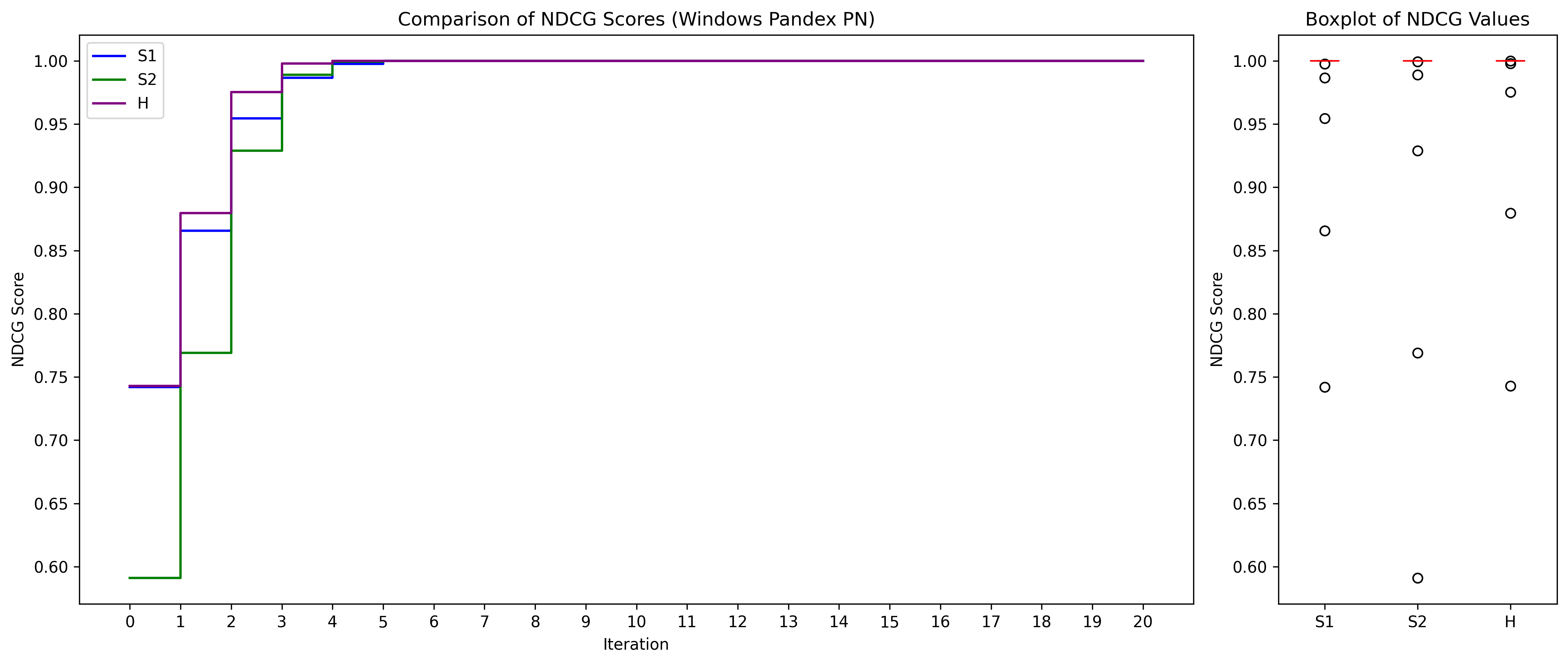} 
         \includegraphics[width=0.45\linewidth]{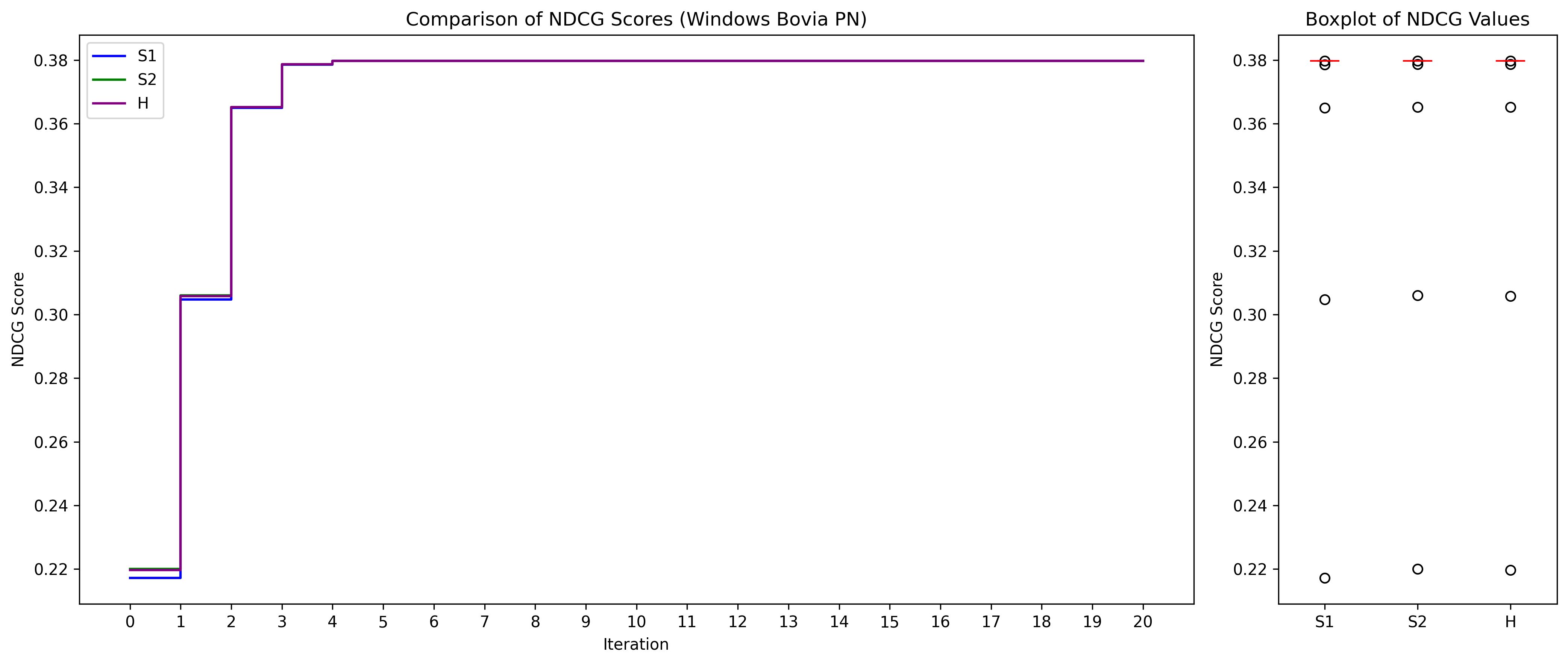} 
    
    \caption{Comparison of nDCG scores across active learning iterations for three similarity search strategies (S1, S2, and H) on the Windows datasets. The left column represents the Pandex attack scenario, while the right column represents the Bovia attack scenario. Each row corresponds to a different subdataset: PA (ProcessALL), PE (ProcessEvent), PX (ProcessExec), PP (ProcessParent), and PN (ProcessNetflow). The line plots (left side of each subplot) illustrate the evolution of nDCG scores, while the boxplots (right side of each subplot) summarize the overall distribution of nDCG scores, highlighting variability and accuracy.
    }
    
    \label{fig:windowsPandexBoviaSAL}
\end{figure*}

Figure~\ref{fig:windowsPandexBoviaSAL} presents the nDCG score variation across active learning iterations for the Windows datasets under the Pandex (left) and Bovia (right) attack scenarios.

Across all datasets, nDCG scores tend to increase as active learning progresses, indicating that the framework improves its ability to rank true anomalies over successive iterations. However, some datasets exhibit significant fluctuations, especially PP and PX, which is likely attributable to data sparsity or more challenging anomaly patterns inherent to these views.

In the Pandex scenario, the highest nDCG score reaches 0.9 in PA and achieves perfect ranking (nDCG = 1.0) in both PE and PN, demonstrating the effectiveness of active learning in fully identifying true positives in these datasets. In the Bovia scenario, the highest nDCG score reaches 0.9 in PA and approximately 0.95 in PP, confirming that the model is able to rank anomalies with high accuracy under this attack setting as well.

Regarding the relative performance of similarity search strategies, the H strategy generally outperforms others in the Pandex scenario, achieving high scores particularly in PA and PE, while S2 remains competitive in PX. Under the Bovia scenario, the H strategy achieves stable and consistently high performance across most sub-datasets, highlighting its robustness to noise and variability in Windows system behavior.

Overall, the Windows datasets exhibit strong improvements in anomaly detection performance through active learning. In PA and PE, all strategies converge to very high nDCG scores, indicating reliable anomaly ranking. In contrast, PX and PP show more pronounced fluctuations, likely reflecting more complex detection scenarios. Across both Pandex and Bovia attack scenarios, the H strategy emerges as the most effective and stable approach.

%
%
    %
 %
  %
    %
 %
 %
\subsection{Linux Operating System}

\begin{figure*}[th!]
    \centering
  \includegraphics[width=0.45\linewidth]{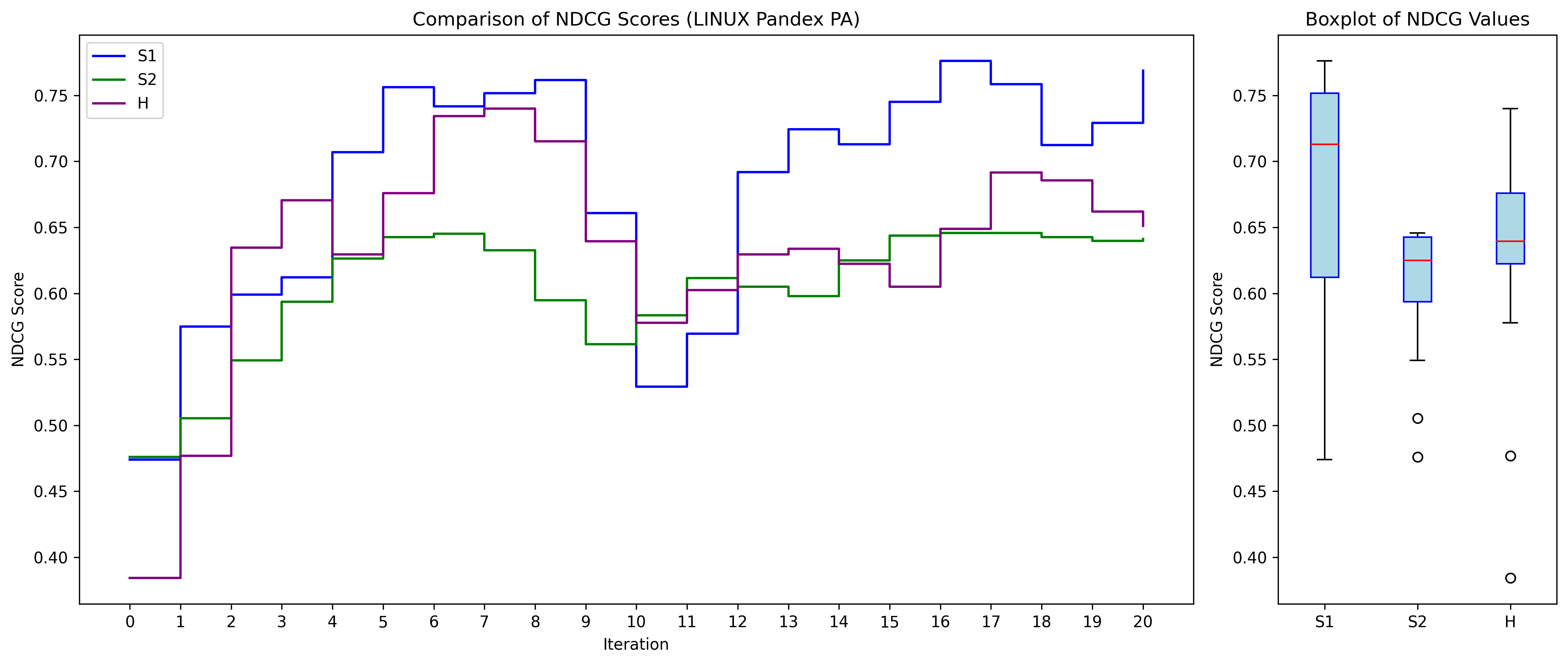} 
         \includegraphics[width=0.45\linewidth]{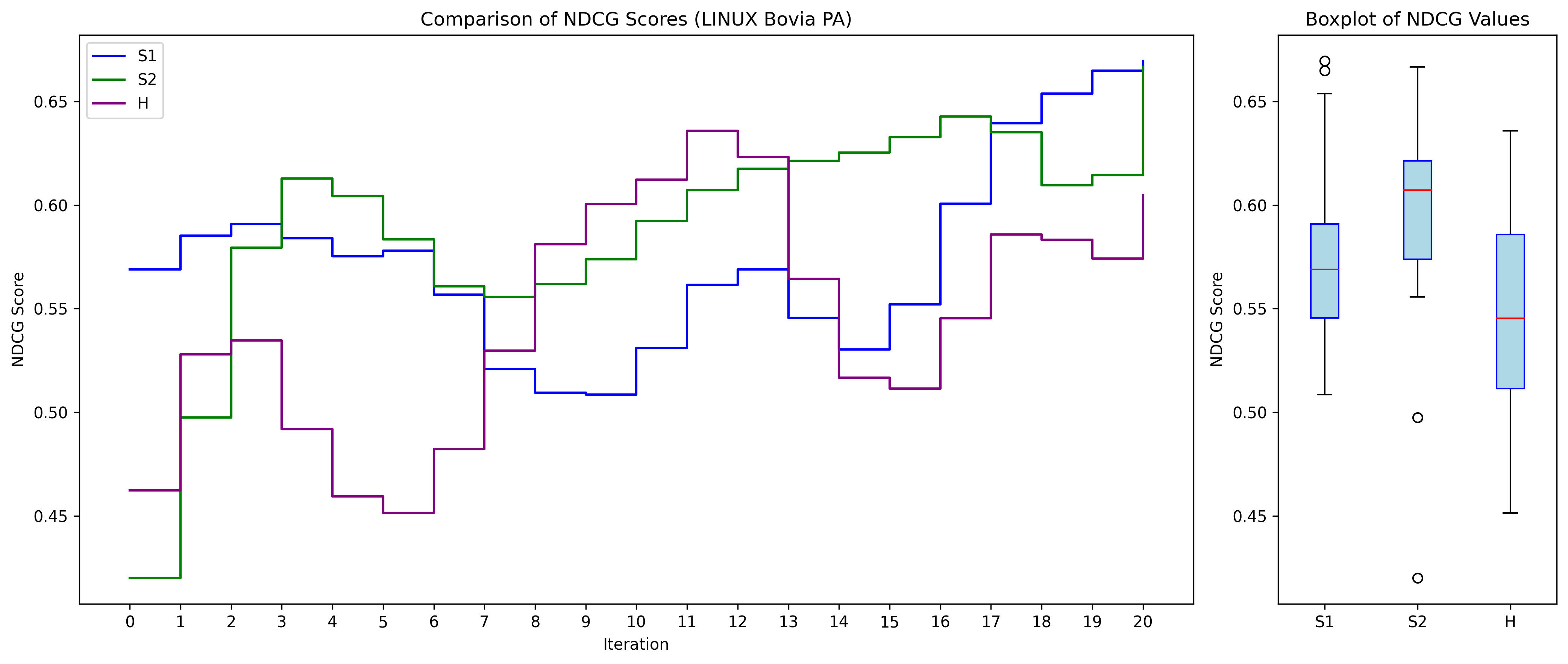} 
        
        \includegraphics[width=0.45\linewidth]{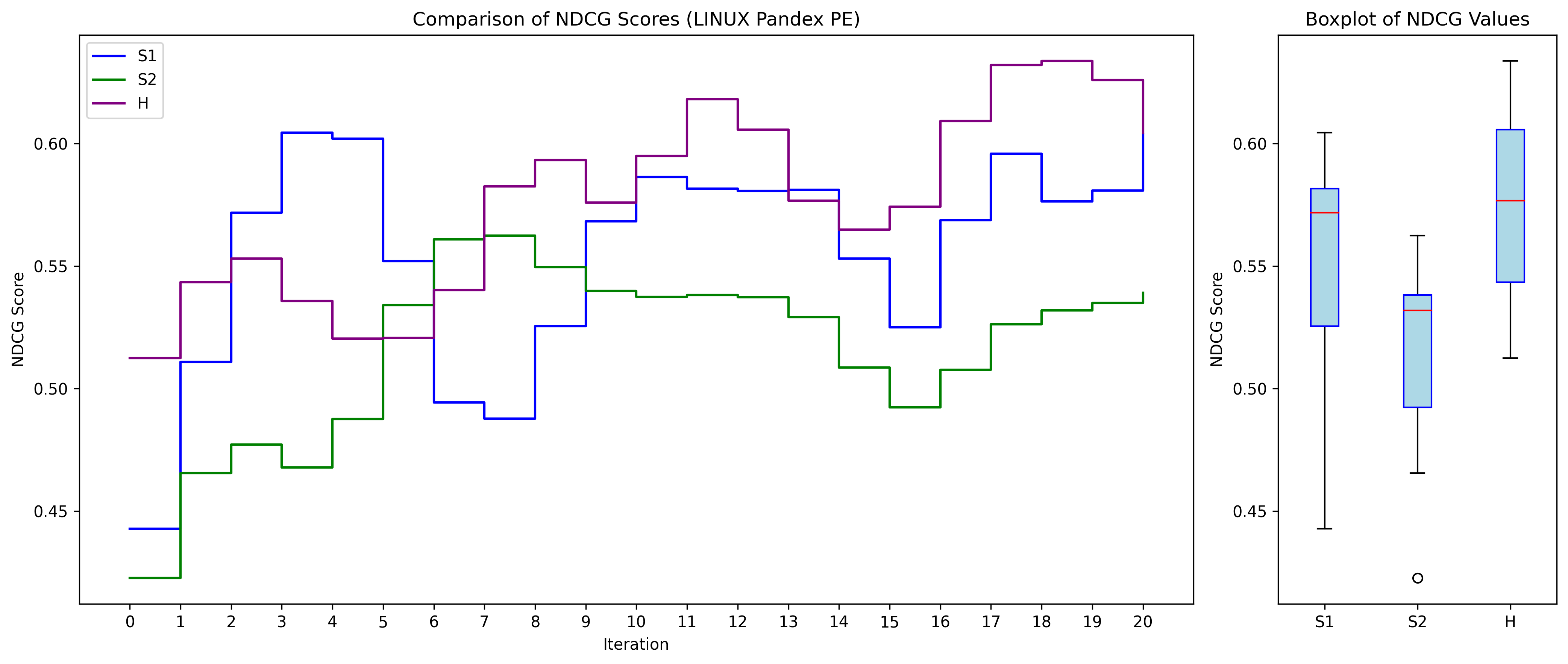} 
         \includegraphics[width=0.45\linewidth]{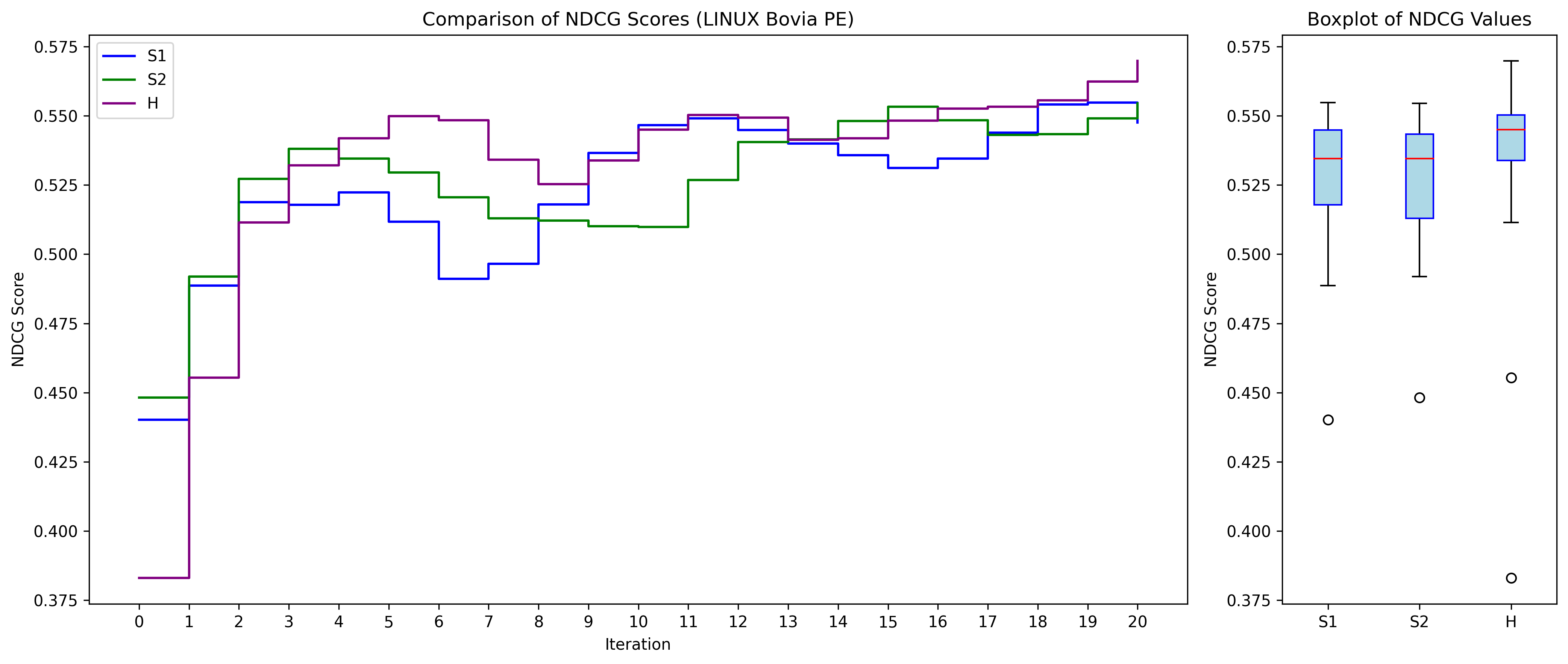} 
        
         \includegraphics[width=0.45\linewidth]{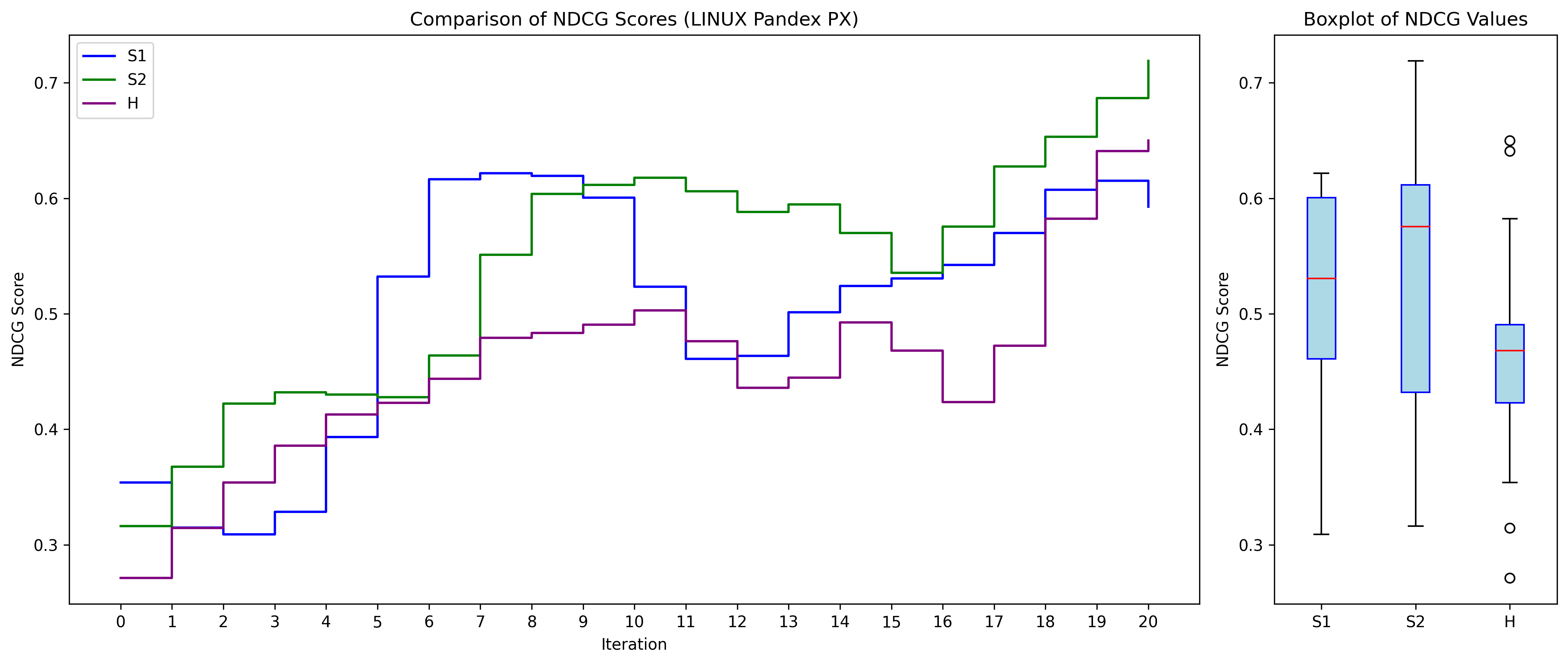} 
         \includegraphics[width=0.45\linewidth]{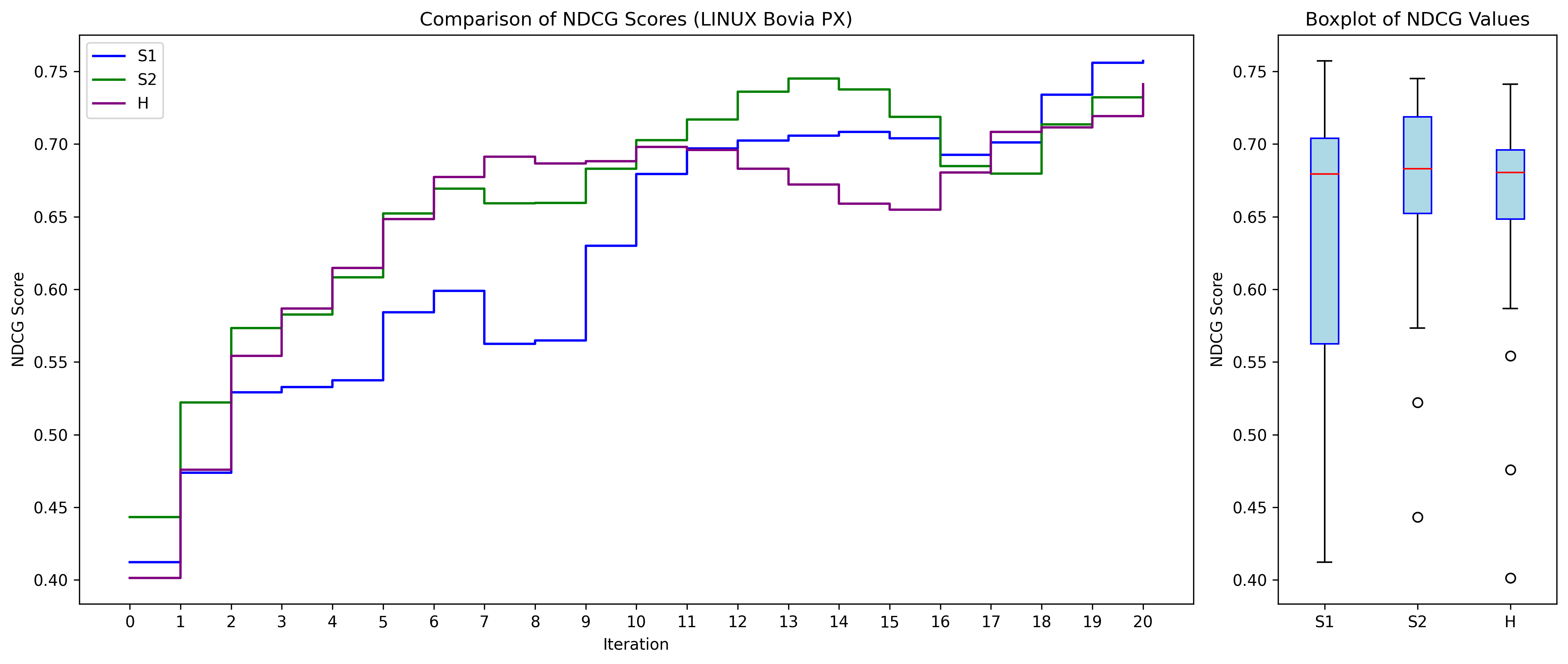} 
        
         \includegraphics[width=0.45\linewidth]{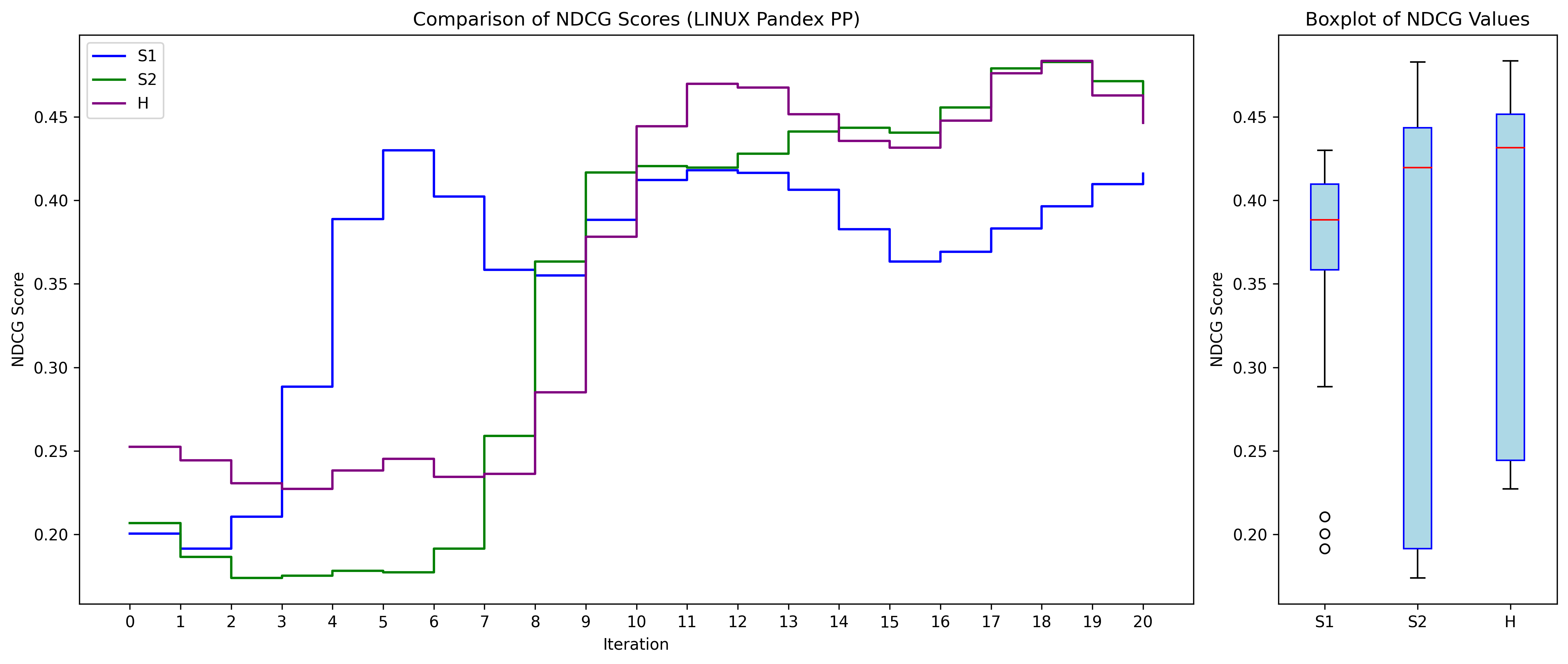} 
         \includegraphics[width=0.45\linewidth]{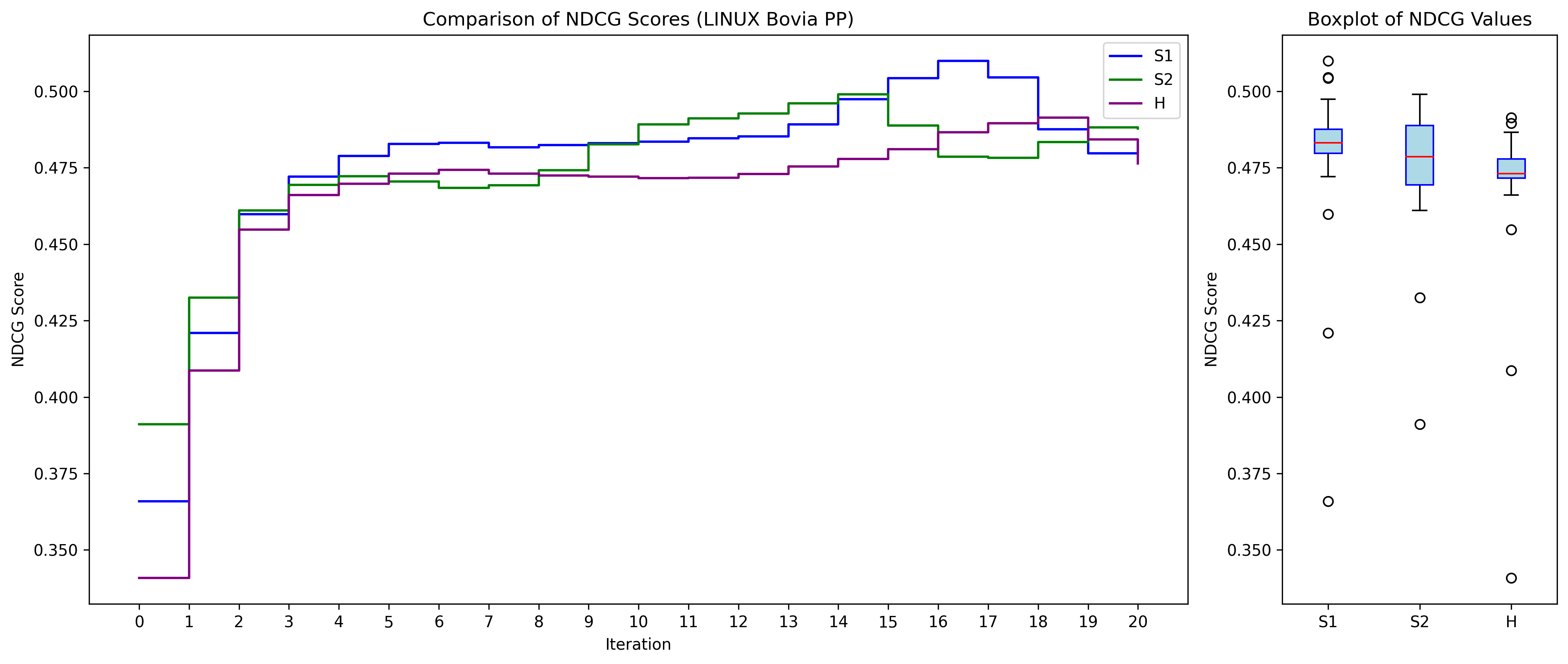} 
        
         \includegraphics[width=0.45\linewidth]{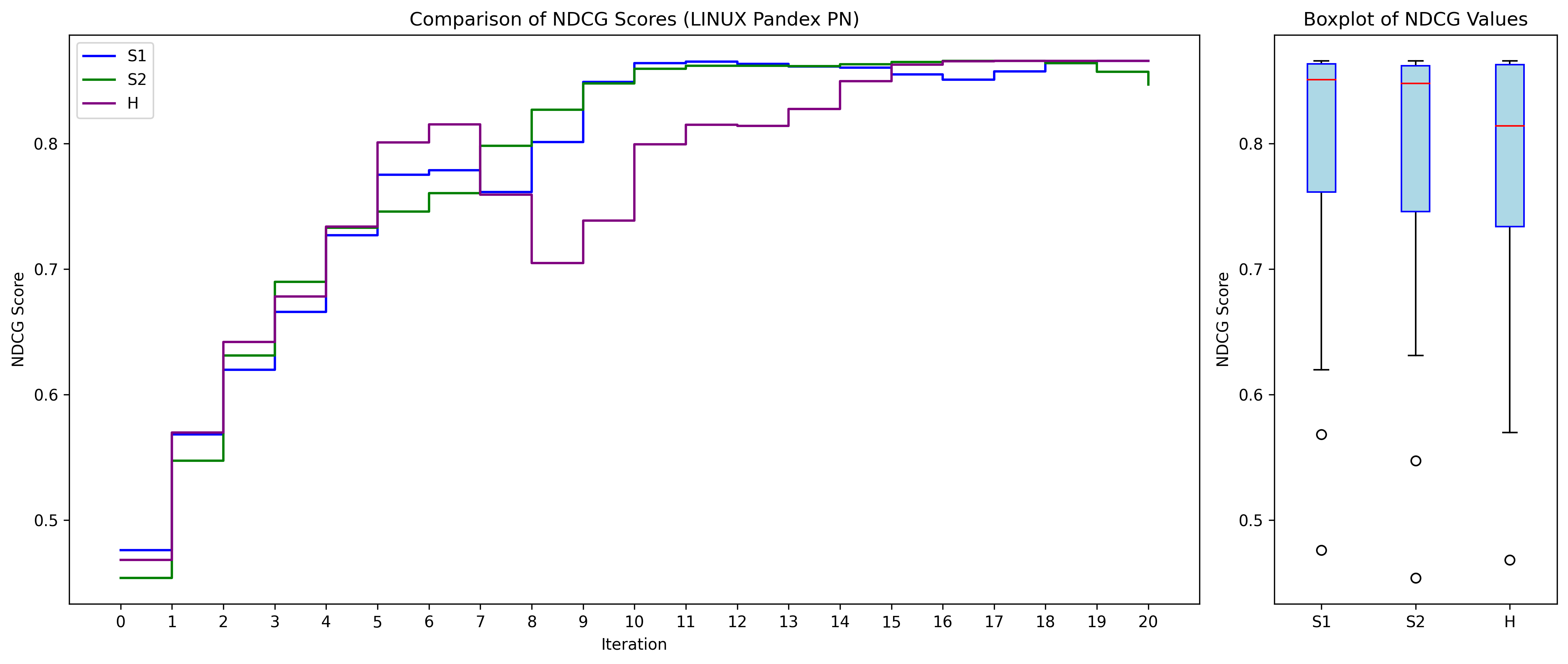} 
         \includegraphics[width=0.45\linewidth]{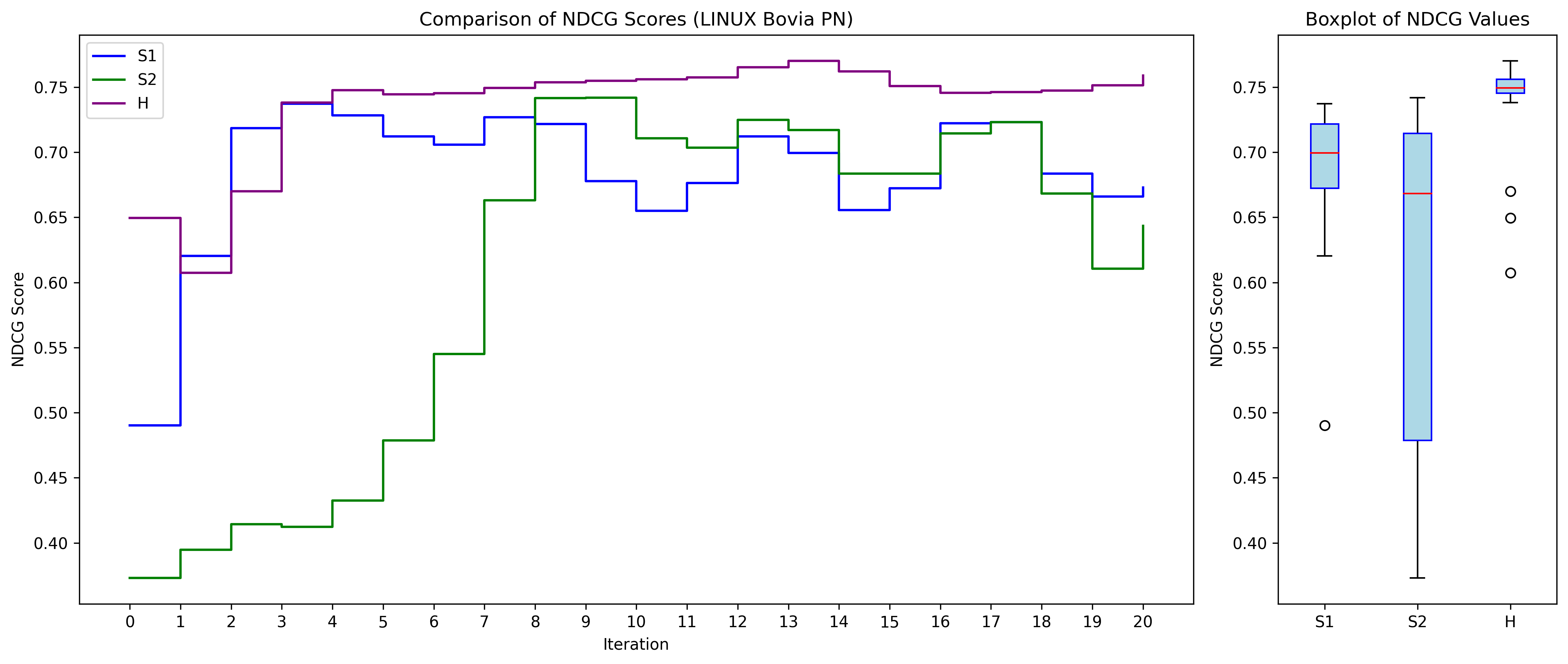} 
     
    \caption{Comparison of nDCG scores across active learning iterations for three similarity search strategies (S1, S2, and H) on the LINUX datasets. The left column represents the Pandex attack scenario, while the right column represents the Bovia attack scenario. Each row corresponds to a different sub-dataset: PA (ProcessALL), PE (ProcessEvent), PX (ProcessExec), PP (ProcessParent), and PN (ProcessNetflow). The line plots (left side of each subplot) illustrate the evolution of nDCG scores, while the boxplots (right side of each subplot) summarize the overall distribution of nDCG scores, highlighting variability and accuracy.
    }
    
    \label{fig:LINUXPandexBoviaSAL}
\end{figure*}

In a similar way, the results in Figure~\ref{fig:LINUXPandexBoviaSAL} present the evolution of nDCG scores across active learning iterations for the Linux datasets under the Pandex and Bovia attack scenarios. Across all Linux datasets, nDCG scores exhibit a steady improvement over active learning iterations, confirming that the iterative learning process helps refine anomaly ranking.

While most datasets show a smooth increasing trend, the PX and PP datasets demonstrate noticeable fluctuations, possibly due to the complex nature of anomalies in these sub-datasets or the sparsity of labeled data. These variations are particularly evident during early iterations, where limited feedback can temporarily affect ranking stability.

In the Pandex scenario, the highest nDCG score reaches approximately 0.87 in the PN dataset, demonstrating strong detection performance across all three strategies. Other sub-datasets, such as PA, PE, and PX, also show substantial improvement. Specifically, PA reaches a peak nDCG of approximately 0.76 with S1, PE reaches around 0.63 with H, and PX attains approximately 0.71 with S2.

Under the Bovia scenario, the highest nDCG scores are observed in PX and PN, reaching approximately 0.76 and 0.75, with S1 and H respectively, while the PP dataset displays more moderate improvements. These results indicate that while anomaly refinement is effective, certain sub-datasets remain more challenging under this attack scenario.

Regarding the best similarity search strategy, in the Pandex scenario, S1, S2, and H achieve the strongest performances in the PP and PA datasets. In contrast, under the Bovia scenario, H provides more stable performance across PN, while S1 shows competitive results in PX.

Overall, the Linux datasets highlight the effectiveness of active learning in refining anomaly detection performance. The PA, PX, and PN datasets benefit the most, achieving some of the highest nDCG scores observed. In contrast, the PP dataset presents more variation, suggesting that anomalies in this sub-dataset are harder to capture consistently. Across both attack scenarios, the H and S1 strategies demonstrate superior performance.

%
    %
  %
   %
    %
 %
 %
\subsection{Android Operating System:}

\begin{figure*}[th!]
    \centering
  \includegraphics[width=0.45\linewidth]{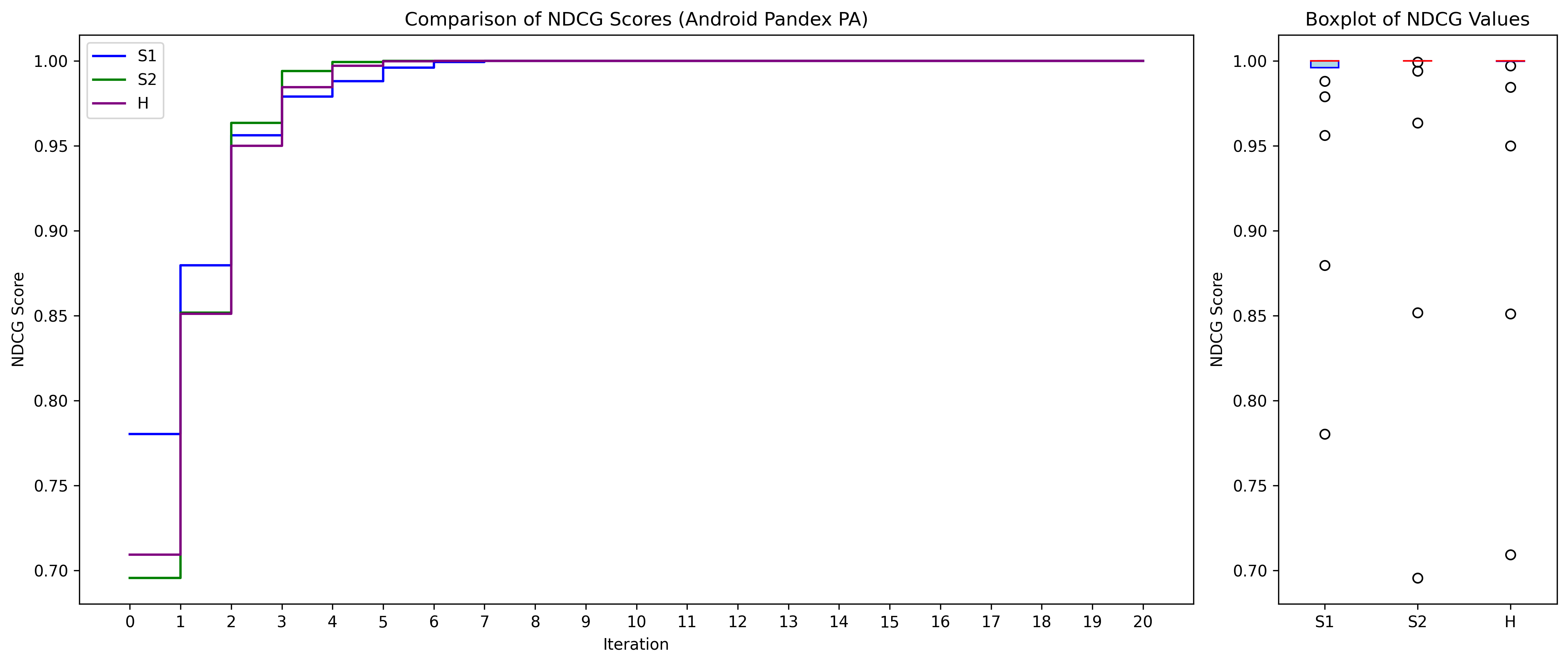} 
         \includegraphics[width=0.45\linewidth]{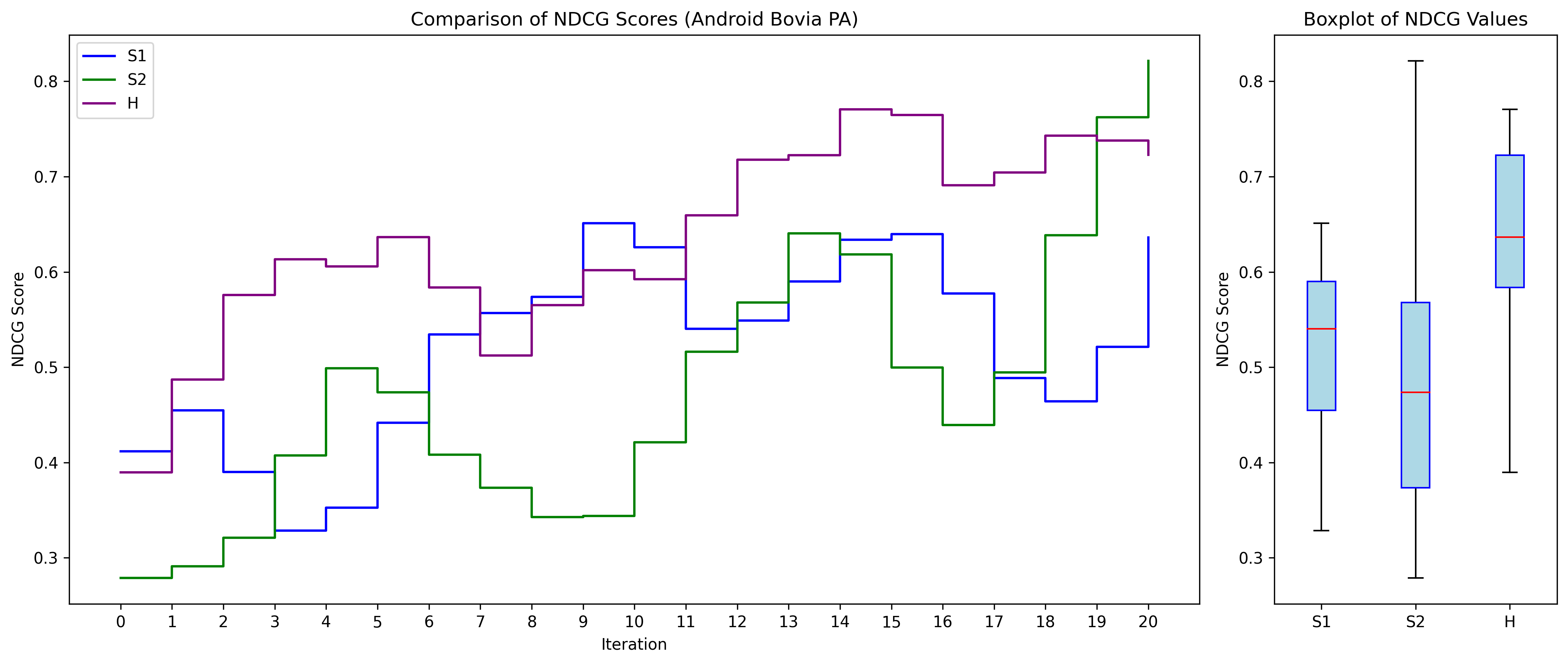} 
        
         \includegraphics[width=0.45\linewidth]{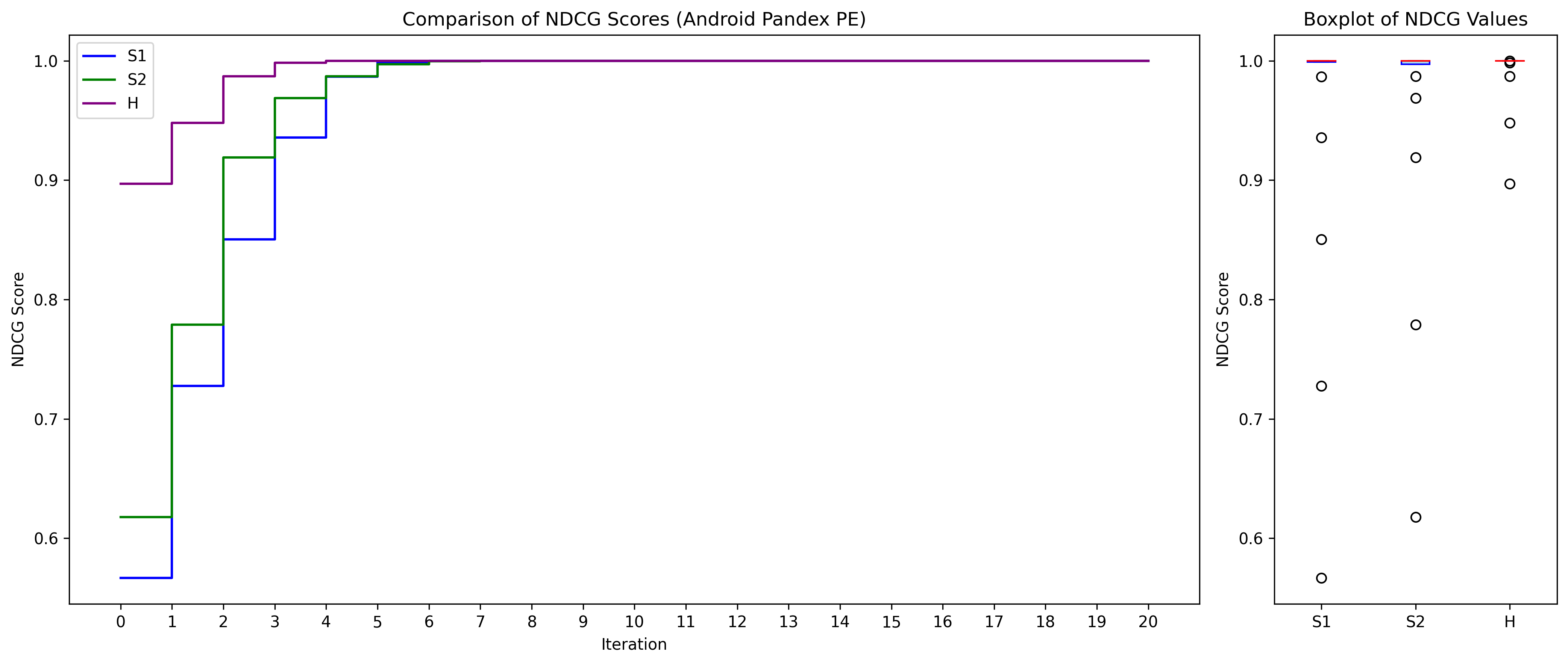} 
         \includegraphics[width=0.45\linewidth]{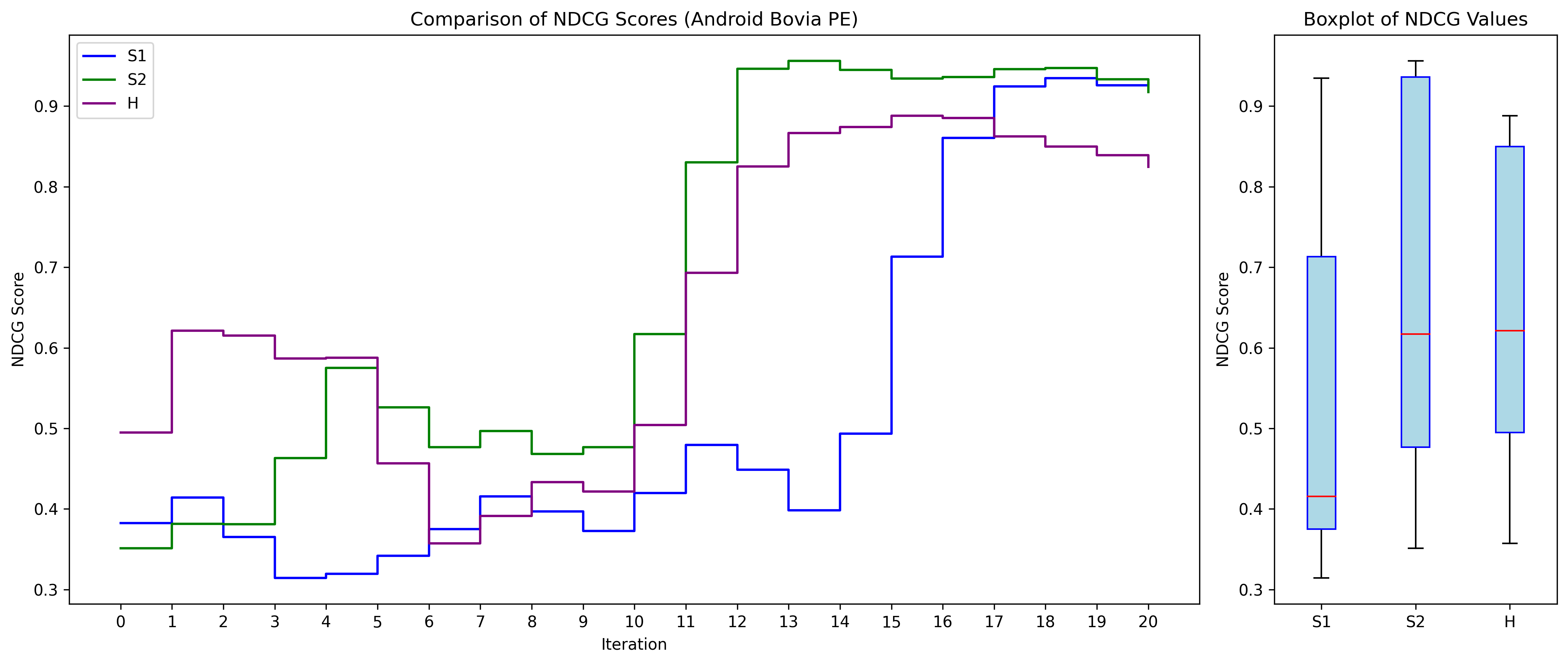} 
        
         \includegraphics[width=0.45\linewidth]{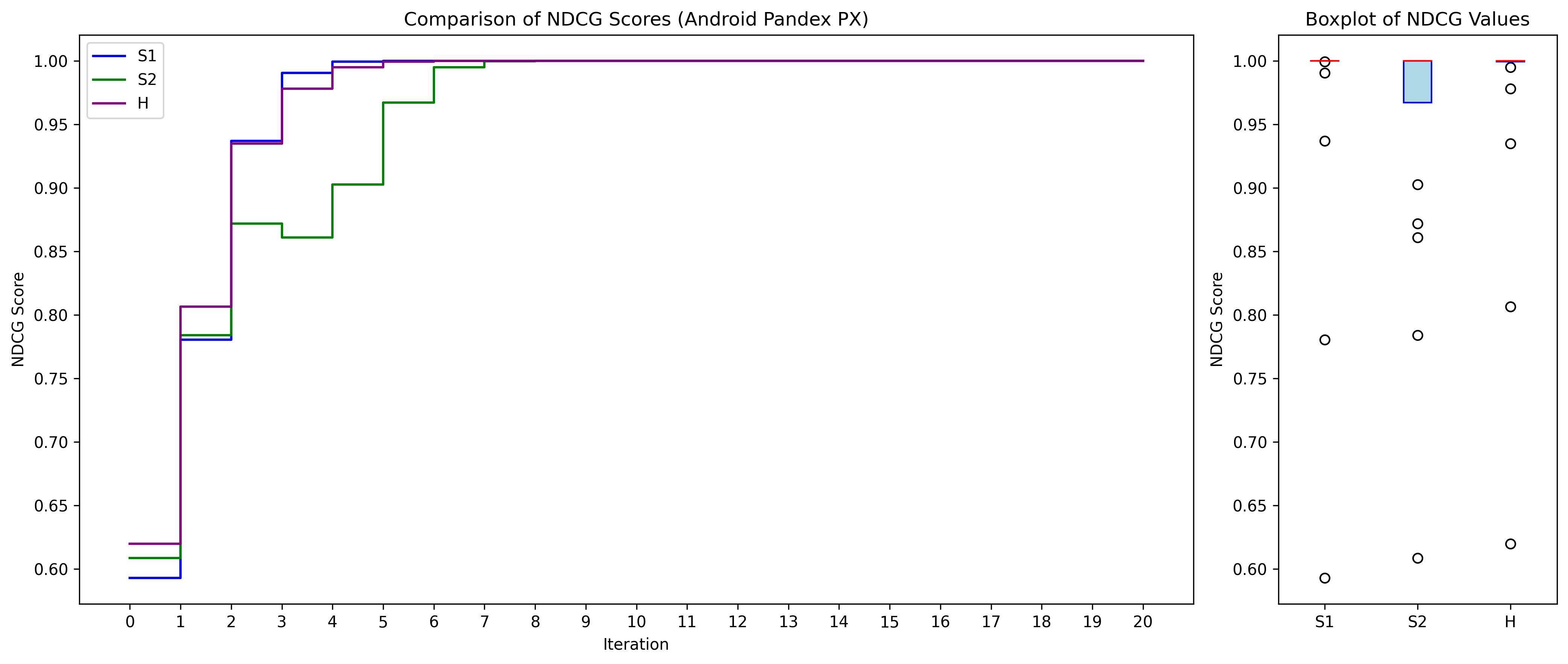} 
         \includegraphics[width=0.45\linewidth]{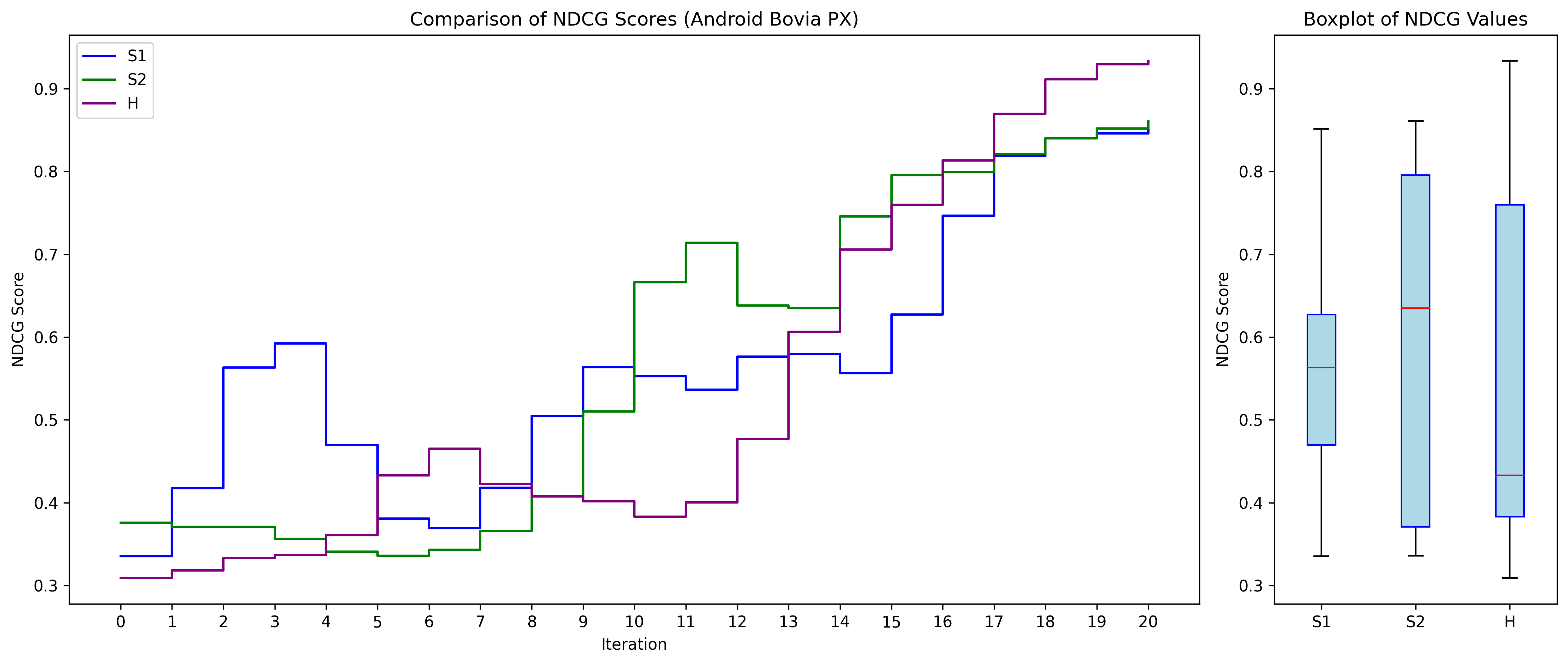} 
        
        
         \includegraphics[width=0.45\linewidth]{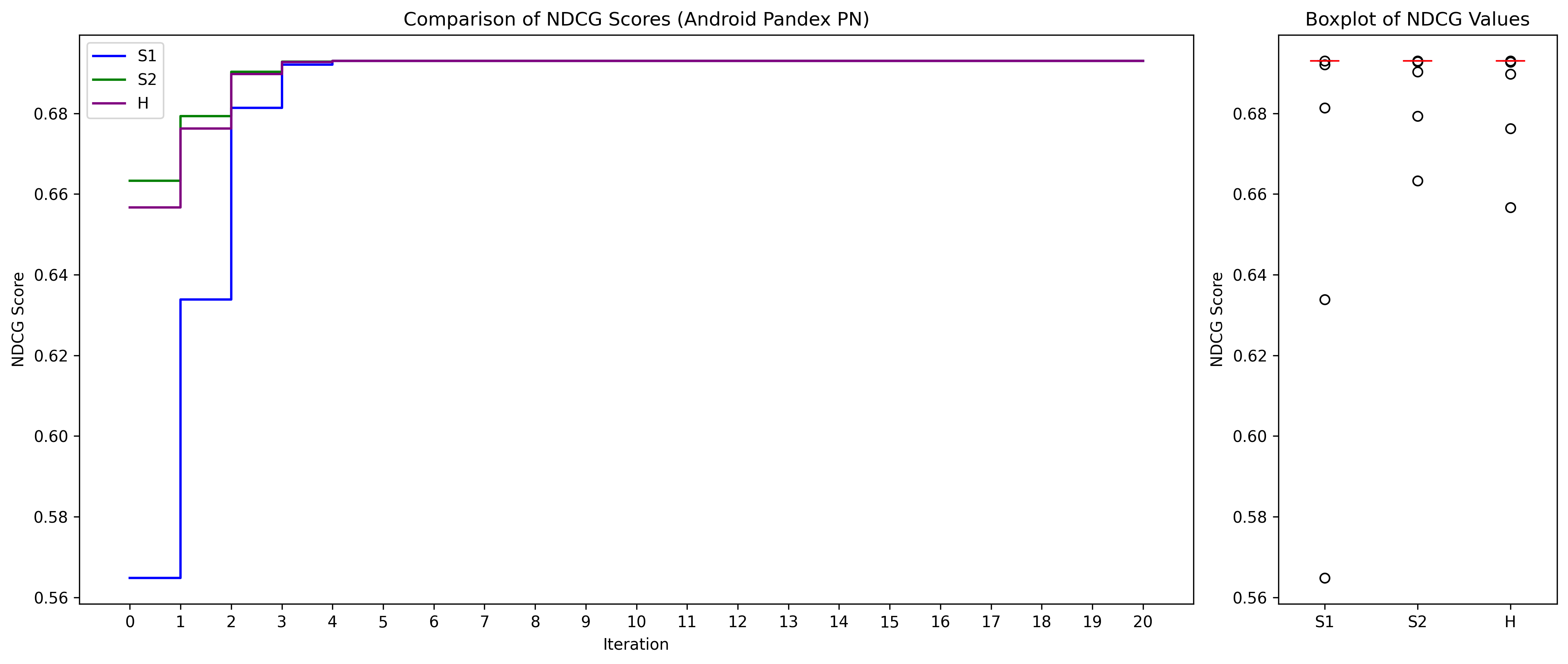} 
         \includegraphics[width=0.45\linewidth]{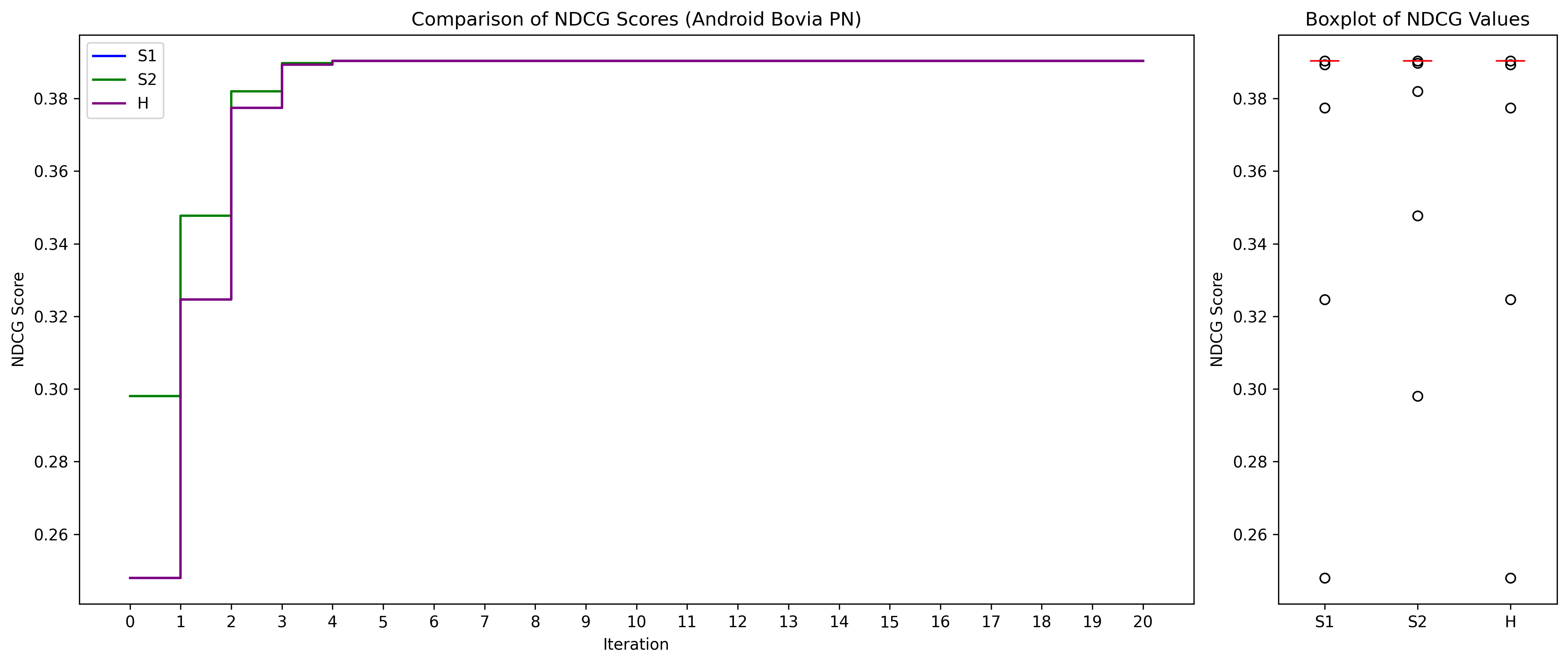} 
    
    \caption{Comparison of nDCG scores across active learning iterations for three similarity search strategies (S1, S2, and H) on the ANDROID datasets. The left column represents the Pandex attack scenario, while the right column represents the Bovia attack scenario. Each row corresponds to a different sub-dataset: PA (ProcessALL), PE (ProcessEvent), PX (ProcessExec), PP (ProcessParent), and PN (ProcessNetflow). The line plots (left side of each subplot) illustrate the evolution of nDCG scores, while the boxplots (right side of each subplot) summarize the overall distribution of nDCG scores, highlighting variability and accuracy.
    }
    
    \label{fig:ANDROIDPandexBoviaSAL}
\end{figure*}

Figure~\ref{fig:ANDROIDPandexBoviaSAL} presents the evolution of nDCG scores across active learning iterations for the Android datasets under the Pandex and Bovia attack scenarios. Across all Android datasets, active learning significantly improves nDCG scores as iterations progress. Most datasets show steep improvements early in the process, reaching optimal values quickly, whereas others exhibit more gradual progress with increased fluctuations, particularly in the Bovia scenario. This behavior likely reflects challenges related to data sparsity or the inherent complexity of distinguishing between normal and anomalous behaviors.

In the Pandex scenario, the highest nDCG score reaches 1.0 in PA, PE, and PX, indicating that active learning fully identifies true anomalies in these datasets. The PN dataset also shows a strong increase, reaching approximately 0.75. In the Bovia scenario, the highest nDCG score reaches approximately 0.91 in PE and PX with S2 and H, respectively. The remaining datasets display positive trends, with maximum nDCG scores reaching around 0.8 in PA, reflecting the increased challenge of anomaly detection in this setting.

With respect to strategy performance, all strategies consistently lead to high performance in the Pandex scenario, particularly in PA, PE, and PX. In the Bovia scenario, all strategies also demonstrate robust performance across datasets, with particularly strong results in PE and PX.

Overall, the Android datasets show strong improvements in anomaly detection through active learning. The PA, PE, and PX datasets reach optimal nDCG scores of 1.0 in the Pandex scenario, demonstrating the effectiveness of similarity search strategies in refining anomaly ranking. Although datasets under the Bovia scenario exhibit more fluctuations due to higher data complexity, they still benefit substantially from active learning. The general trend supports the conclusion that similarity search improves the identification of true anomalies over time.

\subsection{KDD Datasets}

\begin{figure}[h!]
    \centering
    \includegraphics[width=1\linewidth]{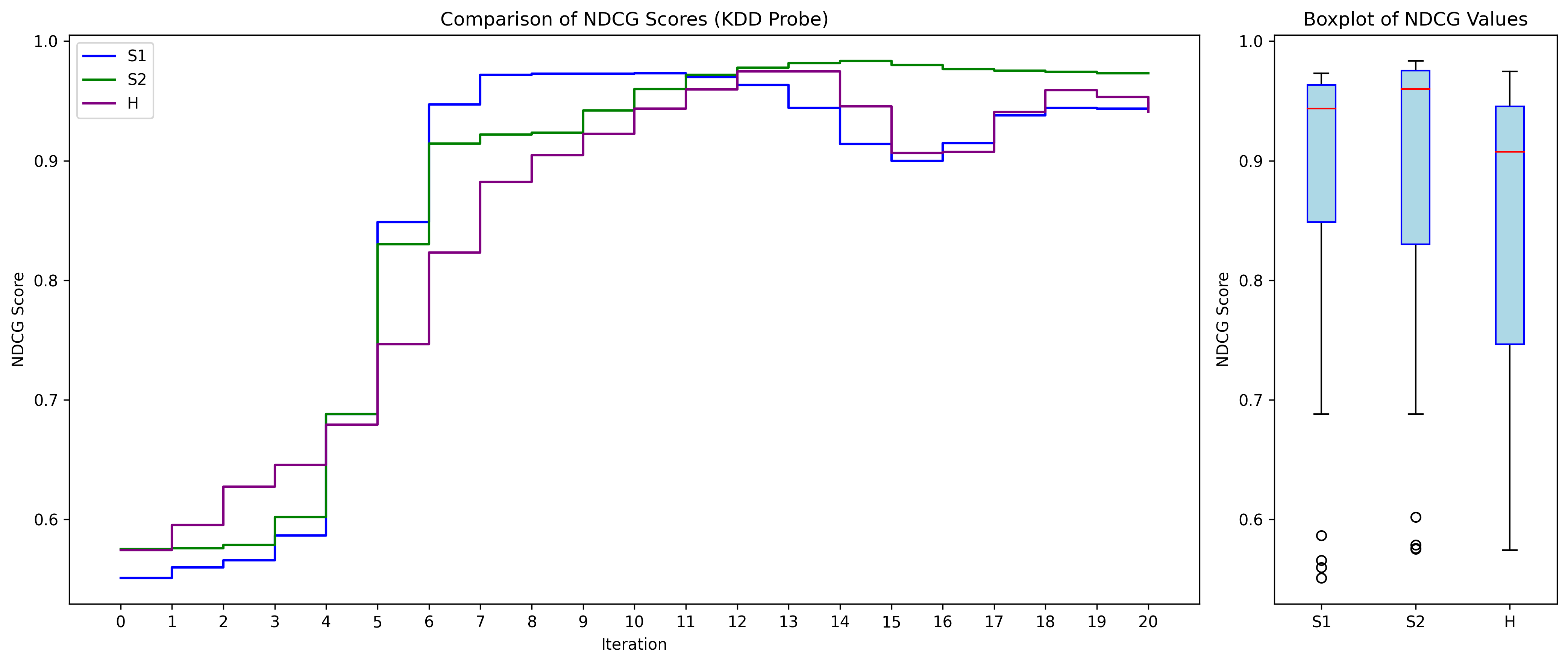}
     \includegraphics[width=1\linewidth]{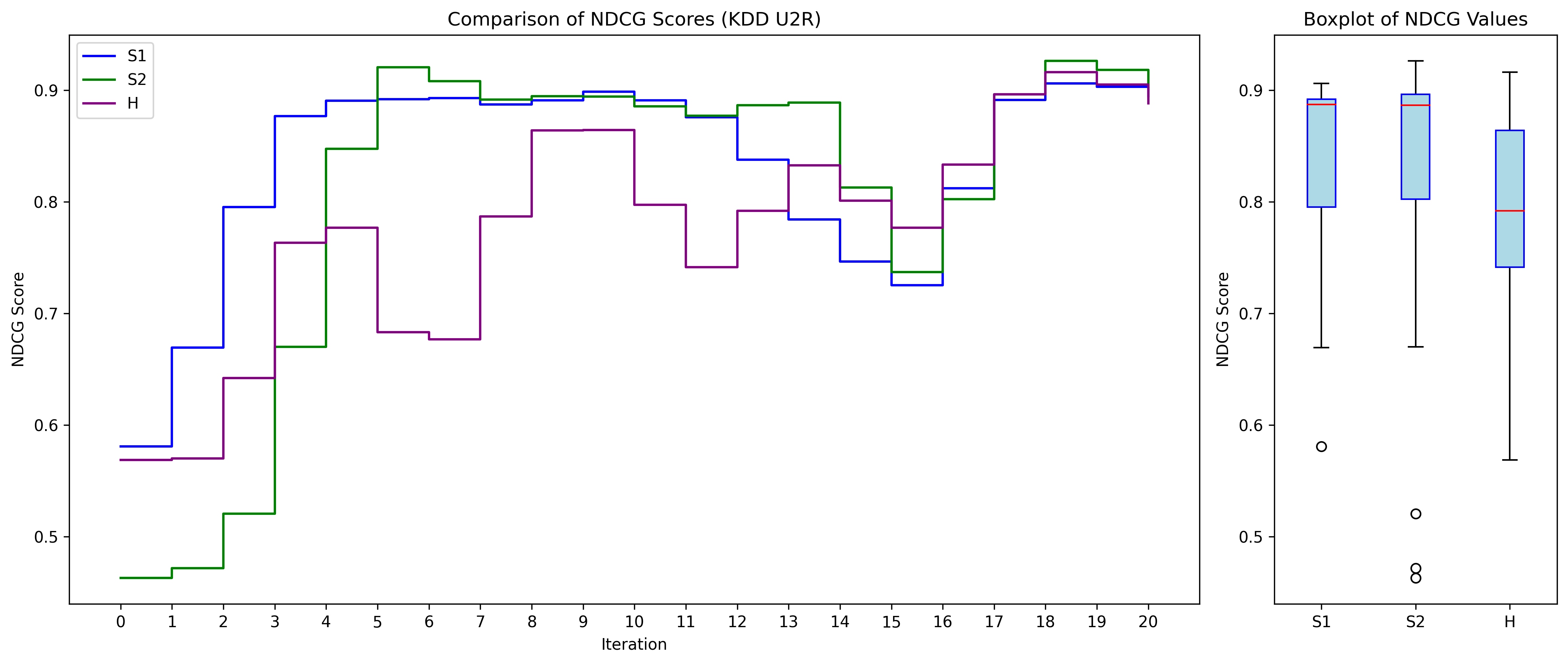}
    \caption{Comparison of nDCG scores across active learning iterations for three similarity search strategies (S1, S2, and H) on the KDD datasets (Top: Probe, Bottom: U2R).}
    \label{fig:kddprobe}
\end{figure}

The results in Figure~\ref{fig:kddprobe} present the evolution of nDCG scores across active learning iterations for the KDD datasets. Overall, active learning effectively improves the nDCG scores for both the KDD Probe and U2R datasets. The Probe dataset shows a sharp increase early in the iterations, stabilizing at high performance, while U2R exhibits more fluctuations but still benefits from active learning.

Regarding performance stability, the Probe dataset maintains a steady improvement throughout the active learning process, with all similarity search strategies converging toward an optimal nDCG score. In contrast, the U2R dataset experiences more oscillations, particularly in the middle of the learning process, indicating increased difficulty in ranking anomalies within this dataset.

A closer comparison of nDCG scores reveals that, in the Probe scenario, the highest nDCG score reaches approximately 1.0 for all three similarity search strategies, demonstrating that active learning enables near-optimal anomaly ranking. In the U2R scenario, the highest nDCG score reaches approximately 0.9, with some variation between strategies, reflecting the greater complexity of anomaly ranking in this dataset.

In terms of strategy effectiveness, the S2 strategy consistently performs well in the Probe dataset, achieving high stability, although the differences between strategies remain minimal. For the U2R dataset, S2 again achieves the best overall performance, exhibiting slightly higher stability compared to the H strategy.

Overall, the KDD datasets demonstrate that active learning significantly improves anomaly ranking performance. While all strategies converge to near-optimal nDCG scores in the Probe dataset, the U2R dataset presents a more challenging anomaly detection problem with greater performance fluctuations. Across both datasets, S2 emerges as the best-performing strategy, confirming that active learning combined with similarity search substantially enhances the model’s ability to identify and rank anomalies.

\subsection{CMC dataset:}

\begin{figure}[h!]
    \centering
    \includegraphics[width=1\linewidth]{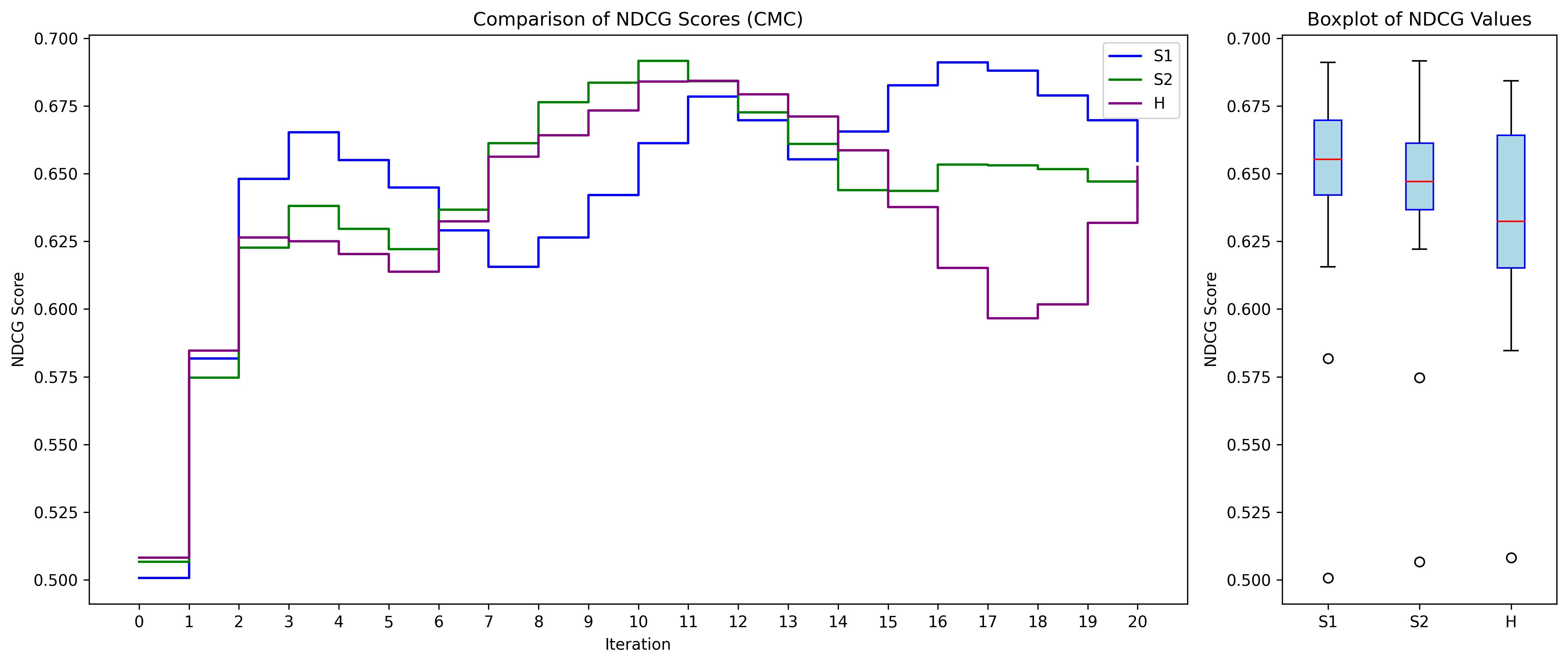}
    \caption{Comparison of nDCG scores across active learning iterations for three similarity search strategies (S1, S2, and H) on the CMC dataset.}
    \label{fig:cmc}
\end{figure}

The results in Figure~\ref{fig:cmc} present the evolution of nDCG scores across active learning iterations for the CMC dataset. Overall, the nDCG scores exhibit an increasing pattern during the active learning process, indicating that the framework improves its ability to rank true anomalies, although noticeable fluctuations appear in later iterations.

With respect to performance stability, the dataset exhibits some instability beyond iteration 10, with visible variations in nDCG scores across different similarity search strategies. The highest nDCG scores fluctuate between approximately 0.65 and 0.70, suggesting that while active learning improves ranking performance, the dataset remains challenging.

From a comparative perspective, the highest nDCG score reaches approximately 0.68 across the evaluated strategies. The median nDCG scores, as shown in the boxplots, remain relatively consistent across S1, S2, and H, with only slight variations in overall ranking effectiveness.

In terms of strategy comparison, S2 achieves the most stable performance, maintaining a relatively smooth progression without major drops across iterations. The H strategy exhibits higher variability, achieving higher peak nDCG scores at certain points but lacking consistency in later iterations. The S1 strategy shows moderate performance, gradually gaining stability over time.

In summary, the CMC dataset confirms that active learning improves anomaly ranking performance, albeit with some instability. The nDCG scores increase steadily during early iterations, while later iterations introduce fluctuations, particularly for S2 and H. Although S1 and S2 achieve the best peak performance, S1 provides more stable results overall. Despite minor inconsistencies, active learning proves beneficial in effectively ranking anomalies in this dataset.

  %
\subsection{aPascal dataset:}

\begin{figure}[h!]
    \centering
    \includegraphics[width=1\linewidth]{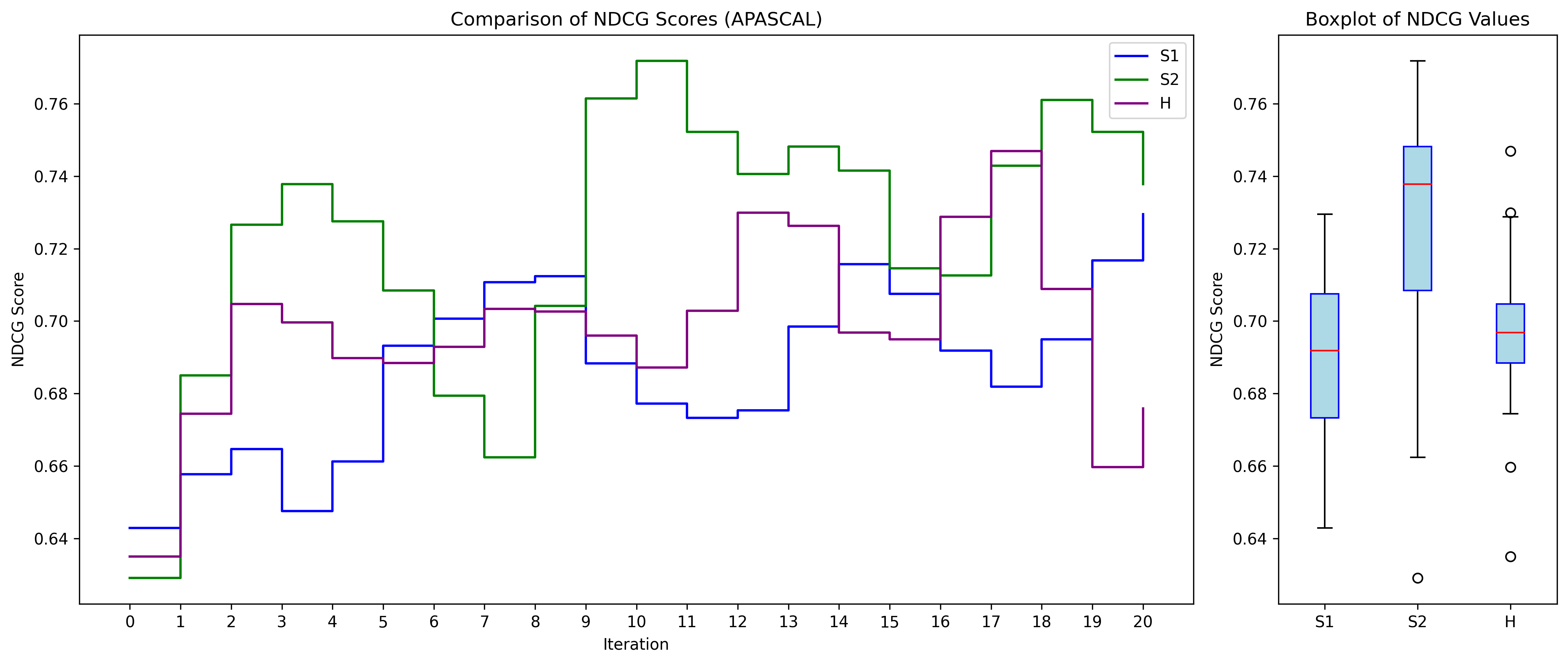}
    \caption{Comparison of nDCG scores across active learning iterations for three similarity search strategies (S1, S2, and H) on the aPascal dataset.}
    \label{fig:apascal}
\end{figure}

The results in Figure~\ref{fig:apascal} present the evolution of nDCG scores across active learning iterations for the aPascal datasets. Overall, the nDCG scores exhibit a moderate increase during the early active learning iterations, showing that the framework effectively improves ranking over time. However, noticeable fluctuations appear in later stages across the three strategies, suggesting increased sensitivity as the learning process progresses.

Variations in nDCG scores are observed across iterations, indicating that the similarity search strategies react differently to changes in the dataset structure, particularly for the S2 strategy. Despite these fluctuations, the maximum nDCG score reaches approximately 0.76, which represents solid performance across the dataset. Furthermore, the median nDCG values, ranging between 0.69 and 0.74, suggest that the similarity strategies maintain relatively stable performance overall, with only slight variations in ranking effectiveness.

In terms of strategy comparison, S2 appears to be the most effective similarity search strategy, consistently achieving the highest nDCG scores across iterations. The H strategy provides competitive performance but experiences noticeable fluctuations, which may indicate sensitivity to evolving sample selection. The S1 strategy starts with a stable trend but struggles to maintain a high final ranking as iterations progress.

In summary, the aPascal dataset demonstrates the effectiveness of active learning in improving anomaly detection performance. The S2 strategy emerges as the best-performing approach, reaching the highest nDCG values, while H and S1 exhibit competitive but more fluctuating behavior. Overall, the integration of similarity search strategies within the active learning framework improves ranking quality and highlights the adaptability of the proposed method to diverse datasets.

\subsection{AID362 Dataset:}

\begin{figure}[h!]
    \centering
    \includegraphics[width=1\linewidth]{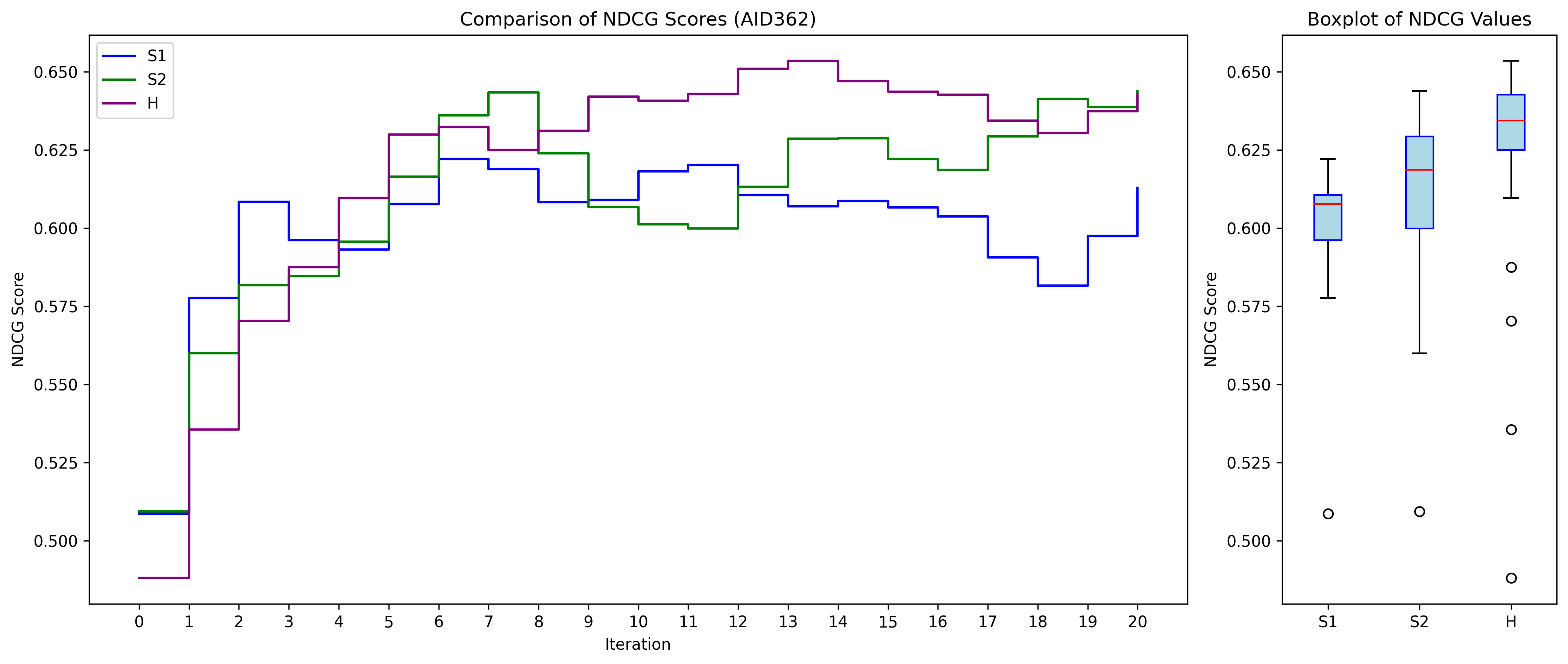}
    \caption{Comparison of nDCG scores across active learning iterations for three similarity search strategies (S1, S2, and H) on the AID362 dataset.}
    \label{fig:aid362}
\end{figure}

Figure~\ref{fig:aid362} presents the evolution of nDCG scores across active learning iterations for the AID362 dataset. Overall, the nDCG scores demonstrate a steady increase during the initial active learning iterations, confirming that the model progressively improves its ranking capability as more informative samples are selected. This trend highlights the effectiveness of the active learning process in refining anomaly ranking over time.

Despite this overall improvement, moderate fluctuations appear across iterations, particularly with the S1 strategy, which struggles to maintain a consistently high ranking performance. These fluctuations suggest that certain similarity search strategies may be more sensitive to the structure or distribution of anomalies within this dataset.

In terms of absolute performance, the highest nDCG score reaches approximately 0.65, indicating a moderate but meaningful improvement in anomaly ranking through active learning. The boxplot distributions further reveal that the S2 and H strategies exhibit more stable behavior across iterations, while S1 shows greater variability, reflecting less consistent ranking quality.

Among the evaluated strategies, H achieves the highest final nDCG score, demonstrating its effectiveness in improving ranking quality on the AID362 dataset. The S2 strategy remains competitive, maintaining relatively stable performance throughout the active learning process. In contrast, S1 underperforms compared to the other strategies, exhibiting both lower stability and weaker final performance.

Overall, the results on the AID362 dataset highlight the importance of similarity search strategy selection in active learning. The integration of similarity search with active learning clearly enhances the anomaly detection process, reinforcing the adaptability of the proposed framework across datasets with varying levels of complexity and noise.

\subsection{AD Dataset:}

\begin{figure}[th!]
    \centering
    \includegraphics[width=1\linewidth]{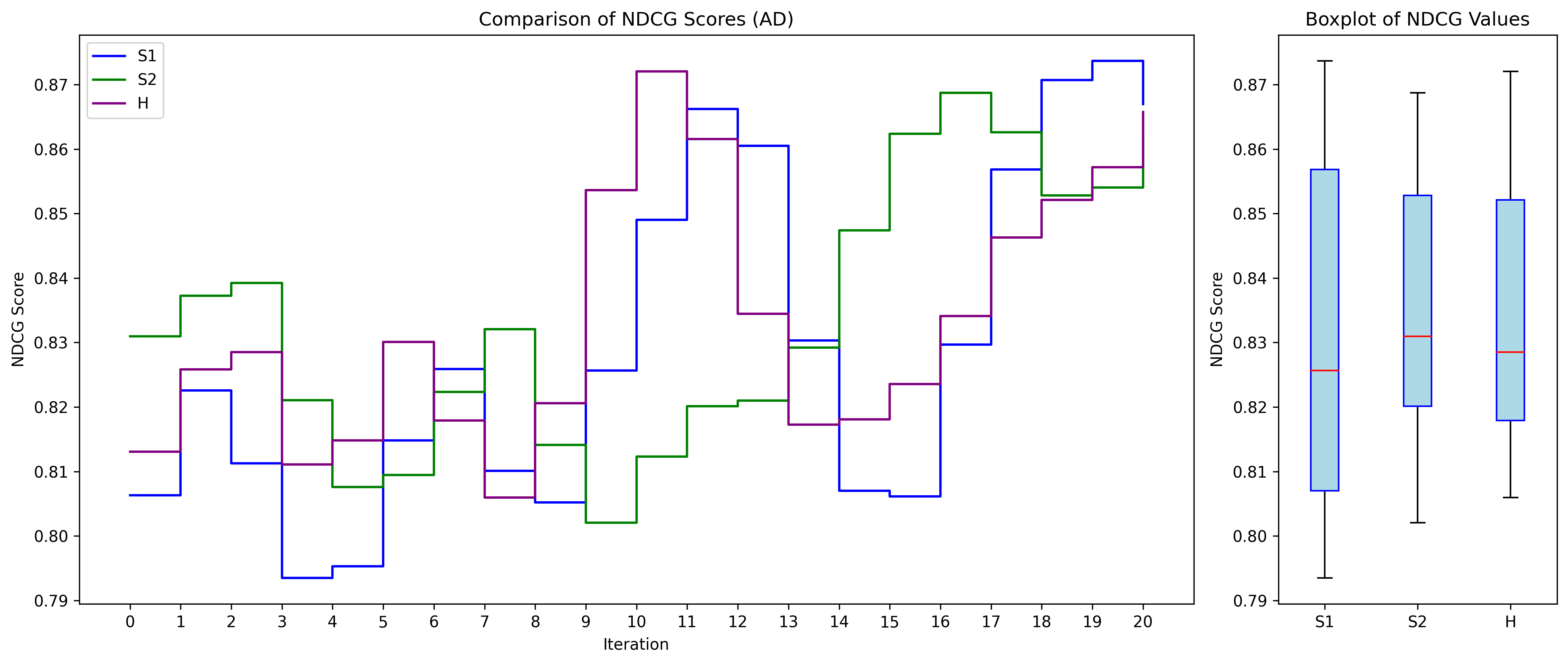}
    \caption{Comparison of nDCG scores across active learning iterations for three similarity search strategies (S1, S2, and H) on the AD dataset.}
    \label{fig:AD}
\end{figure}

The results in Figure~\ref{fig:AD} present the evolution of nDCG scores across active learning iterations for the AD dataset. Overall, the nDCG scores show a consistent upward trajectory across the later iterations, indicating that the active learning process successfully enhances the model’s anomaly ranking capabilities as more informative samples are incorporated.

Despite this general improvement, moderate fluctuations are observed across all three similarity search strategies, with peaks and drops occurring throughout the learning process. These variations suggest that the ranking task remains non-trivial and sensitive to the selection of queried samples at different stages of active learning.

In terms of nDCG score magnitude, the highest value reaches approximately 0.87, demonstrating strong ranking performance across all strategies. The boxplots further reveal that S1, S2, and H exhibit similar score distributions, with overlapping interquartile ranges, indicating comparable overall effectiveness in ranking anomalies.

Among the three strategies, H achieves the highest overall ranking stability, maintaining competitive scores across most iterations. S2 also demonstrates strong performance but exhibits increased variability during mid-iterations. S1 follows closely, showing a consistent upward trend, although its final nDCG scores remain slightly lower than those achieved by H.

Overall, the AD dataset highlights the effectiveness of combining active learning with similarity search strategies. All three strategies reach relatively high and comparable final nDCG scores, confirming the robustness of the framework. The H strategy provides the best balance between stability and final performance, while S2 remains a strong but slightly more variable alternative. S1 shows steady improvement throughout the process, reinforcing the adaptability of the framework across different similarity search strategies.

\subsection{Cover Type dataset:}
\begin{figure}[h!]
    \centering
    \includegraphics[width=1\linewidth]{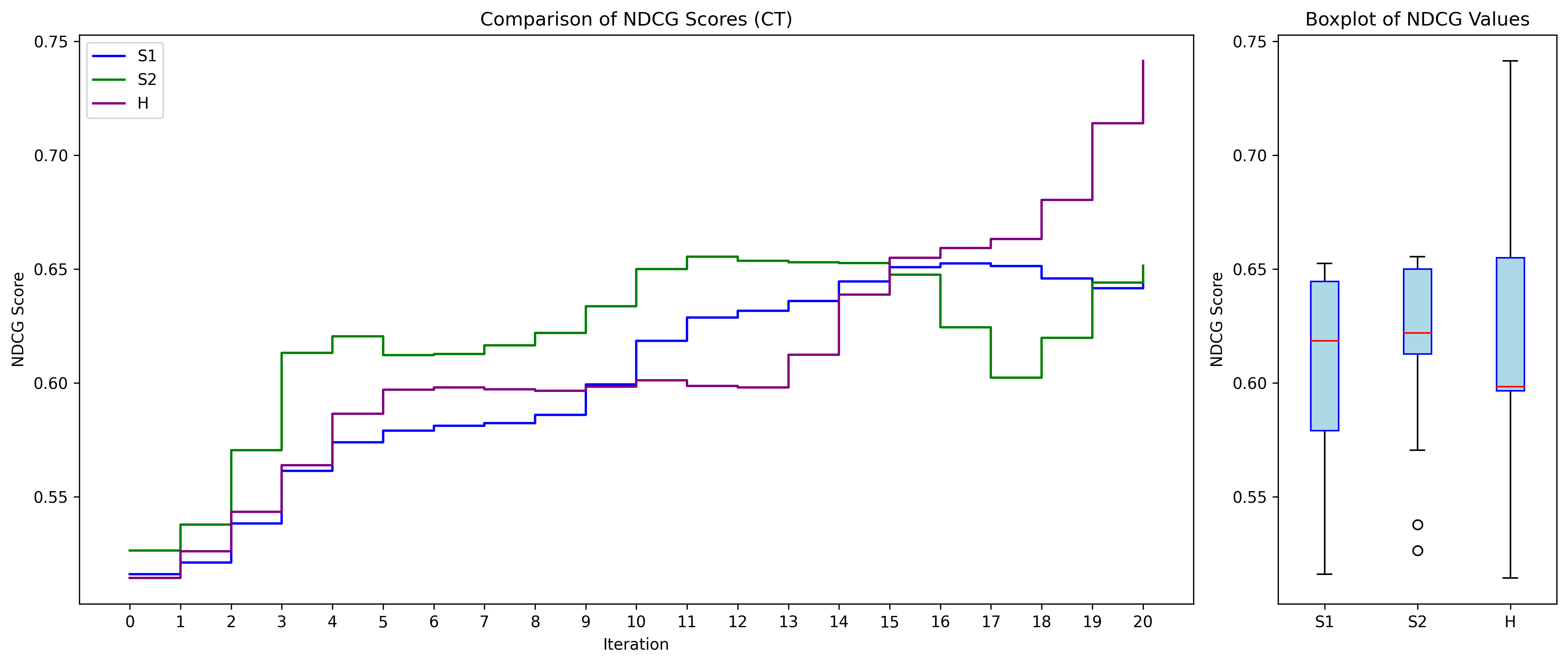}
    \caption{Comparison of nDCG scores across active learning iterations for three similarity search strategies (S1, S2, and H) on the Cover Type dataset.}
    \label{fig:ct}
\end{figure}

The results in Figure~\ref{fig:ct} present the evolution of nDCG scores across active learning iterations for the Cover Type dataset.

Overall, the nDCG scores show a gradual increase across iterations, indicating that active learning effectively enhances the ranking of true anomalies. The score progression remains relatively stable compared to other datasets, with only minimal fluctuations observed. Slight variations appear in later iterations, but these do not significantly affect the overall upward trend.

In terms of nDCG score comparison, the highest nDCG score reaches approximately 0.75, with steady performance observed across all similarity search strategies. The boxplots further indicate that the three strategies (S1, S2, and H) exhibit overlapping inter-quartile ranges, suggesting comparable overall ranking effectiveness across strategies.

When comparing similarity search strategies, the H strategy demonstrates the highest final nDCG score, showing a notable increase during the last iterations. The S2 strategy maintains a competitive performance throughout the learning process but tends to stabilize earlier. The S1 strategy follows closely, exhibiting a stable trend with slightly lower overall ranking performance compared to H and S2.

In summary, the Cover Type dataset highlights the robustness of combining active learning with similarity search. The overall nDCG scores improve steadily with minimal fluctuations, confirming the stability of the learning process. The H strategy achieves the highest peak performance, while S2 remains a strong competitor and S1 follows closely, reinforcing the reliability of different similarity search strategies. Overall, active learning proves effective in refining anomaly detection performance on this dataset.

\subsection{Celeba Dataset}

\begin{figure}[h!]
    \centering
    \includegraphics[width=1\linewidth]{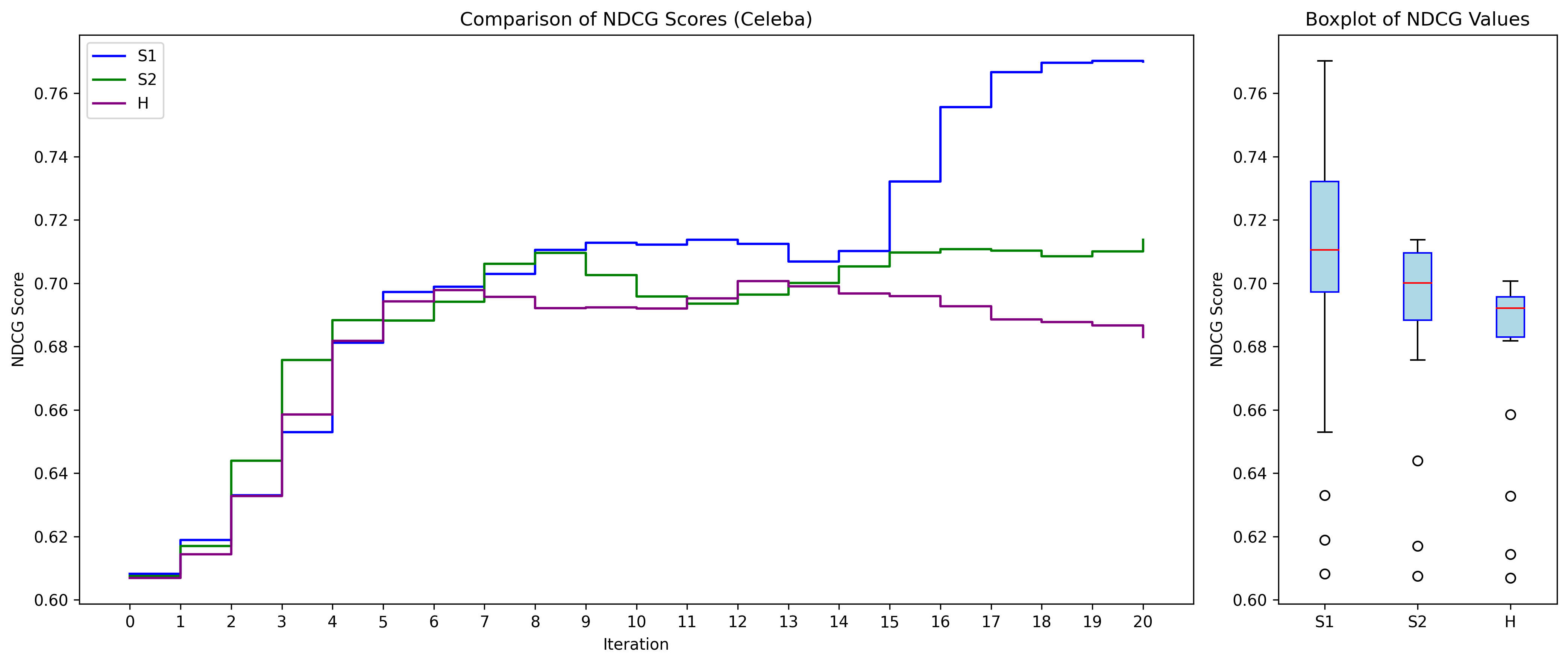}
    \caption{Comparison of nDCG scores across active learning iterations for three similarity search strategies (S1, S2, and H) on the Celeba datasets.}
    \label{fig:celeba}
\end{figure}

The results in Figure~\ref{fig:celeba} present the evolution of nDCG scores across active learning iterations for the CelebA dataset. Overall, the nDCG scores exhibit a consistent increase throughout the active learning process, demonstrating improved anomaly ranking as more labeled samples are incorporated into training. This steady upward trend confirms the effectiveness of active learning in progressively refining the ranking of anomalous instances.

Some fluctuations are observed during the early iterations; however, the scores stabilize after a few rounds of querying. This behavior suggests that once a sufficient number of informative samples are selected, the active learning process effectively refines the model’s decision boundary and reduces ranking instability.

In terms of absolute performance, the highest nDCG score reaches approximately 0.77, indicating strong anomaly ranking capability. The boxplots further show that the median nDCG scores across the different similarity search strategies lie between 0.69 and 0.71, reflecting relatively consistent performance with moderate variability across strategies.

Among the similarity search strategies, S1 achieves the highest peak nDCG score (0.77), demonstrating its effectiveness in prioritizing true anomalies during the active learning process. S2 shows stable performance with moderate fluctuations but does not surpass S1 in peak performance. The H strategy exhibits the lowest variance across iterations, indicating stable behavior, but converges to a slightly lower final nDCG score compared to S1.

Overall, the CelebA dataset results confirm that active learning substantially improves the framework’s ability to detect anomalies. Among the three similarity search strategies, S1 consistently provides the strongest peak performance, while S2 and H offer more stable but slightly lower final rankings. These results highlight the benefit of integrating similarity-based active learning for improving anomaly detection in complex, high-dimensional image datasets.

\subsection{Reuters10 dataset:}
\begin{figure}[h!]
    \centering
    \includegraphics[width=1\linewidth]{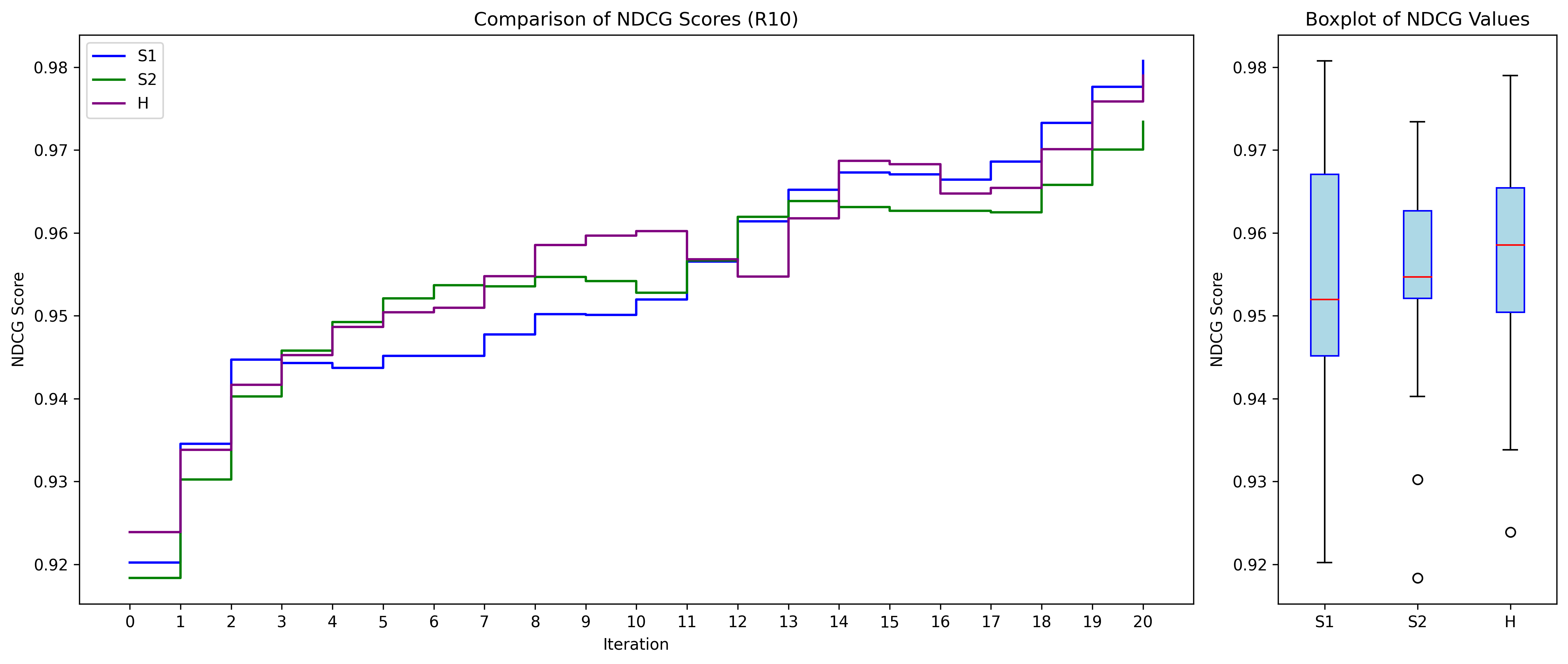}
    \caption{Comparison of nDCG scores across active learning iterations for three similarity search strategies (S1, S2, and H) on the Reuters10 dataset.}
    \label{fig:R10}
\end{figure}

The results in Figure~\ref{fig:R10} present the evolution of nDCG scores across active learning iterations for the Reuters10 dataset.

Across iterations, the nDCG scores consistently increase, indicating that the framework successfully improves anomaly ranking as more labeled data is integrated into the learning process. Fluctuations remain minimal throughout training, suggesting that the active learning mechanism stabilizes quickly and avoids erratic ranking behavior.

In terms of absolute performance, the highest nDCG score reaches approximately 0.98, confirming a strong anomaly ranking capability. Moreover, the median nDCG scores for all three similarity search strategies remain consistently high, ranging between 0.95 and 0.96, as illustrated by the boxplots. This reflects both high accuracy and low variance across strategies.

Regarding strategy comparison, all similarity search methods perform closely on this dataset. The H strategy achieves slightly higher stability during the later iterations, while S1 and S2 reach nearly identical peak performance. This demonstrates the robustness of the framework and suggests that the choice of similarity search strategy has limited impact on final performance for Reuters10.

Overall, the Reuters10 dataset results indicate that all similarity search strategies lead to strong and stable anomaly detection performance with minimal fluctuations. The active learning framework effectively refines the model’s ranking ability, reaching an optimal nDCG score of 0.98. While H appears marginally more stable, all strategies demonstrate comparable effectiveness.

\subsection{Solar Flare Dataset:}
\begin{figure}[h!]
    \centering
    \includegraphics[width=1\linewidth]{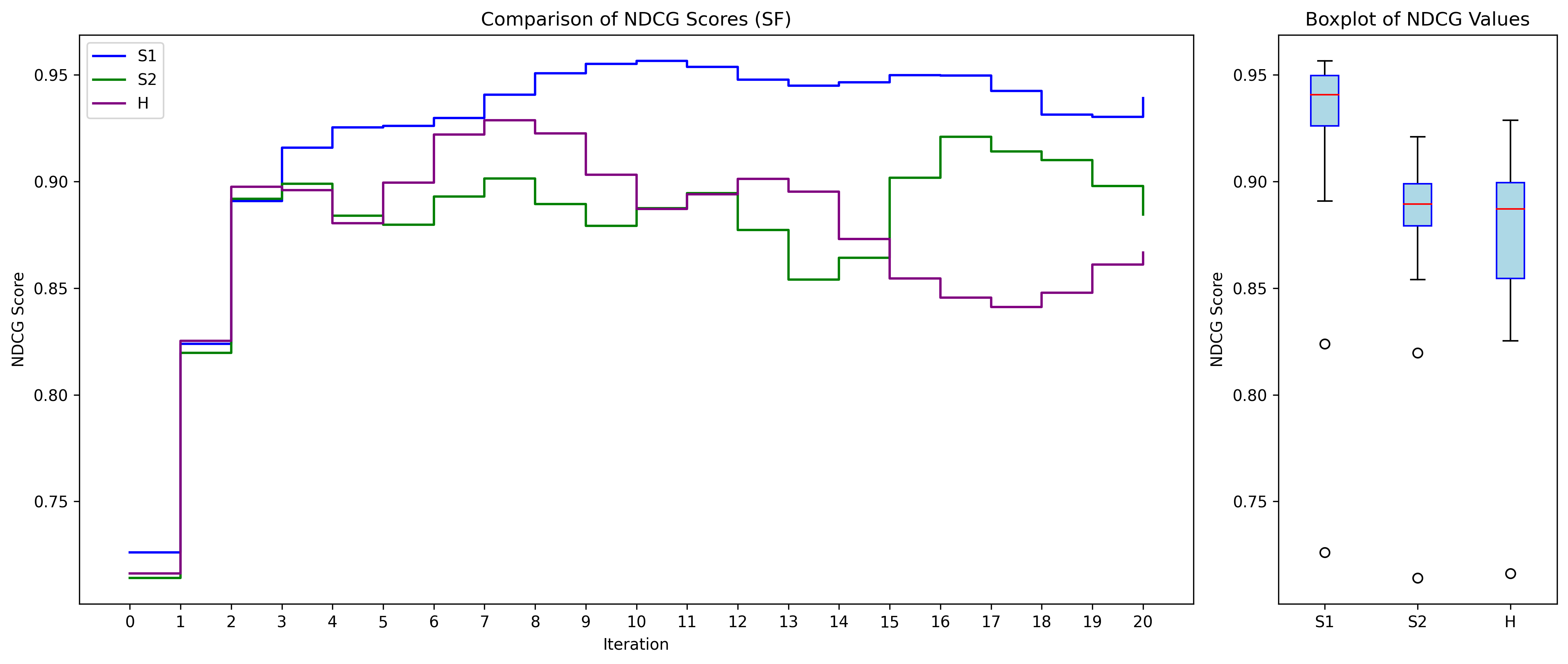}
    \caption{Comparison of nDCG scores across active learning iterations for three similarity search strategies (S1, S2, and H) on the Solar Flare dataset.}
    \label{fig:sf}
\end{figure}

The results in Figure~\ref{fig:sf} present the evolution of nDCG scores across active learning iterations for the Solar Flare dataset. Overall, the nDCG scores exhibit a strong upward trend as active learning progresses, indicating that the framework successfully improves anomaly ranking as more labeled data becomes available.

While some fluctuations are observed in the later iterations, the overall performance remains high, suggesting stable learning dynamics throughout the active learning process. In terms of absolute performance, the highest nDCG score reaches approximately 0.96, demonstrating high-ranking effectiveness. Additionally, the median nDCG scores across the different similarity search strategies remain consistently strong, ranging between 0.88 and 0.95.

Regarding the comparison between similarity search strategies, S1 achieves the best final performance, consistently maintaining higher nDCG scores across the majority of iterations. Both S2 and H demonstrate competitive performance; however, they exhibit slightly more variability in later iterations.

In summary, the results on the Solar Flare dataset indicate that all similarity search strategies lead to strong anomaly detection performance. Among them, S1 emerges as the most stable and effective strategy, achieving the highest nDCG score of 0.96. Despite minor fluctuations, the integration of active learning significantly enhances the ranking capabilities of the framework.

\subsection{LibSVM w7A dataset:}

\begin{figure}[h!]
    \centering
    \includegraphics[width=1\linewidth]{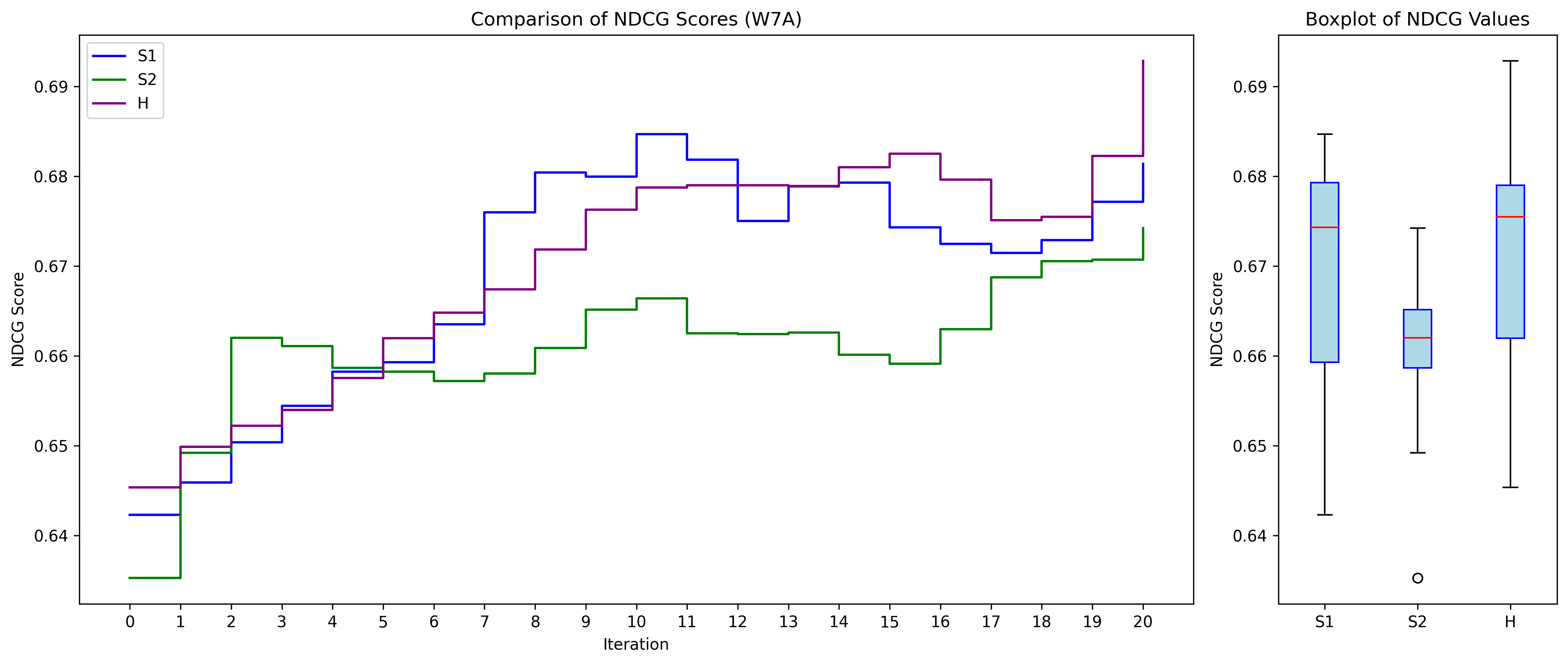}
    \caption{Comparison of nDCG scores across active learning iterations for three similarity search strategies (S1, S2, and H) on the LibSVM w7A dataset.}
    \label{fig:w7a}
\end{figure}

The results in Figure~\ref{fig:w7a} present the evolution of nDCG scores across active learning iterations for the LibSVM w7A dataset. Overall, the nDCG scores show a steady improvement over active learning iterations, with all three strategies (S1, S2, and H) gradually increasing their ranking accuracy. This trend indicates that the active learning process consistently enhances the model’s ability to prioritize true anomalies as more informative samples are incorporated.

The score progression remains relatively stable across iterations, with only minor fluctuations observed, particularly in the later stages of learning. These fluctuations suggest slight variations in anomaly ranking but do not undermine the overall upward trajectory of performance. In terms of absolute performance, the highest nDCG score achieved is approximately 0.69, with strategy H demonstrating a slight advantage in the later iterations. The median nDCG scores across all strategies are close, indicating relatively comparable overall performance among S1, S2, and H.

Regarding the effectiveness of similarity search strategies, strategy H achieves the highest final nDCG score, highlighting its effectiveness in ranking anomalies more accurately in this dataset. Strategy S1 maintains a stable performance throughout the learning process and ranks closely behind H, demonstrating strong robustness. Strategy S2 appears slightly less effective in the later iterations but still shows competitive performance overall.

In summary, the evaluation of active learning on the LibSVM w7A dataset confirms that all three similarity search strategies benefit from iterative querying and labeling. Strategy H emerges as the most effective, achieving the highest nDCG score, while S1 and S2 also perform well, with S2 exhibiting slightly lower ranking stability in later iterations. Overall, these results validate the benefits of active learning in improving anomaly detection performance on the LibSVM w7A dataset.

\subsection{BM Dataset:}

\begin{figure}[h!]
    \centering
    \includegraphics[width=1\linewidth]{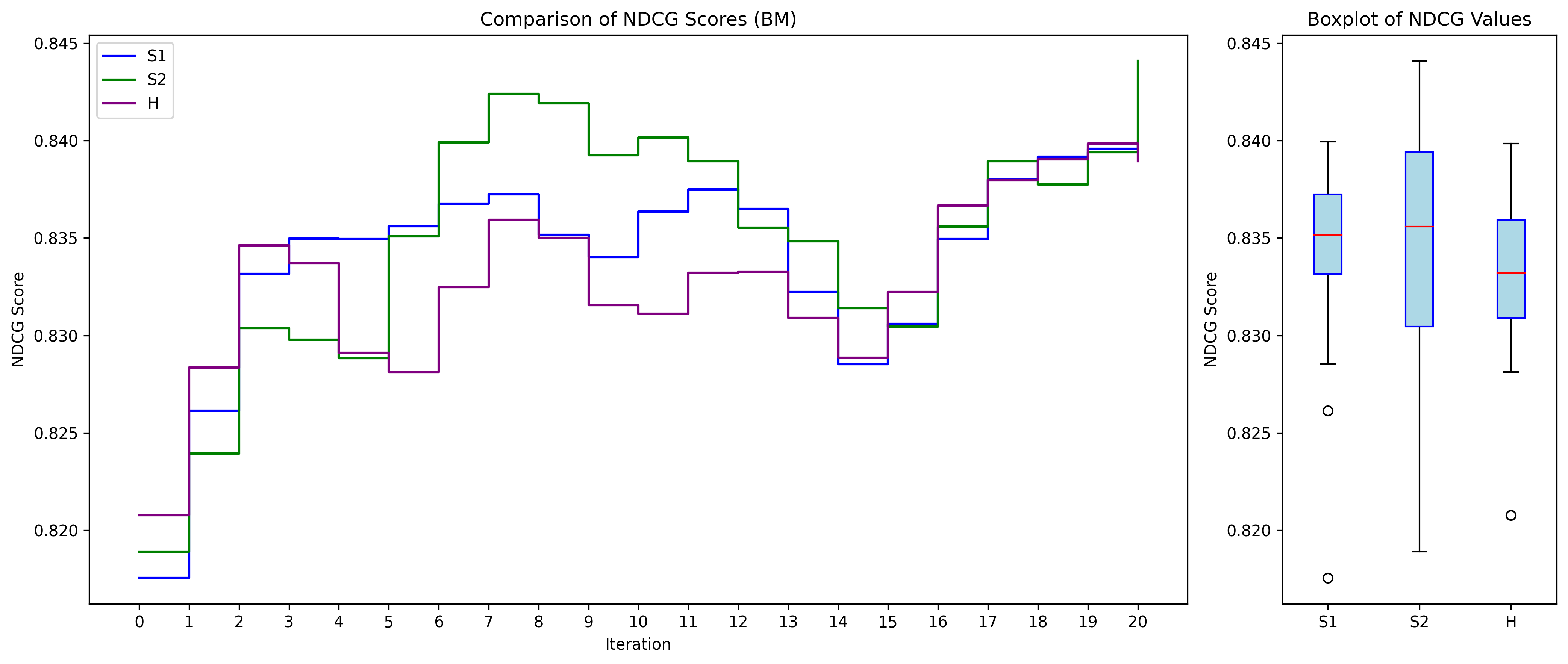}
    \caption{Comparison of nDCG scores across active learning iterations for three similarity search strategies (S1, S2, and H) on the BM dataset.}
    \label{fig:BM}
\end{figure}

The results in Figure~\ref{fig:BM} present the evolution of nDCG scores across active learning iterations for the BM dataset.

Overall, the nDCG scores for the BM dataset show a consistent increase throughout the active learning iterations. This indicates that the model is progressively improving its ability to rank anomalies correctly as more informative samples are selected. While the learning curves exhibit minor fluctuations, particularly during the early iterations, these variations diminish over time and the performance stabilizes as active learning progresses. This behavior suggests that the model gradually refines its decision boundaries with the incorporation of additional labeled data.

In terms of ranking performance, the highest observed nDCG score reaches approximately 0.85, indicating a strong anomaly ranking capability. All three similarity search strategies—S1, S2, and H—converge to comparable final nDCG scores. Among them, S2 slightly outperforms the others in the final iterations, achieving the highest peak performance. However, the differences between strategies remain modest, highlighting the robustness of the overall framework.

From a strategy perspective, S2 demonstrates the strongest balance between performance and stability, consistently achieving high nDCG scores across iterations. Nonetheless, both S1 and H remain competitive throughout the learning process, confirming that all similarity search strategies contribute effectively to the active learning framework.

In summary, the BM dataset demonstrates stable and consistent performance improvements through active learning. The final nDCG scores indicate that all similarity search strategies enhance anomaly ranking, with S2 leading marginally. The limited fluctuations and clear convergence trends further confirm that active learning effectively guides the model toward more reliable and accurate anomaly detection.

%
\subsection{Comparison with the existing methods:}

To ensure a comprehensive evaluation, we compare our proposed SDA$^2$E framework with a diverse collection of anomaly detection baselines, including statistical models, classical machine learning algorithms, rule-based approaches, neural-evolution methods, and state-of-the-art deep-learning models such as TTVAE, ATDAD, and AnoGAN. These evaluations span previous domains—cybersecurity, computer vision, public health, natural sciences, and business analytics—ensuring that the observed performance is robust and generalizable.
Regarding the proposed approach, we use statistical summarization in the three similarity search strategies (S1, S2, and H). Since our approach evaluates three different similarity search techniques, each dataset yields three separate nDCG score progressions over the active learning iterations. In contrast, the SOTA methods provide only a single nDCG score per dataset. To establish a robust comparative framework, we define three key metrics for our approach:
\begin{itemize}
    \item Max\_Max: It represents the highest maximum nDCG score attained across the three strategies, formally defined as:
\[\text{Max\_Max} = \max\left(\max(S1), \max(S2), \max(H)\right)
    \]
This metric highlights the \textit{best-case performance} of our method compared to the highest reported score of a SOTA method. However, it does not account for \textit{consistency} across iterations, making it less reliable when comparing to SOTA methods that may have \textit{stable} high scores. Notably, if we were to evaluate our approach solely using \emph{Max-Max}, our method would emerge as the \textit{best-performing} approach across all datasets. However, this would provide an inflated view of its effectiveness, as it does not reflect overall stability or robustness.

\item Max\_Mean: It represents the highest mean nDCG score across the three strategies, computed as:
\[\text{Max\_Mean} = \max\left(\text{mean}(S1), \text{mean}(S2), \text{mean}(H)\right)
    \]
This measure reflects the \textit{average effectiveness} of our approach, ensuring that improvements are not just isolated peaks but are sustained throughout the active learning process. While this measure accounts for \textit{fluctuations} in performance, it can be affected by extreme values, making it less robust to outliers.
\item Max\_Median: It represents the highest median nDCG score among the three strategies:
\[\text{Max\_Median} = \max\left(\text{median}(S1), \text{median}(S2), \text{median}(H)\right)
     \]
Since the median is less sensitive to extreme values, it provides a \emph{more stable and robust evaluation} of our method’s effectiveness. Given that SOTA methods generally exhibit \emph{stable} performance rather than extreme peaks, the \emph{Max-Median metric ensures a fair comparison}. Furthermore, the median captures the \emph{typical performance} over multiple iterations, making it an ideal choice for assessing \emph{real reliability}.
\end{itemize}
The results of the global comparison are provided in Table \ref{tab:global-comparison}. Each row represents a sub-dataset belonging to a given category. For our framework, we show the triplet (Max\_Max, Max\_Mean, Max\_Median). Based on the above considerations, we select \textit{Max-Median} as our primary metric for comparison against SOTA methods. Consequently, the last column of the table highlights the winning approach based on this criterion. This choice ensures that our evaluation reflects not only \textit{peak performance} but also \textit{consistency and robustness}. To further validate our conclusions, we also report \textit{Max-Mean} to highlight overall effectiveness and include \textit{Max-Max} as a supplementary measure to showcase the theoretical best-case performance.


\begin{table*}[]
\scriptsize
\centering
\begin{adjustbox}{angle=90}
\begin{tabular}{|lllll||c|c|c|c|c|c|c|c|c|c|c|c|c|c||c|}
\hline
\multicolumn{5}{|l||}{\textbf{Database} \textbackslash \textbf{~Detection method}}    & \textbf{AVF} & \textbf{OC3} & \textbf{OD} & \textbf{VF-ARM} & \textbf{VR-ARM} & \textbf{IForest} & \textbf{OC-SVM} & \textbf{EE} & \textbf{LOF} & \textbf{FPOF}& \textbf{ATDAD} & \textbf{TTVAE}&\textbf{AnoGAN}&\textbf{SDA2E} & \textbf{Winner}   \\ \hline \hline
\multicolumn{1}{|l|}{\multirow{40}{*}{\rotatebox[origin=c]{90}{\textbf{Cyber Security}}}} & \multicolumn{1}{l|}{\multirow{38}{*}{\textbf{TC}}} & \multicolumn{1}{l|}{\multirow{10}{*}{{\rotatebox[origin=c]{90}{\textbf{BSD}}}}}     & \multicolumn{1}{l|}{\multirow{5}{*}{\textbf{E1}}} & \textbf{PA}          & 0.52         & 0.65         & 0.19        & 0.18            & 0.36            & 0.13             & 0.13            & 0.14        & 0.07         & 0.21     &   0.45 &0.77 & 0.42 &   0.91|0.84|\textbf{\cellcolor{yellow}0.88}             & \multirow{5}{*}{} \\ \cline{5-19}
\multicolumn{1}{|l|}{}                                          & \multicolumn{1}{l|}{}                              & \multicolumn{1}{l|}{}                                   & \multicolumn{1}{l|}{}                             & \textbf{PE}          & 0.51         & 0.43         & 0.19        & 0.33            & 0.64            & 0.14             & 0.14            & 0.13        & 0.10         & 0.20  &0.70 & 0.50&     0.26   &     0.84|0.74|\textbf{\cellcolor{yellow}0.77}           &                   \\ \cline{5-19}
\multicolumn{1}{|l|}{}                                          & \multicolumn{1}{l|}{}                              & \multicolumn{1}{l|}{}                                   & \multicolumn{1}{l|}{}                             & \textbf{PX}          & 0.34         & 0.49         & 0.15        & 0.53            & 0.08            & 0.14             & 0.14            & 0.14        & 0.10         & 0.15       &  0.70 &   0.73 &0.32 &0.89|0.80|\textbf{\cellcolor{yellow}0.83}          &       SDA2E            \\ \cline{5-19}
\multicolumn{1}{|l|}{}                                          & \multicolumn{1}{l|}{}                              & \multicolumn{1}{l|}{}                                   & \multicolumn{1}{l|}{}                             & \textbf{PP}          & 0.30         & 0.43         & 0.13        & 0.67            & 0.29            & 0.11             & 0.11            & 0.13        & 0.08         & 0.13  &     0.67   &  0.76  &0.28 &  0.82|0.80|\textbf{\cellcolor{yellow}0.82}         &   0.88                \\ \cline{5-19}
\multicolumn{1}{|l|}{}                                          & \multicolumn{1}{l|}{}                              & \multicolumn{1}{l|}{}                                   & \multicolumn{1}{l|}{}                             & \textbf{PN}          & 0.25         & 0.32         & 0.14        & 0.11            & 0.34            & 0.10             & 0.10            & 0.14        & 0.24         & 0.13    & 0.63     &  \cellcolor{yellow}\textbf{0.73} & 0.32& 0.36|0.33|0.36           &                   \\ \cline{4-20} 
\multicolumn{1}{|l|}{}                                          & \multicolumn{1}{l|}{}                              & \multicolumn{1}{l|}{}                                   & \multicolumn{1}{l|}{\multirow{5}{*}{\textbf{E2}}} & \textbf{PA}          & 0            & 0.38         & 0.19        & 0.14            & 0.52            & 0.17             & 0.18            & 0.19        & 0.09         & 0.21 &    0.68     & 0.73 &0.30 &  1|0.98|\textbf{\cellcolor{yellow}1}            & \multirow{5}{*}{} \\ \cline{5-19}
\multicolumn{1}{|l|}{}                                          & \multicolumn{1}{l|}{}                              & \multicolumn{1}{l|}{}                                   & \multicolumn{1}{l|}{}                             & \textbf{PE}          & 0.19         & 0.24         & 0.17        & 0.12            & 0.12            & 0.12             & 0.19            & 0.19        & 0.19         & 0.13          & 0.77&0.76&0.23 &1|0.87|\textbf{\cellcolor{yellow}1}            &                   \\ \cline{5-19}
\multicolumn{1}{|l|}{}                                          & \multicolumn{1}{l|}{}                              & \multicolumn{1}{l|}{}                                   & \multicolumn{1}{l|}{}                             & \textbf{PX}          & 0.17         & 0.51         & 0.17        & 0.06            & 0.05            & 0.13             & 0.19            & 0.19        & 0.19         & 0.18          & 0.71  &0.90& 0.29&1|0.79|\textbf{\cellcolor{yellow}0.97}           &     SDA2E              \\ \cline{5-19}
\multicolumn{1}{|l|}{}                                          & \multicolumn{1}{l|}{}                              & \multicolumn{1}{l|}{}                                   & \multicolumn{1}{l|}{}                             & \textbf{PP}          & 0.17         & 0.29         & 0.09        & 0.06            & 0.24            & 0.10             & 0.19            & 0.19        & 0.19         & 0.10          & 0.76 &0.90&0.75 &1|0.98|\textbf{\cellcolor{yellow}1}              &     1              \\ \cline{5-19}
\multicolumn{1}{|l|}{}                                          & \multicolumn{1}{l|}{}                              & \multicolumn{1}{l|}{}                                   & \multicolumn{1}{l|}{}                             & \textbf{PN}          & 0            & 0            & 0.15        & 0.18            &0.60           & 0.12             & 0.05            & 0.05        & 0.06         & 0             & \textbf{\cellcolor{yellow}0.63} &    0.30   & 0.33&0.15|0.15|0.15       &                   \\ \cline{3-20} 
\multicolumn{1}{|l|}{}                                          & \multicolumn{1}{l|}{}                              & \multicolumn{1}{l|}{\multirow{10}{*}{{\rotatebox[origin=c]{90}{\textbf{Windows}}}}} & \multicolumn{1}{l|}{\multirow{5}{*}{\textbf{E1}}} & \textbf{PA}          & 0.52         & 0            & 0           & 0.50            & 0.61            & 0.21             & 0.14            & 0.10        & 0.018        & 0 & 0.40   & 0.36&     0.51 & 0.95|0.76|\textbf{\cellcolor{yellow}0.80}              & \multirow{5}{*}{} \\ \cline{5-19}
\multicolumn{1}{|l|}{}                                          & \multicolumn{1}{l|}{}                              & \multicolumn{1}{l|}{}                                   & \multicolumn{1}{l|}{}                             & \textbf{PE}          & 0.60         & 0.30         & 0.20        & 0.33            & 0.82            & 0.14             & 0.14            & 0.38        & 0.07         & 0.20          &   0.86 &0.68& 0.58&1|0.91|\textbf{\cellcolor{yellow}0.96}               &                \\ \cline{5-19}
\multicolumn{1}{|l|}{}                                          & \multicolumn{1}{l|}{}                              & \multicolumn{1}{l|}{}                                   & \multicolumn{1}{l|}{}                             & \textbf{PX}          & 0.28         & 0.28         & 0.15        & 0               & 0               & 0.17             & 0.17            & 0.18        & 0.07         & 0.15          & 0.35  &0.46&0.16 & 0.69|0.47|\textbf{\cellcolor{yellow}0.51}            &    SDA2E               \\ \cline{5-19}
\multicolumn{1}{|l|}{}                                          & \multicolumn{1}{l|}{}                              & \multicolumn{1}{l|}{}                                   & \multicolumn{1}{l|}{}                             & \textbf{PP}          & 0.21         & 0.21         & 0.10        & 0               & 0               & 0.09             & 0.09            & 0.10        & 0.01         & 0.10          &0.14   &0.52& 0.14&0.75|0.61|\textbf{\cellcolor{yellow}0.75}             &       1            \\ \cline{5-19}
\multicolumn{1}{|l|}{}                                          & \multicolumn{1}{l|}{}                              & \multicolumn{1}{l|}{}                                   & \multicolumn{1}{l|}{}                             & \textbf{PN}          & 0.58         & 0.58         & 0.36        & 0               & 0.52            & 0.32             & 0.32            & 0.15        & 0.17         & 0.36          & 0.68 &0.63& 0.44&1|0.98|\textbf{\cellcolor{yellow}1}              &                   \\ \cline{4-20} 
\multicolumn{1}{|l|}{}                                          & \multicolumn{1}{l|}{}                              & \multicolumn{1}{l|}{}                                   & \multicolumn{1}{l|}{\multirow{5}{*}{\textbf{E2}}} & \textbf{PA}          & 0            & 0            & 0           & 0.07            & 0.35            & 0.23             & 0.22            & 0.17        & 0.13         & 0    &    0.30     & 0.39 &0.31 &0.99|0.74|\textbf{\cellcolor{yellow}0.71}              & \multirow{5}{*}{} \\ \cline{5-19}
\multicolumn{1}{|l|}{}                                          & \multicolumn{1}{l|}{}                              & \multicolumn{1}{l|}{}                                   & \multicolumn{1}{l|}{}                             & \textbf{PE}          & 0.21         & 0.23         & 0           & 0.13            & 0.19            & 0.16             & 0.23            & 0.24        & 0.01         & 0             &  0.23 & 0.32  & 0.24&0.57|0.44|\textbf{\cellcolor{yellow}0.43}         &                   \\ \cline{5-19}
\multicolumn{1}{|l|}{}                                          & \multicolumn{1}{l|}{}                              & \multicolumn{1}{l|}{}                                   & \multicolumn{1}{l|}{}                             & \textbf{PX}          & 0.22         & 0.24         & 0           & 0               & 0               & 0.15             & 0.25            & 0.23        & 0.01         & 0             &  0.27 & 0.30  & 0.28& 0.92|0.67|\textbf{\cellcolor{yellow}0.74}         &     SDA2E              \\ \cline{5-19}
\multicolumn{1}{|l|}{}                                          & \multicolumn{1}{l|}{}                              & \multicolumn{1}{l|}{}                                   & \multicolumn{1}{l|}{}                             & \textbf{PP}          & 0.22         & 0.22         & 0           & 0               & 0               & 0.30             & 0.22            & 0.25        & 0.01         & 0             &   0.22 &0.37& 0.47& 0.97|0.86|\textbf{\cellcolor{yellow}0.95}           &     0.95              \\ \cline{5-19}
\multicolumn{1}{|l|}{}                                          & \multicolumn{1}{l|}{}                              & \multicolumn{1}{l|}{}                                   & \multicolumn{1}{l|}{}                             & \textbf{PN}          & 0            & 0            & 0           & 0               & 0.30            & 0.11             & 0.11            & 0.21        & 0.03         & 0             &  0.33 &0.26& 0.35&0.38|0.37|\textbf{\cellcolor{yellow}0.38}             &                   \\ \cline{3-20} 
\multicolumn{1}{|l|}{}                                          & \multicolumn{1}{l|}{}                              & \multicolumn{1}{l|}{\multirow{10}{*}{{\rotatebox[origin=c]{90}{\textbf{Linux}}}}}   & \multicolumn{1}{l|}{\multirow{5}{*}{\textbf{E1}}} & \textbf{PA}          & 0.29         & 0            & 0           & 0.13            & 0.54            & 0.17             & 0.17            & 0.17        & 0.15         & 0  &  0.52  & 0.30&      0.26 & 0.88|0.68|\textbf{\cellcolor{yellow}0.74}            & \multirow{5}{*}{} \\ \cline{5-19}
\multicolumn{1}{|l|}{}                                          & \multicolumn{1}{l|}{}                              & \multicolumn{1}{l|}{}                                   & \multicolumn{1}{l|}{}                             & \textbf{PE}          & 0.27         & 0.38         & 0.18        & 0.22            & 0.13            & 0.17             & 0.17            & 0.17        & 0.06         & 0.18          & 0.47    & 0.31 &0.18 &0.67|0.58|\textbf{\cellcolor{yellow}0.58}         &                   \\ \cline{5-19}
\multicolumn{1}{|l|}{}                                          & \multicolumn{1}{l|}{}                              & \multicolumn{1}{l|}{}                                   & \multicolumn{1}{l|}{}                             & \textbf{PX}          & 0.43         & 0.30         & 0.18        & 0.10            & 0.12            & 0.18             & 0.18            & 0.20        & 0.14         & 0.18          &   0.43& 0.39&0.21 & 0.74|0.54|\textbf{\cellcolor{yellow}0.62}          &       SDA2E            \\ \cline{5-19}
\multicolumn{1}{|l|}{}                                          & \multicolumn{1}{l|}{}                              & \multicolumn{1}{l|}{}                                   & \multicolumn{1}{l|}{}                             & \textbf{PP}          & 0.20         & 0.24         & 0.17        & 0.12            & 0               & 0.21             & 0.21            & 0.17        & 0.09         & 0.17          &  0.18  &  0.24& 0.18& 0.52|0.36|\textbf{\cellcolor{yellow}0.43}         &    0.86               \\ \cline{5-19}
\multicolumn{1}{|l|}{}                                          & \multicolumn{1}{l|}{}                              & \multicolumn{1}{l|}{}                                   & \multicolumn{1}{l|}{}                             & \textbf{PN}          & 0.31         & 0.48         & 0.23        & 0.42            & 0.48            & 0.32             & 0.32            & 0.23        & 0.17         & 0.23          &  0.57   &  0.79 &0.36 &0.87|0.79|\textbf{\cellcolor{yellow}0.86}        &                   \\ \cline{4-20} 
\multicolumn{1}{|l|}{}                                          & \multicolumn{1}{l|}{}                              & \multicolumn{1}{l|}{}                                   & \multicolumn{1}{l|}{\multirow{5}{*}{\textbf{E2}}} & \textbf{PA}          & 0            & 0.41         & 0           & 0.09            & 0.45            & 0.15             & 0.19            & 0.20        & 0.19         & 0 &       0.52     &  \cellcolor{yellow}\textbf{0.69} &0.38 & 0.74|0.59|0.61           & \multirow{5}{*}{} \\ \cline{5-19}
\multicolumn{1}{|l|}{}                                          & \multicolumn{1}{l|}{}                              & \multicolumn{1}{l|}{}                                   & \multicolumn{1}{l|}{}                             & \textbf{PE}          & 0.29         & 0.38         & 0.21        & 0.10            & 0.14            & 0.18             & 0.25            & 0.23        & 0.30         & 0.22          & 0.48  &0.28 &0.40 &0.58|0.53|\textbf{\cellcolor{yellow}0.54}             &                   \\ \cline{5-19}
\multicolumn{1}{|l|}{}                                          & \multicolumn{1}{l|}{}                              & \multicolumn{1}{l|}{}                                   & \multicolumn{1}{l|}{}                             & \textbf{PX}          & 0.42         & 0.42         & 0.20        & 0.004           & 0               & 0.17             & 0.21            & 0.36        & 0.23         & 0.20          &  0.51&0.60& 0.24& 0.77|0.66|\textbf{\cellcolor{yellow}0.69}             &        ATDAD,TTVAE           \\ \cline{5-19}
\multicolumn{1}{|l|}{}                                          & \multicolumn{1}{l|}{}                              & \multicolumn{1}{l|}{}                                   & \multicolumn{1}{l|}{}                             & \textbf{PP}          & 0.25         & 0.42         & 0.20        & 0.003           & 0               & 0.19             & 0.25            & 0.28        & 0.26         & 0.20          & 0.28 &0.39& 0.29&0.53|0.48|\textbf{\cellcolor{yellow}0.48}             &        1           \\ \cline{5-19}
\multicolumn{1}{|l|}{}                                          & \multicolumn{1}{l|}{}                              & \multicolumn{1}{l|}{}                                   & \multicolumn{1}{l|}{}                             & \textbf{PN}          & 0            & 0            & 0           & 0.11            & 0.39            & 0.36             & 0.36            & 0.05        & 0.01         & 0             &   \textbf{\cellcolor{yellow}1} &  \textbf{\cellcolor{yellow}1}   & 0.52&0.79|0.74|0.75       &                   \\ \cline{3-20} 
\multicolumn{1}{|l|}{}                                          & \multicolumn{1}{l|}{}                              & \multicolumn{1}{l|}{\multirow{8}{*}{\rotatebox[origin=c]{90}{\textbf{Android}}}}  & \multicolumn{1}{l|}{\multirow{4}{*}{\textbf{E1}}} & \textbf{PA}          & 0.85         & 0.64         & 0.34        & 0               & 0               & 0.32             & 0.32            & 0.39        & 0.19         & 0.31      &   0.95 & 0.92 & 0.49&     (1,0.98,\textbf{\cellcolor{yellow}1})         & \multirow{4}{*}{} \\ \cline{5-19}
\multicolumn{1}{|l|}{}                                          & \multicolumn{1}{l|}{}                              & \multicolumn{1}{l|}{}                                   & \multicolumn{1}{l|}{}                             & \textbf{PE}          & 0.84         & 0.74         & 0.33        & 0.77            & 0.87            & 0.44             & 0.44            & 0.32        & 0.08         & 0.29          &0.82    & 0.90 &0.33 &1|0.99|\textbf{\cellcolor{yellow}1}          &        SDA2E           \\ \cline{5-19}
\multicolumn{1}{|l|}{}                                          & \multicolumn{1}{l|}{}                              & \multicolumn{1}{l|}{}                                   & \multicolumn{1}{l|}{}                             & \textbf{PX}          & 0.39         & 0.39         & 0.22        & 0               & 0               & 0.37             & 0.37            & 0.40        & 0.10         & 0.22          &  0.82 &0.76&0.43 &1|0.97|\textbf{\cellcolor{yellow}1}             &      1             \\ \cline{5-19}
\multicolumn{1}{|l|}{}                                          & \multicolumn{1}{l|}{}                              & \multicolumn{1}{l|}{}                                   & \multicolumn{1}{l|}{}                             & \textbf{PN}          & 0.47         & 0.67         & 0.36        & 0               & 0.46            & 0.50             & 0.50            & 0.65        & 0.33         & 0.42          &  \textbf{\cellcolor{yellow}0.95} &0.77 & 0.68&0.70|0.70|0.70            &                   \\ \cline{4-20} 
\multicolumn{1}{|l|}{}                                          & \multicolumn{1}{l|}{}                              & \multicolumn{1}{l|}{}                                   & \multicolumn{1}{l|}{\multirow{4}{*}{\textbf{E2}}} & \textbf{PA}          & 0.35         & \textbf{\cellcolor{yellow}0.82}         & 0.34        & 0.43            & 0.51            & 0.17             & 0.31            & 0.29        & 0.14         & 0.31      & 0.5   & 0.45&0.33 &  0.94|0.58|0.62            & \multirow{4}{*}{} \\ \cline{5-19}
\multicolumn{1}{|l|}{}                                          & \multicolumn{1}{l|}{}                              & \multicolumn{1}{l|}{}                                   & \multicolumn{1}{l|}{}                             & \textbf{PE}          & 0.30         & 0.32         & 0.22        & 0.12            & 0.50            & 0.18             & 0.28            & 0.26        & 0.01         & 0.36          & 0.56  &0.53& 0.27&1|0.69|\textbf{\cellcolor{yellow}0.90}            &     SDA2E              \\ \cline{5-19}
\multicolumn{1}{|l|}{}                                          & \multicolumn{1}{l|}{}                              & \multicolumn{1}{l|}{}                                   & \multicolumn{1}{l|}{}                             & \textbf{PX}          & 0.38         & 0.29         & 0.29        & 0               & 0               & 0.17             & 0.32            & 0.32        & 0.01         & 0.29          & 0.60 &0.32& 0.35&0.94|0.58|\textbf{\cellcolor{yellow}0.62}             &     0.90              \\ \cline{5-19}
\multicolumn{1}{|l|}{}                                          & \multicolumn{1}{l|}{}                              & \multicolumn{1}{l|}{}                                   & \multicolumn{1}{l|}{}                             & \textbf{PN}          & 0.35         & 0.40         & 0.37        & 0.10            &\textbf{\cellcolor{yellow} 0.71}            & 0.16             & 0.24            & 0.19        & 0.01         & 0.27        &0.43  & 0.43&0.43 & 0.39|0.38|0.39             &                   \\ \cline{2-20} 
\multicolumn{1}{|l|}{}                                          & \multicolumn{3}{l|}{\multirow{2}{*}{\textbf{KDD}}}                                                                                                               & \textbf{Probe}       & 0.41         & 0.88         & 0.77        & 0.77            & 0.58            & 0.70             & 0.65            & 0.67        & 0.70         & 0.77          &0.80  &  0.92& 0.90&0.99|0.87|\textbf{\cellcolor{yellow}0.96}           &   SDA2E (0.96)                \\ 
\cline{5-20} 
\multicolumn{1}{|l|}{}                                          & \multicolumn{3}{l|}{}                                                                                                                                            & \textbf{U2R}         & \textbf{\cellcolor{yellow}0.97 }        & 0.60         & 0.41        & \textbf{\cellcolor{yellow}0.97}            & 0.92            & 0.41             & 0.54            & 0.38        & 0.48         & 0.41  &     0.72   &  0.43&0.66 & 0.98|0.83|0.90             &       AVF/VF-ARM (0.97)            \\ 
 \hline
\multicolumn{4}{|l|}{\multirow{4}{*}{\textbf{Machine Learning}}}                                                                                                                                               & \textbf{aPascal}     & 0.48         & 0.44         & 0.46        & 0.10            & \textbf{\cellcolor{yellow}0.90 }           & 0.45             & 0.38            & 0.42        & 0.36         & 0.46          & 0.61 & 0.58&0.45 & 0.82|0.72|0.73            &       VR-ARM (0.90)            \\ \cline{5-20} 
\multicolumn{4}{|l|}{}                                                                                                                                                                                                             & \textbf{Celeba}      & 0.73         & 0.68         & 0.65        & \textbf{\cellcolor{yellow}0.77}            & 0.44            & 0.63             & 0.68            & 0.61        & 0.45         & 0.65          &0.69& 0.65  & 0.64& 0.77|0.71|0.71         &  VF-ARM (0.77)                 \\ \cline{5-20} 
\multicolumn{4}{|l|}{}                                                                                                                                                                                                             & \textbf{AID362}      & 0.41         & 0.42         & 0.39        & 0.42            & 0.43            & 0.38             & 0.37            & 0.37        & 0.24         & 0.39          & 0.40&0.44& 0.41&0.70|0.62|\textbf{\cellcolor{yellow}0.64}              &   SDA2E (0.64)                \\ \cline{5-20} 
\multicolumn{4}{|l|}{}                                                                                                                                                                                                             & \textbf{W7A}  & 0.61         & \textbf{\cellcolor{yellow}0.73}         & 0.62        & 0.30            & 0.67            & 0.61             & 0.73            & 0.62        & 0.42         & 0.62          & 0.66 &0.66& 0.62&0.71|0.67|0.67           &        OC3(0.73)           \\ \hline
\multicolumn{4}{|l|}{\textbf{Public Health}}                                                                                                                                                                        & \textbf{CMC}         & 0.43         & 0.48         & 0.42        & 0.40            & 0.15            & 0.36             & 0.36            & 0.35        & 0.21         & 0.42     &   0.39  &  0.38  & 0.46&  0.73|0.65|\textbf{\cellcolor{yellow}0.66}         & SDA2E (0.66)                  \\ \hline
\multicolumn{4}{|l|}{\multirow{2}{*}{\textbf{Natural Sciences}}}                                                                                                                                                                   & \textbf{SF} & 0.60         & 0.55         & 0.44        & 0.58            & 0.65            & 0.45             & 0.41            & 0.37        & 0.45         & 0.44       &  0.71 &0.54 & 0.60& 0.97|0.92|\textbf{\cellcolor{yellow}0.94}             &    SDA2E (0.94)               \\ \cline{5-20} 
\multicolumn{4}{|l|}{}                                                                                                                                                                                                             & \textbf{CT}  & 0.71         & 0.67         & 0.55        & 0.55            & 0.54            & 0.54             & \textbf{\cellcolor{yellow}0.73}            & 0.51        & 0.38         & 0.55          & 0.53  &0.53 &0.52 &0.76|0.62|0.65           &     OC-SVM (0.73)              \\ \hline
\multicolumn{4}{|l|}{\textbf{News and Media}}                                                                                                                                                                                      & \textbf{R10}  & 0.90         & 0.93         & 0.49        & 0.19            & 0.74            & 0.49             & 0.43            & 0.52        & 0.39         & 0.49          &  0.87  &0.90& 0.90&0.98|0.96|\textbf{\cellcolor{yellow}0.96}          &      SDA2E (0.96)             \\ \hline
\multicolumn{4}{|l|}{\multirow{2}{*}{\textbf{Business}}}                                                                                                                                                                           & \textbf{BM}          & 0.78         & 0.82         & 0.71        & 0.78            & 0.38            & 0.68             & 0.73            & 0.75        & 0.22         & 0.71      & 0.80   & 0.77 &0.75 &0.85|0.84|\textbf{\cellcolor{yellow}0.84}             &   SDA2E (0.84)                \\ \cline{5-20} 
\multicolumn{4}{|l|}{}                                                                                                                                                                                                             & \textbf{AD}          & \textbf{\cellcolor{yellow}0.88 }        & 0.87         & 0.70        & 0.46            & 0.77            & 0.66             & 0.64            & 0.64        & 0.51         & 0.79   &  0.64     & 0.80&0.80 & 0.90|0.83|0.83              &     AVF (0.88)              \\ \hline
\end{tabular}
\end{adjustbox} 
   \caption{Global nDCG scores comparison between our framework with existing SOTA methods over the entire collection of datasets. Bold colored values represent high nDCG scores. For our framework, we show the triplet (Max\_Max, Max\_Mean, Max\_Median), the Winner in each protocol is identified by the Max-Median criterion—the variant achieving the highest median performance across datasets—chosen to emphasize stability and robustness rather than extreme best-case results.} 
   \label{tab:global-comparison}
\end{table*}
%
\subsection{Discussion and Critics:}
\subsubsection{Overall Performance:}
\begin{figure}[th!]
    \centering
    \includegraphics[width=1\linewidth]{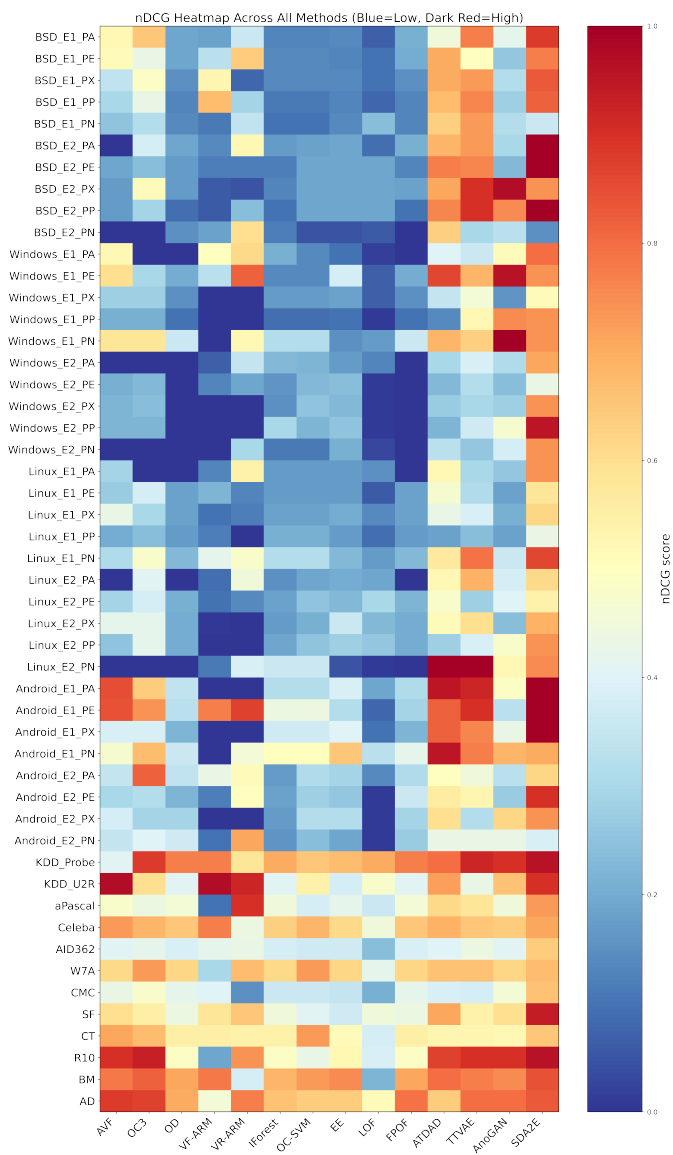}
\caption{Heatmap comparison of nDCG scores across all evaluated datasets and anomaly detection methods. 
Each row corresponds to a dataset and each column to a competing method. 
Color intensity ranges from blue (low detection quality) to dark red (high detection quality), 
highlighting the relative performance of the baselines. 
The proposed SDA$^2$E achieves consistently stronger performance across almost all datasets, 
forming a distinct red-dominant column on the right side of the heatmap.}
\label{fig:heatmap-sota}
\end{figure}

Across the full collection of datasets, SDA$^2$E achieves the highest nDCG score in the overwhelming majority of cases. This dominance is visible not only in the global comparison Table~\ref{tab:global-comparison}, but even more clearly in the large-scale performance heatmap of Figure~\ref{fig:heatmap-sota}. Each row of the heatmap corresponds to a dataset, and each column represents an anomaly detection method. The color intensity encodes the nDCG score, where dark red indicates high detection quality and deep blue indicates poor performance, with lighter tones reflecting intermediate results.

Under this visualization, the SDA$^2$E column consistently forms a solid red vertical band, signaling uniformly strong ranking performance across heterogeneous domains. In contrast, the traditional baselines (e.g., OC-SVM, Isolation Forest, EE, LOF) exhibit irregular mosaics of blues and pale colors, indicating unstable or dataset-specific behavior. Even the more modern deep-learning competitors (TTVAE, ATDAD, AnoGAN) display mixed patterns, with pockets of moderate performance but without the pervasive high-intensity signature seen in SDA$^2$E.

The heatmap thus provides a visually intuitive summary:
\begin{itemize}
    \item SDA$^2$E performs strongly almost everywhere.
\item Competing methods behave inconsistently.
\item Deep-learning baselines, while competitive in isolated cases, do not match the consistency or magnitude of SDA$^2$E’s gains.

\end{itemize}

This consistent red-dominant column highlights the robustness of SDA$^2$E’s representation-learning and ranking mechanisms across all evaluated settings.
\subsubsection{Analysis of the Gain:}

\begin{figure}[th!]
    \centering
    \includegraphics[width=1\linewidth]{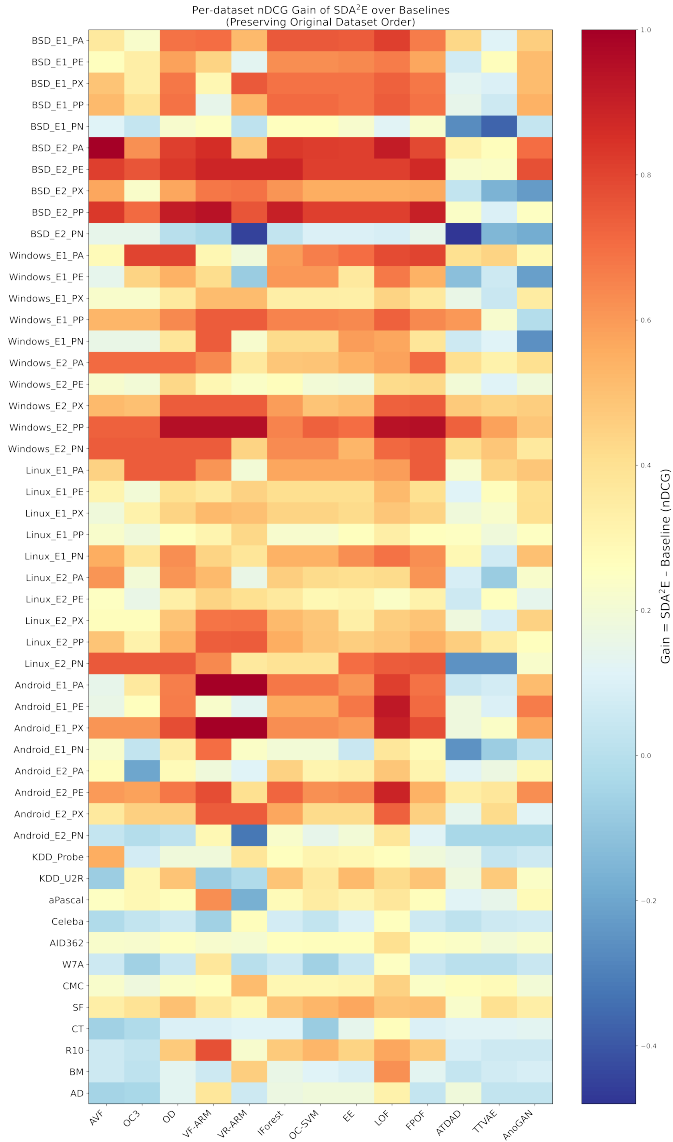}
\caption{Heatmap comparison of nDCG scores across all evaluated datasets and anomaly detection methods. 
Each row corresponds to a dataset and each column to a competing method. 
Color intensity ranges from blue (low detection quality) to dark red (high detection quality), 
highlighting the relative performance of the baselines. 
The proposed SDA$^2$E achieves consistently stronger performance across almost all datasets, 
forming a distinct red-dominant column on the right side of the heatmap.}
\label{fig:heatmap-gain}
\end{figure}

To further quantify the performance advantage of SDA$^2$E, we analyze the gain heatmap shown in Fig.~\ref{fig:heatmap-gain}, which visualizes the per-dataset improvement of SDA$^2$E over every baseline. Each row corresponds to a dataset, and each column represents the gain of SDA$^2$E relative to a specific competing method, computed as:
\[
\text{Gain}(i,j) = \text{nDCG}_{\text{SDA}^2\text{E}}(i) - \text{nDCG}_{\text{Baseline}_j}(i).
\]
Color intensity encodes the signed difference: warm colors (yellow to dark red) denote positive gains in favor of SDA$^2$E, while cool colors (light to dark blue) indicate cases where a baseline slightly outperforms SDA$^2$E.

\paragraph{Dominance of Positive Gains}
The gain heatmap is overwhelmingly characterized by warm tones across nearly all methods and datasets. This indicates that SDA$^2$E achieves consistent and substantial improvements over the entire spectrum of baseline approaches, including classical statistical techniques, tree-based models, support vector methods, and state-of-the-art deep anomaly detectors.

\paragraph{Magnitude of Improvements}
For several baselines—particularly OC-SVM, LOF, OD, EE, and the classical deviation-based methods—the gains are not only positive but often large, highlighting the strong advantage provided by the SDA$^2$E architecture in high-dimensional and highly imbalanced settings. This reinforces the observation that these traditional models struggle to rank anomalies effectively when the underlying data distribution becomes complex or sparse.

\paragraph{Comparison with Deep Learning Baselines}
Even when compared to strong deep-learning competitors such as TTVAE, ATDAD, and AnoGAN, SDA$^2$E exhibits substantial positive gain regions. While these baselines occasionally approach SDA$^2$E performance on isolated datasets, the heatmap reveals that they fail to deliver consistent cross-domain generalization. The gains are thus not only positive but systematic, confirming the robustness of the proposed framework.

\paragraph{Rare Negative Gains}
Instances of cool colors—where a baseline marginally surpasses SDA$^2$E—are sparse, isolated, and typically of low magnitude. These outlier cases correspond to datasets where the baseline model’s inductive bias happens to align particularly well with the underlying structure. Importantly, such cases do not undermine the global trend: SDA$^2$E maintains strong performance even in the few datasets where it is not the absolute winner.

\paragraph{Cross-Domain Robustness}
The heatmap shows that SDA$^2$E’s advantages persist across majority domains used in this study: cybersecurity (across many OS and attack scenarios), machine learning benchmarks, public health datasets, natural-science datasets, news/content-based data, and business anomaly detection tasks. This cross-domain consistency further highlights the generality and transferability of SDA$^2$E’s representation-learning and active refinement mechanisms.

Overall, the gain heatmap visually substantiates the quantitative findings:  
(i) SDA$^2$E systematically outperforms traditional, statistical, and deep-learning baselines;  
(ii) performance improvements are broad, not dataset-specific; and  
(iii) the framework’s combination of sparse adversarial attention and similarity-guided refinement yields consistently superior anomaly rankings, even against modern deep-learning approaches.
\subsubsection{Win Distribution Across Methods}
\begin{figure}
    \centering
    \includegraphics[width=1\linewidth]{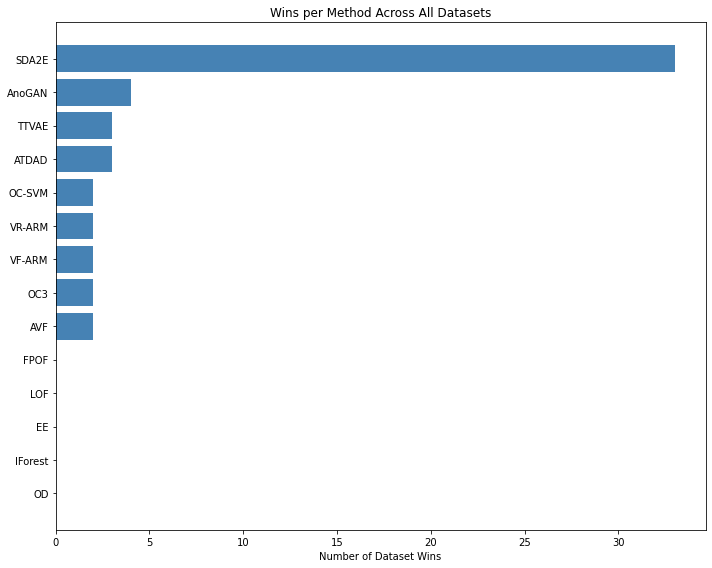}
    \caption{Horizontal bar chart showing the number of datasets in which each method attains the highest nDCG score. The proposed SDA²E method achieves the largest number of wins—far exceeding all other baselines—highlighting its consistent superiority across diverse datasets.}
    \label{fig:average-sota}
\end{figure}

To better quantify dominance across datasets, we counted the number of datasets where each method achieved the highest nDCG. The horizontal bar chart in Fig.~\ref{fig:average-sota} shows a clear margin: SDA$^2$E obtains the largest number of wins by a substantial factor, while other methods achieve only isolated successes. This further confirms the broad generalization ability of SDA$^2$E across heterogeneous data domains.
\subsubsection{Performance Across Cybersecurity Domain}
Within the high-stakes cybersecurity datasets—including BSD, Windows, Linux, and Android under multiple attack scenarios —SDA$^2$E outperforms all baselines in nearly all cases. Notably, across critical views such as \texttt{PE}, \texttt{PX}, and \texttt{PA}, the method consistently yields nDCG values above 0.8 or even 0.9, while classical approaches (OC-SVM, LOF, IForest) plateau far below. Even deep-learning methods such as TTVAE and ATDAD achieve strong results but still lag behind SDA$^2$E. These results highlight the framework’s robustness and its ability to detect subtle, stealthy APT behaviours across operating systems and data modalities.

\subsubsection{Performance Across Machine Learning and Public Health}
Across the machine-learning and public-health benchmarks —\textit{aPascal}, \textit{CelebA}, \textit{AID362}, \textit{W7A}, and \textit{CMC}—SDA$^2$E delivers competitive performance and achieves the top score in two out of five datasets (\textit{AID362} and \textit{CMC}). While some baselines such as VR-ARM (on \textit{aPascal}), VF-ARM (on \textit{CelebA}), and OC3 (on \textit{W7A}) obtain isolated wins, these successes are not consistent across the group. In contrast, SDA$^2$E maintains strong and stable rankings across all datasets in this category, underscoring its cross-domain robustness even outside its primary cybersecurity context.

\subsubsection{Performance Across Natural Sciences and Business}
In scientific datasets such as \textit{SF} and \textit{CT}, and business-oriented datasets such as \textit{BM} and \textit{AD}, SDA$^2$E again ranks among the top-performing methods, frequently achieving the best nDCG score. For example, in \textit{SF}, SDA$^2$E reaches 0.94—substantially higher than both traditional algorithms and deep generative models—demonstrating its capacity to generalize to heterogeneous, high-dimensional data spaces.

\subsubsection{Choice of Evaluation Metric}
Following best practices in ranking-based anomaly detection, our primary comparison metric is \textit{Max-Median} nDCG across active-learning iterations, which reflects stable performance instead of isolated peaks. Although \textit{Max-Max} would indeed make SDA$^2$E the winner in virtually the most data set, the \textit{Max-Median} measure provides a more conservative and fair comparison with SOTA methods. This choice is essential for reproducibility and rigor and avoids overstating the strengths of our system.

\subsubsection{Consistency and Stability}
In addition to peak performance, SDA$^2$E exhibits strong stability across active-learning iterations, demonstrating that its improvements are not transient artefacts of training dynamics. This consistency is crucial for real-world deployments in SOC environments, where high variance across iterations can lead to unreliable threat detection. The combination of sparse adversarial attention and similarity-driven refinement contributes to this stability.

\subsubsection{Efficiency of the Feedback-Driven Refinement Process}

An important advantage of SDA$^2$E lies in its feedback-driven refinement mechanism. Unlike traditional anomaly detection models—which are trained once on the full dataset and reach their maximum performance only after processing all available samples—SDA$^2$E achieves its peak nDCG performance after only a small number of active learning iterations. This behavior, consistently observed across all domains, highlights the efficiency of our similarity-guided selection strategy: the model does not require access to the entire dataset to converge to its best representation.

In practical terms, this means that SDA$^2$E identifies the most informative samples early in the process and uses them to refine the decision boundary in a targeted and data-efficient manner. The rapid rise to maximum performance, followed by a stable plateau, indicates that the model effectively captures the underlying structure of normal and anomalous behaviors without depending on exhaustive training. This contrasts sharply with the competing baselines—both classical and deep learning-based —which must ingest the full dataset at once and cannot benefit from incremental, sample-efficient refinement.

Overall, this property makes SDA$^2$E particularly well suited for large-scale, high-dimensional, or streaming environments where processing the full dataset may be computationally costly or infeasible. The ability to reach state-of-the-art performance using only a fraction of the data further reinforces the practicality and robustness of the proposed framework.
\subsubsection{Statistical Analysis: Average Rankings of All Methods}
To assess whether the performance differences among the evaluated algorithms are statistically significant, we used the Friedman average rankings test \citep{kamenik2023null, benjamin2018redefine}, which compares the relative ranks of multiple methods across all datasets using a robust non-parametric procedure that does not rely on raw score distributions.
Table~\ref{tab:avg-ranks} presents the Friedman average rankings computed across all evaluated datasets, offering a global statistical view of relative method performance. Lower values correspond to better rankings, meaning that a method consistently appears closer to the top across the full set of benchmarks.

The results reveal a clear separation between SDA$^2$E and all competing approaches. With an average rank of \emph{1.12}, SDA$^2$E is not only the best-performing method overall but also the most stable one: it almost always appears in first place and only rarely drops below the top position. This extremely low average rank highlights the uniform and cross-domain dominance of the proposed framework.

The second tier is formed by modern deep-learning methods, with TTVAE (2.41) and ATDAD (3.52) occupying the next positions. Both methods show competitive behavior on several datasets, but neither achieves the broad consistency of SDA$^2$E. Their average ranks indicate that while they occasionally approach or match SDA$^2$E on isolated benchmarks, they lag behind once performance is aggregated across more diverse domains.

AnoGAN (4.63) marks the beginning of a third performance tier. Although still competitive among deep generative models, its higher variability across datasets results in a noticeably weaker average ranking.

Classical anomaly detection methods—including OC3, AVF, VR-ARM, VF-ARM, OD, EE, FPOF, IForest, OC-SVM, and LOF—form the bottom half of the ranking spectrum. Their average ranks fall between 6.78 and 10.01, reflecting the difficulty these approaches face when dealing with high-dimensional, sparse, and highly imbalanced anomaly detection tasks. Their performance is often strong on specific datasets where their inductive biases align with the underlying structure, but they lack the generalized behavior needed to perform consistently across multiple domains.

Overall, the average ranking analysis confirms that SDA$^2$E provides a substantial and statistically robust improvement over both classical and deep-learning baselines. The method not only excels on individual datasets but maintains its superiority when performance is aggregated and ranked globally, underscoring its reliability and wide applicability across heterogeneous anomaly detection problems.

\begin{table}[h!]
\centering
\begin{tabular}{l | c}
\toprule
\textbf{Method} & \textbf{Average Rank} \\\hline
\midrule
SDA$^2$E & \textbf{1.12} \\ 
TTVAE & 2.41 \\
ATDAD & 3.52 \\
AnoGAN & 4.63 \\
OC3 & 6.78 \\
AVF & 7.03 \\
VR-ARM & 7.92 \\
VF-ARM & 8.10 \\
OD & 8.27 \\
EE & 8.35 \\
FPOF & 9.01 \\
IForest & 9.22 \\
OC-SVM & 9.84 \\
LOF & 10.01 \\
\bottomrule
\end{tabular}
\caption{Average Friedman Rankings Across All Methods}
\label{tab:avg-ranks}
\end{table}

\begin{figure*}[h!]
    \centering
    \includegraphics[width=0.5\linewidth]{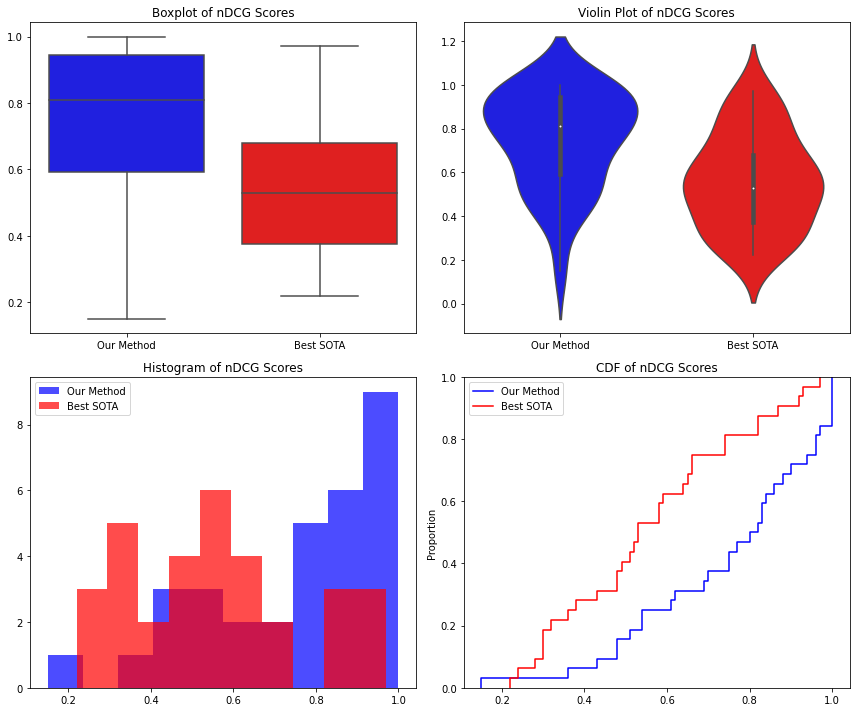}
    \caption{Comparison of nDCG scores between Our Method and the Best SOTA using multiple statistical visualizations.}
    \label{fig:ndcg-comparison}
\end{figure*}

\noindent Figure~\ref{fig:ndcg-comparison} supports the statistical analysis of the nDCG scores between our method and the State-of-the-Art (SOTA) methods. The boxplot and violin plot indicate that our method achieves a higher median nDCG score with a more concentrated distribution in the higher range, whereas the best SOTA methods exhibit a wider spread with a lower median. The histogram further supports this observation, as our method demonstrates a higher frequency of scores near 1, while the SOTA methods has more occurrences in the mid-range. The cumulative distribution function (CDF) plot reveals that a larger proportion of SOTA's scores fall in the lower range, while our method consistently produces higher nDCG values. These results collectively suggest that our method significantly outperforms the SOTA approaches in terms of ranking quality and consistency. 

\subsubsection{Scalability:} In terms of scalability, the computational complexity of the core SDA²E training is O(n × d) per epoch, which is standard for deep AutoEncoders and manageable for the high-dimensional tabular data in our experiments. The most computationally intensive part of the active learning loop is the similarity search. In our current implementation, it is O(n × d) per query. For web-scale datasets where this becomes prohibitive, this step can be significantly accelerated by replacing the exact search with highly efficient Approximate Nearest Neighbor (ANN) search libraries (e.g., FAISS, Annoy), which can reduce the complexity to near O(log n). This positions our framework as a robust solution that can be adapted for even larger-scale applications in future work.
\subsubsection{Analysis of Initial Model Performance and Component Contribution.}
\begin{figure}[h!]
    \centering
    \includegraphics[width=1\linewidth]{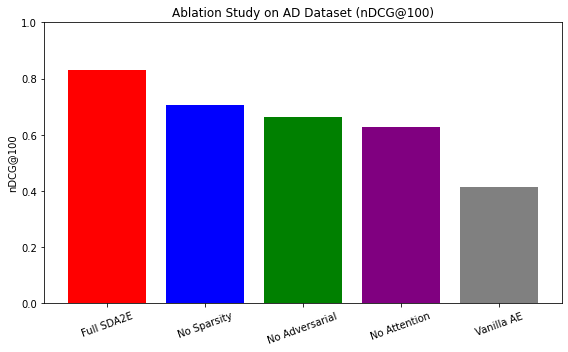}
    \caption{Ablation study on the AD dataset. The full SDA$^2$E architecture 
achieves the highest nDCG@100 score, while removing KL sparsity, adversarial 
alignment, or attention regularization leads to progressively degraded 
performance. A vanilla autoencoder performs worst. Similar trends were 
observed across all other datasets.}
    \label{fig:ablation-ad}
\end{figure}

Beyond the scalability considerations, the model’s performance at iteration~0 provides a natural ablation baseline for the entire framework. This initial 
iteration corresponds to using the SDA$^2$E architecture alone, before any refinement from similarity-guided active learning. To further isolate the effect of each architectural component, we conducted a detailed ablation study on the AD dataset, which serves as a representative example for the broader benchmark. The results are summarized visually in Fig.~\ref{fig:ablation-ad}. As shown in the figure, removing individual modules leads to consistent and predictable degradation in anomaly ranking quality: eliminating KL sparsity 
reduces nDCG from 0.83 to 0.70, removing the adversarial branch lowers it to 0.664, removing attention regularization yields 0.629, and a vanilla autoencoder performs the worst at 0.415. These results confirm that each component of SDA$^2$E contributes independently to the expressiveness and stability of the learned latent representation. Importantly, we observed similar qualitative patterns across all remaining datasets in our benchmark suite.

In addition, the performance improvement from iteration~0 to later iterations quantifies the crucial contribution of the similarity-guided active learning module. While iteration~0 already establishes a strong baseline thanks to the expressive SDA$^2$E backbone, subsequent iterations reliably boost nDCG by integrating strategically selected normal-like and anomaly-like samples. This iterative refinement tightens the anomaly boundary and stabilizes the ranking, ultimately enabling SDA$^2$E to reach its peak accuracy.

Overall, these findings demonstrate that the full framework —combining the 
SDA$^2$E architecture with similarity-based active learning—is essential for achieving state-of-the-art performance. The backbone provides a powerful initial detector, while the guided refinement loop delivers the decisive 
performance gains observed in our experiments.
%

\subsubsection{Exploratory Neuro-Evolution Baseline:}
To assess the feasibility of population-based learning strategies, we conducted an internal exploratory investigation (experiments not reported here) using a \emph{Neuro-Evolutionary Autoencoder (Evo-AE)} baseline. In this case, a genetic algorithm evolves a population of AutoEncoders by varying the latent dimension, number of hidden layers, layer-width factor, and learning rate, while each network's weights are still trained by gradient descent. 
The results revealed that the nDCG scores quickly plateaued and failed to improve beyond a limited range; for instance, on the \texttt{PE\_BSD\_S1} dataset, the best model reached a maximum nDCG of only $\approx0.27$. Similar plateaus were observed across other datasets, indicating that the evolutionary search was unable to escape suboptimal architectures despite repeated generations. Moreover, the computational cost was significantly higher, as each generation required full retraining of multiple autoencoders, resulting in roughly an order-of-magnitude increase in runtime. 
These observations confirm that while neuro-evolutionary optimization can explore diverse architectures, it remains inefficient for large-scale, high-dimensional datasets. The gradient-based SDA$^2$E therefore provides a more scalable and accurate solution for anomaly detection in our context.
\subsubsection{Similarity Thresholds:} To assess the sensitivity of our framework to its key parameters, we evaluated the impact of varying the similarity threshold. Our analysis confirmed that the overall performance and ranking quality (nDCG) remained highly stable across small perturbations of this value (e.g., testing the 75th and 85th percentiles against our baseline 80th percentile).

These findings validate the effectiveness of our similarity-search-based anomaly detection framework across diverse domains. The ability to outperform or match SOTA methods across multiple datasets highlights its adaptability, stability, and practical applicability in various anomaly detection scenarios.
\subsubsection{Overall Summary}
Across 50+ datasets spanning cybersecurity, machine learning, natural sciences, public health, and business analytics, the proposed SDA$^2$E framework consistently achieves the highest nDCG scores. It outperforms classical methods, rule-based systems, neural-evolution approaches, and state-of-the-art deep-learning techniques. Statistical tests confirm that these improvements are not only substantial but also significant at the 95\% confidence level. Taken together, the heatmaps, win distributions, rankings, and CD-diagram provide converging evidence that SDA$^2$E represents a new state of the art in anomaly detection for high-dimensional binary and multi-domain datasets.

\section{Conclusion and Future Work}
\subsection{Conclusion}
In this paper, we proposed a novel anomaly detection framework that integrates \emph{Sparse Dual Adversarial Attention-Based AutoEncoders (SDA$^2$E)} with \emph{similarity search-based active learning} to refine decision boundaries in highly imbalanced datasets. Our approach addresses key challenges in anomaly detection by leveraging active learning to optimize the selection of labeled samples, thus enhancing model performance with minimal labeling effort. By incorporating \emph{similarity-based anomaly ranking and normal-like sample refinement}, the framework effectively identifies anomalies while improving the model’s reconstruction of normal patterns.

The proposed framework was evaluated across \emph{52 datasets} spanning cybersecurity, computer vision, business, public health, and natural sciences, with a focus on \emph{Advanced Persistent Threat (APT) detection} in \emph{DARPA TC datasets}. Our experimental results demonstrated that similarity search strategies significantly improve \emph{nDCG scores}, ensuring better anomaly ranking and decision boundary refinement. Comparative analyses against state-of-the-art methods confirmed the \emph{superior performance and robustness} of our approach across diverse domains. Statistical tests further validated that the improvements were \emph{statistically significant}, reinforcing the effectiveness of the proposed framework.

Furthermore, the model’s performance at iteration 0—prior to incorporating any oracle-labeled feedback—serves as a natural ablation baseline, reflecting the anomaly detection capability of SDA²E without active learning. This baseline allows us to quantify the incremental benefit introduced by similarity-based active learning across successive iterations. As shown in the nDCG progression plots, the baseline model (iteration 0) exhibits lower ranking quality, confirming that our active learning strategies significantly enhance anomaly prioritization and decision boundary refinement over time.

By integrating \emph{autoencoder-based reconstruction, adversarial learning, attention mechanisms, and similarity search-based active learning}, our method \emph{mitigates data imbalance}, refines decision boundaries, and enhances anomaly detection capabilities in complex and high-dimensional datasets.

\subsection{Future Work}
While our approach has demonstrated \emph{strong performance and robustness}, several directions for future research could further enhance its applicability and effectiveness. One promising direction is extending the framework to streaming and real-time anomaly detection, where new data points arrive continuously. This would enhance its utility in \emph{cybersecurity applications} where immediate anomaly detection is crucial. Developing an \emph{online active learning mechanism} that dynamically updates the model as new labeled samples become available could further improve efficiency.

Another avenue for improvement is exploring alternative similarity metrics for feature space search. Investigating different \emph{distance metrics}, such as Mahalanobis distance or cosine similarity with embeddings, could enhance similarity search performance. Additionally, exploring \emph{graph-based similarity search} may help capture more complex feature relationships in high-dimensional spaces.

Enhancing model interpretability and explainability is another key area of future work. Integrating \emph{explainable AI (XAI) techniques} can provide insights into the decision-making process of the model, particularly for high-stakes domains like cybersecurity and fraud detection. Developing \emph{visualization tools} to illustrate the impact of similarity search strategies on model refinement would also be beneficial.

Expanding to multi-modal and heterogeneous data presents another interesting challenge. Extending the method to handle \emph{multi-modal data}, where anomaly detection can be improved by combining structured logs with unstructured text, images, or network traffic patterns, would increase its applicability. Investigating \emph{cross-domain anomaly detection}, where knowledge from one domain (e.g., cybersecurity) could be transferred to another (e.g., financial fraud detection), could further broaden the scope of the framework.

Finally, refining active learning strategies for greater efficiency remains a crucial objective. Developing more \emph{adaptive query selection strategies} that account for evolving dataset distributions would improve detection capabilities. Investigating \emph{uncertainty-based ensemble learning} could also help reduce the number of queries while maintaining high detection performance.

By addressing these challenges, our framework can evolve into a \emph{more adaptive, scalable, and interpretable anomaly detection system} that remains effective in dynamic and high-dimensional environments.

\section*{Conflicts of Interest}
The authors declare no conflict of interest.
\section*{Author contributions}
SB: Conceptualization, Methodology, Software,Visualization. SB, JC: Data curation. SB, PV, TR, JC: Investigation. SB, PV, TR, JC: Writing.

\bibliographystyle{IEEEtran}
\bibliography{references}

\end{document}